\documentclass[nohyperref]{article}

\usepackage{microtype}
\usepackage{graphicx}
\usepackage{subfigure}
\usepackage{booktabs} % for professional tables

\usepackage{hyperref}

\usepackage[accepted]{icml2023}

\usepackage{amsmath}
\usepackage{amssymb}
\usepackage{mathtools}
\usepackage{amsthm}
\usepackage{bm}
\usepackage{multirow}
\usepackage[capitalize,noabbrev]{cleveref}
\usepackage{caption}

\theoremstyle{plain}
\newtheorem{theorem}{Theorem}[section]

\theoremstyle{definition}
\newtheorem{definition}[theorem]{Definition}

\theoremstyle{remark}

\usepackage[textsize=tiny]{todonotes}

\icmltitlerunning{High-dimensional Clustering onto Hamiltonian Cycle}

\begin{document}

\twocolumn[
\icmltitle{High-dimensional Clustering onto Hamiltonian Cycle}




\begin{icmlauthorlist}
\icmlauthor{Tianyi Huang}{1,2}
\icmlauthor{Shenghui Cheng}{1,2}
\icmlauthor{Stan Z. Li}{1,2}
\icmlauthor{Zhengjun Zhang}{3,4}

\end{icmlauthorlist}

\icmlaffiliation{1}{School of Engineering, Westlake University, Hangzhou, China}
\icmlaffiliation{2}{Westlake Institute for Advanced Study, Hangzhou, China}
\icmlaffiliation{3}{School of Economics and Management, the University of Chinese Academy of Sciences, Beijing, China}
\icmlaffiliation{4}{School of Computer, Data \& Information Sciences, the University of Wisconsin, Madison, USA}

\icmlcorrespondingauthor{Shenghui Cheng}{Chengshenghui@westlake.edu.cn}

\icmlkeywords{Machine Learning, ICML}

\vskip 0.3in
]



\printAffiliationsAndNotice{}  % leave blank if no need to mention equal contribution

\begin{abstract}
Clustering aims to group unlabelled samples based on their similarities and is widespread in high-dimensional data analysis.
However, most of the clustering methods merely generate pseudo labels and thus are unable to simultaneously present the similarities between different clusters and outliers.
%For example, in image clustering for face identification, it is difficult to clean the face images that are in-between different clusters and low-quality face images without the presentation of similar clusters and outliers.
%This could lead to low identification accuracy.
This paper proposes a new framework called High-dimensional Clustering onto Hamiltonian Cycle (HCHC) to solve the above problems.
First, HCHC combines global structure with local structure in one objective function for deep clustering, improving the labels as relative probabilities, to mine the similarities between different clusters while keeping the local structure in each cluster.
Then, the anchors of different clusters are sorted on the optimal Hamiltonian cycle generated by the cluster similarities and mapped on the circumference of a circle.
Finally, a sample with a higher probability of a cluster will be mapped closer to the corresponding anchor.
In this way, our framework allows us to appreciate three aspects visually and simultaneously - clusters (formed by samples with high probabilities), cluster similarities (represented as circular distances), and outliers (recognized as dots far away from all clusters).
The theoretical analysis and experiments illustrate the superiority of HCHC.
\end{abstract}

\section{Introduction}
High-dimensional data, i.e., the data described by a large number of features, are widely existing in many research fields, such as image processing, pattern recognition, and bioinformatics~\cite{buhlmann2011statistics,donoho2000high}.
Analyzing high-dimensional data is a significant but challenging task~\cite{verleysen2003learning}.
Clustering is widespread in high-dimensional data analysis~\cite{mccarthy2004applications,jain1999data}.
It can group samples so that the samples within the same cluster are broadly more similar to one another than those in other clusters~\cite{hartigan1975clustering,niu2022spice,huang2020}.
Then by simultaneously presenting the clusters, similarities, and outliers, we can well recognize the insight of high-dimensional data.
For example, in the clustering of community detection, we can find the topological information between the different clusters of the community nodes by presenting the similarities between these clusters and thus explain the community detection result~\cite{fortunato2010community}.
%In image clustering for face identification, we can clean the face images that are in-between different clusters and low-quality face images by the presentation of similar clusters and outliers and then improve the identification accuracy.
However, most of the traditional clustering methods merely generate pseudo labels.
In this case, it is hard to get the knowledge of similarities and outliers in the clustering.

A frequently-used tool to explore the outliers and similarities between the different clusters is the dendrogram in hierarchical clustering~\cite{guha1998cure,karypis1999chameleon}.
But, some critical knowledge for explaining the recognized similar clusters and outliers cannot be presented in this way.
For example, the dendrogram cannot present the in-between samples of the different clusters to explain the recognized similar clusters.
It also cannot present the clustering probability distributions of the samples to explain the recognized outliers.
Deep clustering methods, such as DEC, have been proposed to cluster high-dimensional data by simultaneously learning clustering probability distributions and the embedded features of the samples~\cite{aljalbout2018clustering,xie2016unsupervised}.
Then, we can extract similarities and outliers in the generated clustering probability distributions~\cite{li2020consistent}.
Unfortunately, there does not exist an effective way to simultaneously present the mined knowledge, including clusters, similarities, and outliers.
Although some embedding methods, such as MDS and $t$-SNE~\cite{kruskal1978multidimensional,van2008visualizing,mcinnes2018umap}, can visualize the distances between the samples or the local-manifold in each cluster, the visualized result may be inconsistent with the clustering result.
One reason is that it is hard to visualize all of the distinguishing information of high-dimensional in $2$D space.
The other one is that these methods may have their limits in the visualization as shown in Appendix~\ref{SuppC11}~\cite{pmlr-v162-zu22a,li2020consistent}.
For example, in the embedding space of deep clustering, MDS cannot well keep the local-strcutre of data and $t$-SNE cannot capture the global structure of data.

This paper proposes High-dimensional Clustering onto Hamiltonian Cycle (HCHC) to solve the above problems.
It comprises two key components: (1) extracting the clustering probability distributions by deep clustering and (2) mapping the clustering probability distributions onto the optimal Hamiltonian cycle.
For extracting the clustering probability distributions, we construct a new deep clustering method, GLDC, by combining global structure with local structure in one objective function.
GLDC constructs a weighted adjacency matrix associated with a similarity graph of the samples.
By learning the structure of the unconnected samples and the connected samples, respectively, the loss function of GLDC can well mine the similarities between different clusters while keeping the local-structure of the samples in the same cluster.
Thus, the extracted clustering probability distributions in GLDC can well represent the similarities between different clusters and the samples in the same cluster.
Then, enlightened from RadViz Deluxe~\cite{Cheng2017,cheng2018colormap,grinstein2002information}, an effective visualization method to analyze the relationships between different features in data, we utilize the Hamiltonian cycle to present the mined knowledge in the extracted clustering probability distributions.
We find that by mapping the anchors of different clusters on the circumference of a circle with Hamiltonian cycle, the similarities between the different clusters can be well presented.
Concretely, to present the similarities between the different clusters, the anchors of different clusters are sorted on the optimal Hamiltonian cycle generated by the similarities between these clusters and mapped on the circumference of a circle by their orders.
Based on the polar coordinates of the anchors on the circumference, a sample with a higher probability of a cluster will be mapped closer to the corresponding anchor.
In this way, the data can be visualized by the following three aspects:
(1) the samples with a high probability in the same cluster can be mapped together;
(2) similar clusters can be mapped close to each other;
(3) the samples with low probability to any cluster can be regarded as outliers and mapped far away from all clusters.

An example of our HCHC is shown in Appendix~\ref{SuppA} and a head-to-head comparison between HCHC and Radviz Deluxe is shown in Appendix~\ref{SuppA1}.
Compared with visualizing the deep features by existing embedding methods, such as $t$-SNE, our HCHC can better match the clustering result while presenting the outliers and similarities between the different clusters.
Compared with the dendrogram in hierarchical clustering, our HCHC can better explain not only the recognized similar clusters by showing the in-between samples of different clusters but also the recognized outliers by showing their clustering probability distributions.
We perform experiments on six real-world datasets and a COVID-19 dataset to illustrate the effectiveness of our HCHC.
The source code can be downloaded from https://github.com/TianyiHuang2022.

\section{Related Work}
In this section, we review high-dimensional clustering and visualization.
They are highly related to our HCHC.

\subsection{High-dimensional Clustering}
Clustering has been a long-standing problem in machine learning~\cite{hartigan1975clustering,braun2022iterative,pmlr-v162-wang22r}.
There are many well-known clustering methods, such as $k$-means~\cite{macqueen1967some}, DBSCAN~\cite{ester1996density}, and Gaussian Mixture models~\cite{rasmussen1999infinite}.
Clustering can also be combined with other techniques like category discovery and semantic instance segmentation~\cite{shi2000normalized,wang2021progressive}.
However, clustering high-dimensional data is a hard issue, because of the large time complexity and the complex structure of data resulting from high-dimensional space~\cite{aljalbout2018clustering}.
Spectral clustering methods are often used to address this issue~\cite{macgregor2022tighter,ng2001spectral,von2007tutorial,bianchi2020spectral,pmlr-v162-macgregor22a}.
It explores the manifold structure of high-dimensional data in a low-dimensional space by the eigenvectors of the Laplacian matrix from the corresponding similarity graph.
Another important tool for clustering high-dimensional data is multitask clustering.
It can handle high-dimensional data by exploiting the knowledge shared by related tasks, such as inter-task clustering correlation and intra-task learning correlation~\cite{yang2014multitask,zhang2016self}.
Unfortunately, most of the above clustering methods just generate the pseudo sample label, a binary choice of belonging to a cluster or not, and thus are unable to represent the other interesting knowledge in high-dimensional clusterings, such as the similarities between clusters and outliers.

With the advances in deep learning, combining neural networks into clustering tasks has drawn significant attention in the literature ~\cite{chang2017deep,chen2015deep,ji2019invariant,tian2014learning}.
Deep clustering can mine not only clusters but also similarities and outliers in high-dimensional data by learning the clustering probability distributions.
We will give a brief introduction of the promising works in deep clustering including DEC, IDEC, deep spectral clustering, and data augmentation next~\cite{xie2016unsupervised,guo2017improved,shaham2018spectralnet,li2021contrastive}.

DEC starts with pretraining a nonlinear mapping by an autoencoder and then removes the decoder.
The remaining encoder is finetuned by minimizing the KL divergence.
To enhance the local structure preservation,
IDEC improves DEC by incorporating the autoencoder into DEC in the whole training process~\cite{guo2017improved}.
%Spectral clustering is a leading and highly popular field in clustering~\cite{shi2000normalized}.
%To keep the local manifold of data in a low dimensional space, it works by embedding data in the eigenspace of the Laplacian matrix, derived from the pairwise similarities, and then applying $k$-means to this representation to obtain the clusters.
By combining with deep learning, spectral clustering can work with large datasets~\cite{shaham2018spectralnet,bianchi2020spectral}.
In deep spectral clustering, the loss function is based on a weighted adjacency matrix associated with a similarity graph which includes the pairwise similarities between data points in each mini-batch.
In this way, the time complexity of deep spectral clustering is mainly dependent on the size of the mini-batch.
Recently, data augmentation is combined into deep image clustering and achieves great success, especially in contrastive learning~\cite{li2021contrastive,van2020scan,niu2022spice}.
The augmentations of an image sample can be computed by the transformation functions, e.g., random rotation, shifting, and cropping~\cite{guo2018deep,park2021improving}.
Then the loss function for learning the relationship between this sample and its augmentation can be computed by the dissimilarity between their corresponding embedded features.
%The differences in the deep clustering methods of data augmentation are mainly in the computation of this dissimilarity.

\subsection{High-dimensional Visualization}
High-dimensional visualization is the visual representation of the data with a large number of features~\cite{grinstein2001high}.
In this process, samples are mapped to numerical form and translated into a 2D graphical representation.
Many embedding methods can map high-dimensional data in a 2D space for visualization.
In this way, the important information from the features or the sample similarities in data can be well presented to us.
These methods are also used to visualize the embedded features in deep clustering for presenting the mined knowledge~\cite{xie2016unsupervised,huang2022consolidation}.

MDS~\cite{kruskal1978multidimensional}, PCA~\cite{abdi2010principal}, Isomap~\cite{balasubramanian2002isomap}, $t$-SNE~\cite{van2008visualizing} and UMAP~\cite{mcinnes2018umap} are five commonly used embedding methods.
MDS is a multivariate statistical method for estimating the scale values along one or more continuous dimensions such that those dimensions account for proximity measures defined over pairs of samples.
PCA extracts the important information from the features in a set of new orthogonal variables called principal components to display the similarities between the samples in the data.
Isomap extends classical multidimensional scaling by considering approximate geodesic distance instead of Euclidean distance.
t-SNE is a variation of stochastic neighbour embedding (SNE)~\cite{hinton2002stochastic}.
Compared with SNE, $t$-SNE is much easier to optimize and produces significantly better visualizations by reducing the tendency to crowd points together in the centre of the map.
UMAP is based on Riemannian geometry and algebraic topology. 
This method is competitive with $t$-SNE for visualization quality and arguably preserves more of the global structure with superior run time performance.
However, most of the above embedding methods only map the data samples without considering the relations between the features.

Radviz can solve this problem by assigning the features to points called dimensional anchors placed on the circumference of a circle~\cite{grinstein2002information}.
Original RadVis computes the polar coordinates of the anchors by a function of the effectiveness for discriminating each class of the samples in data~\cite{mccarthy2004applications}.
The details of RadVis are in Appendix~\ref{SuppB}.
In unsupervised learning, the polar coordinates of the anchors can be computed by the similarities between the features~\cite{sharko2008vectorized}.
Then the locations of the samples are determined by a weighting formula where sample features with higher values will receive a higher attraction to the corresponding anchors. 
%There are also some other visualization methods similar to RadViz.
%For example, star coordinate visualization maps each sample by computing linear combinations of its feature values that represent radial axes~\cite{kandogan2000star}.
%Parallel coordinate displays each feature as a vertical axis and each sample is a piecewise linear line going across each axis at its respective feature values~\cite{inselberg1990parallel}.
%However, these methods still cannot present the relations between the features well enough without mapping the similarities between these features.
As an improvement of RadViz, RadViz Deluxe sorts the feature anchors by Hamiltonian cycle and then maps the similarities between these features by adjusting the distances between the anchors on the circumference~\cite{Cheng2017}.
Thus the similarities between the different features can be better presented.

\section{GLDC onto Hamiltonian Cycle}
In this section, we propose HCHC to cluster high-dimensional data and then visualize the clustering results including clusters, cluster similarities, and outliers.
The details of the motivation of HCHC are shown in Appendix~\ref{SuppC}.
\begin{figure}[h]
\centering
  \includegraphics[width=3.2in]{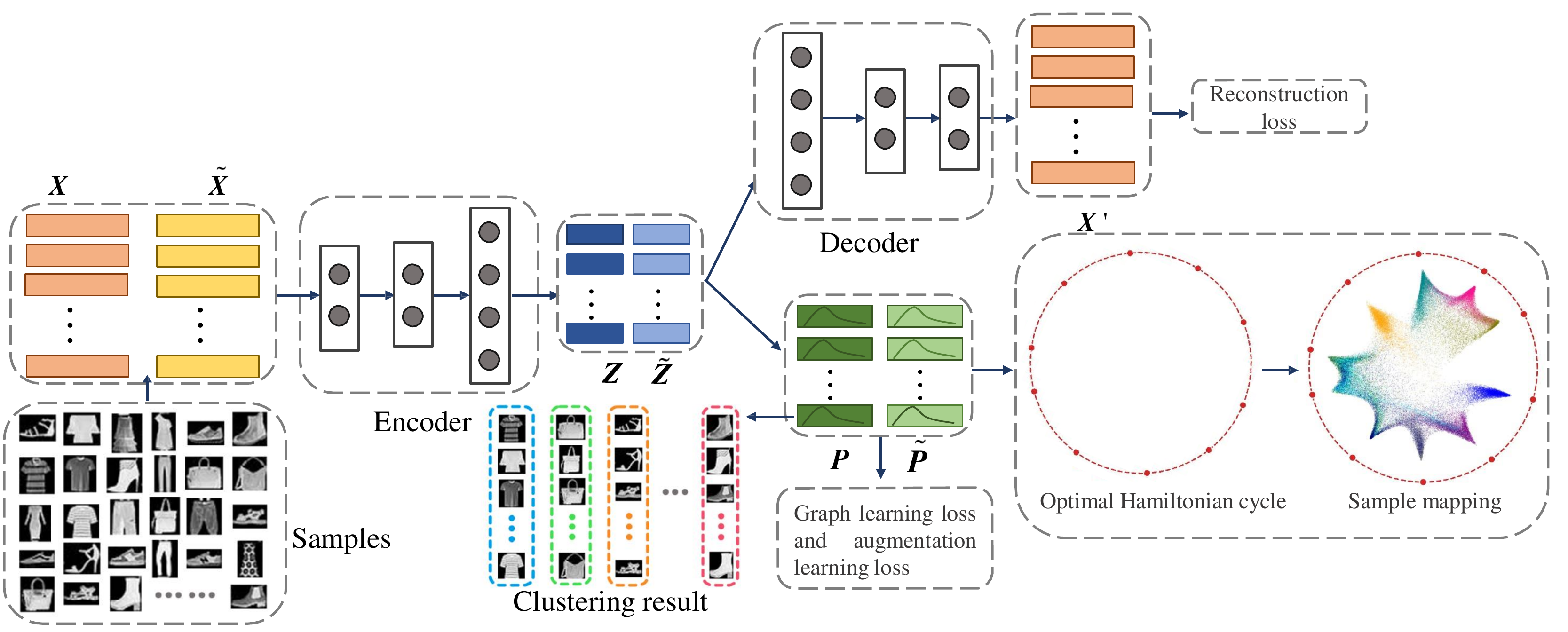}\\
  \caption{The architecture of our HCHC.}
  \label{framework}
\end{figure}
The architecture of HCHC is shown in Fig.~\ref{framework}.
The dataset $\bm{X}$ and its augmentation $\bm{\widetilde{X}}$ are the inputs of our GLDC network.
In the pretraining of our GLDC, the autoencoder is trained by minimizing the reconstruction loss to encode $\bm{X}$ in $\bm{Z}$.
After pretraining, GLDC computes the clustering probability distribution of each sample in $\bm{Z}$ by minimizing our clustering loss which includes the reconstruction loss, the graph learning loss, and the augmentation learning loss.
Finally, the clustering probability distributions from GLDC are visualized based on a Hamiltonian cycle.

\subsection{Combining Global-structure with Local-structure}
In this subsection, we propose a new deep clustering method, GLDC, by incorporating global-structure with local-structure in one objective function.
At the beginning of GLDC, self-training is used as the pretraining by minimizing the reconstruction loss of data.
In this way, we can capture the most salient features~\cite{guo2017improved,lin2007large}.

For a given dataset $\bm{X}=\{\bm{x_1};\bm{x_2};\cdots; \bm{x_n}\}$, an encoder $G_{\theta}$, and the corresponding decoder $G_{\theta'}$, the reconstruction loss in the pretraining can be written as
\begin{small} 
\begin{eqnarray}
\label{mresloss}
L_r = \sum_{i}^{n}||\bm{x_i}-G_{\theta'}(\bm{z_{i}})||^2_2 = \sum_{i}^{n}||\bm{x_i}-G_{\theta'}(G_{\theta}(\bm{x_{i}}))||^2_2
\end{eqnarray}
\end{small} 
%In this process, the encoder $G_{\theta}$ and decoder $G_{\theta'}$ are updated by
%\begin{eqnarray}
%\label{preupdate1}
% \theta = \theta - \lambda\frac{\partial L_r}{\partial \theta} \\
%\label{preupdate2}
% \theta' = \theta' - \lambda\frac{\partial L_r}{\partial \theta'}
%\end{eqnarray}
After the pretraining, in the space of $\bm{Z}$ of the $l$-th mini-batch, the network is optimized by a weighted adjacency graph which is associated with a weighted adjacency matrix $\bm{W^{l}}$.
Each entry in $\bm{W^{l}}$ is defined as
\begin{eqnarray}
\label{congraph}
w_{i,j}^{l} =
\begin{cases}
e^{-\frac{||\bm{z_i}-\bm{z_j}||^2_2}{\sigma^2}}&      \bm{x_{j}} \in \mathcal{N}^{k}_{i}\\
0,&      \text{otherwise}\\
\end{cases}
\end{eqnarray}
where $\mathcal{N}^{K}_{i}$ is the set of $k$-nearest neighbors of $\bm{z_{i}}$ in the mini-batch.
Different from the existing deep spectral clustering where merely the local-structure of the data is represented in $\bm{Z}$, our GLDC represents the local-structure and the global-structure of data in the output $\bm{p_{i}}$, the learned probability distribution of $\bm{x_{i}}$ for different clusters.
In our network,
we calculate the $\bm {p_i}=\{p_{i,1},...,p_{i,c}\}$ by 
\begin{eqnarray}
\label{compro}
\bm{p_i} = \mathcal{P}_{\phi}(\bm{z_i}) = \mathcal{P}_{\phi}(G_{\theta}(\bm{x_{i}}))
\end{eqnarray}

Then the graph learning loss based on $\bm{W^{l}}$ in the mini-batch is defined as~\cite{rebuffi2021lsd}.
\begin{eqnarray}
L_w =-\frac{1}{B^2}(\sum_{i,j}^{B} w_{i,j}^l\log P(i = j) \nonumber\\
+(1-w_{i,j}^l)\log P(i \neq j))
\end{eqnarray}
where $P(i = j)$ is the probability that $\bm{x_i}$ and $\bm{x_j}$ belong to the same cluster while $P(i \neq j)$  is the probability that $\bm{x_i}$ and $\bm{x_j}$ belong to the different clusters.
The item $w_{i,j}^l\log P(i = j)$ is to make the connected samples in the adjacency matrix have similar clustering probability distributions and thus can keep the local-structure of data in the same cluster.
$(1-w_{i,j}^l)\log P(i \neq j)$ is to make the unconnected samples have diverse clustering probability distributions, i.e., far points should be in different clusters, and thus can be used to analyze the global-structure of the data.
Therefore, in our GLDC clustering, the local structure of data and the similarities between different clusters can be considered together and presented in our visualization.
Because
\begin{eqnarray}
P(i = j) &=&\sum_{h=1}^{c} P(i=h,j=h) \nonumber\\
&=&\sum_{h=1}^{c} p_{i,h} \times p_{j,h} \nonumber\\
&=&\bm{p_i}^{\intercal}\bm{p_j}
\end{eqnarray}
where $c$ is the cluster number, $L_w$ can be written as~\cite{rebuffi2021lsd}
\begin{eqnarray}
\label{graphloss}
L_w =-\frac{1}{B^2}(\sum_{i,j}^Bw_{i,j}^l\log \bm{p_i}^{\intercal}\bm{p_j} +  \nonumber\\
(1-w_{i,j}^l)\log (1-\bm{p_i}^{\intercal}\bm{p_j}))
\end{eqnarray} 
We also use a generalized data augmentation method to improve our clustering.
In our GLDC, $T(\bm{x_i})$ is defined as
\begin{eqnarray}
\label{genaug}
\bm{\widetilde{x}_i} = T(\bm{x_i}) = \bm{x_i} + \epsilon, \epsilon \sim \mathcal{N}(0,\xi)
\end{eqnarray}
where $\mathcal{N}(0,\xi)$ is a Gaussian distribution.
Define $\bm{\widetilde{p}_i}$ as the output of the probability of $\bm{\widetilde{x}_i}$, then the loss of the data augmentation learning can be defined as
\begin{eqnarray}
\label{augloss}
L_a = \sum_{i}||\bm{p_i} - \bm{\widetilde{p}_i}||^2_2
\end{eqnarray}

The overall clustering loss is given by
\begin{eqnarray}
\label{cluloss}
L_{clu} =L_r + \beta_1 L_w+ \beta_2 L_a
\end{eqnarray}
The algorithm of GLDC is summarized in Appendix~\ref{SuppC1}.
For the input dataset $\bm{X} = \{\bm{x_1},\bm{x_2}, \cdots, \bm{x_n}\}$, in this algorithm we can get the distributions $\bm{P} = \{\bm{p_1},\bm{p_2}, \cdots, \bm{p_n}\}$, where $p_{i,j}$ is the the probability that $\bm{x_i}$ belongs to cluster $j$.
Based on $\bm{p_i}$, the label $c_i$ assigned to $\bm{x_i}$ can be obtained by
\begin{eqnarray}
c_i = \arg \max_{j}p_{i,j}
\end{eqnarray}

%%%%%%%%%%%%%%%%%%%%%%%%%%%%%%%%%%%%%%%%%%%%%%%%%%%%%%%%%%%%%%%%%%%%%%%%%%%%%%%%%%%%%%
\subsection{Mapping the Distributions by the Optimal Hamiltonian Cycle}
\begin{figure}
\centering
  \includegraphics[width=3.2in]{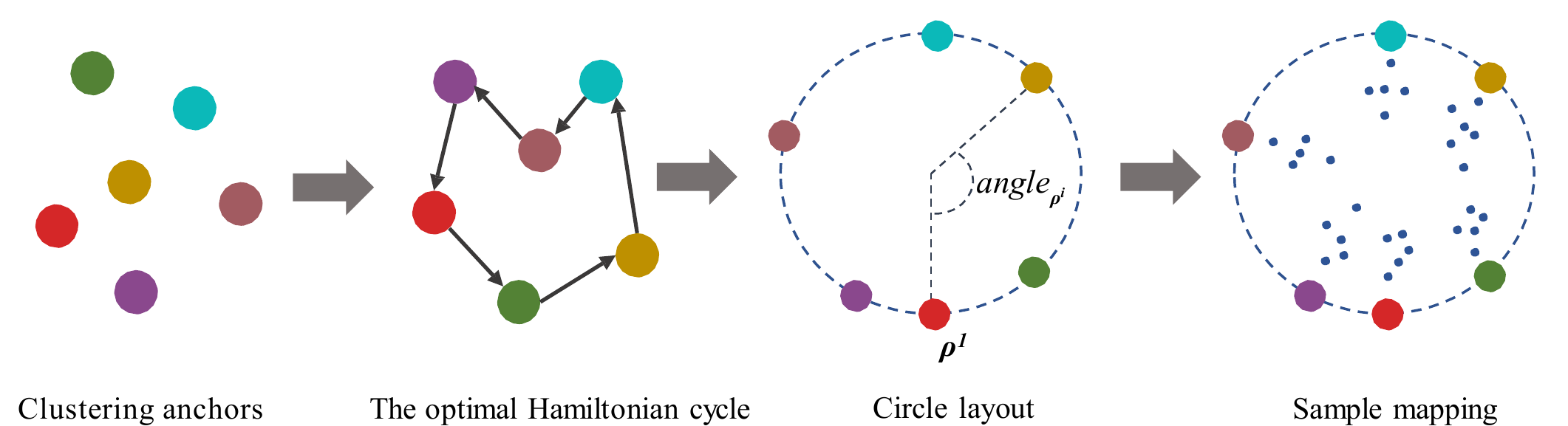}\\
  \caption{Illustration of our mapping process by optimal Hamiltonian cycle.}
  \label{mappingprocess}
\end{figure}

After the deep clustering, we can get the distribution matrix $\bm{P} = \{\bm{\rho_1};\bm{\rho_2}; \cdots; \bm{\rho_c}\}$, where $c$ is the number of the clusters and $\bm{\rho_i}=\{p_{1,i},p_{2,i},\cdots, p_{n,i}\}$.
Then we show the clustering results in $\bm{P}$ by the optimal Hamiltonian cycle of the different clusters.
This is the shortest cycle that passes through every $\bm{\rho_i}$ exactly once, except the first passed one~\cite{dirac1952some}.
The mapping process is shown in Fig.~\ref{mappingprocess} and detailed as follows.

First, we use Pearson correlation coefficient to compute the similarities between $\bm{\rho_i}$ and $\bm{\rho_j}$ as
\begin{small}
\begin{eqnarray}
\label{rhop}
s(\bm{\rho_i},\bm{\rho_j}) = \frac{\sum_{l=1}^{n}(p_{l,i}-\overline{\bm{\rho_i}})(p_{l,j}-\overline{\bm{\rho_j}})}{\sqrt{\sum_{l=1}^{n}(p_{l,i}-\overline{\bm{\rho_i}})^2} \sqrt{\sum_{l=1}^{n}(p_{l,j}-\overline{\bm{\rho_j}})^2}}
\end{eqnarray}
\end{small}
Thus, the dissimilarities between the $\bm{\rho_i}$ and $\bm{\rho_j}$ can be defined as
\begin{eqnarray}
\label{disfp}
dis(\bm{\rho_i},\bm{\rho_j}) = \frac{1-s(\bm{\rho_i},\bm{\rho_j})}{\sum_{i=1}^{c-1}\sum_{j=i+1}^{c}(1-s(\bm{\rho_i},\bm{\rho_j}))}
\end{eqnarray}
Secondly, based on the above dissimilarities, $\{\bm{\rho_1};\bm{\rho_2};\cdots; \bm{\rho_c}\}$ are ordered by an optimal Hamiltonian cycle.
We use dynamic programming to get the optimal Hamiltonian cycle by the following definition.

\begin{definition}
Define $dp(\bm{\rho_i},\bm{\rho_j},\bm{state})$ as the minimum path between $\bm{\rho_i}$ and $\bm{\rho_j}$ with the path state $\bm{state}=\{pass(\bm{\rho_c}),
\cdots pass(\bm{\rho_2}),pass(\bm{\rho_1})\}$,
where if $\bm{\rho_i}$ has been passed, $pass(\bm{\rho_i})=1$, otherwise, $pass(\bm{\rho_i})=0$.
\end{definition}

In the dynamic programming, $dp(\bm{\rho_i},\bm{\rho_j},\bm{state})$ can be updated by
\begin{eqnarray}
\label{dpfhc} 
dp(\bm{\rho_i},\bm{\rho_j},\bm{state})= \min \{dp(\bm{\rho_i},\bm{\rho_k},\bm{state} \nonumber\\
\oplus (1<<(j-1)) + dis(\bm{\rho_k},\bm{\rho_j})\}
\end{eqnarray}
where $\oplus$ is XOR operation and $<<$ is shift-arithmetic-left operation.
Then we can get $\Pi^* = \{\bm{\rho^1};\bm{\rho^2};\cdots; \bm{\rho^c}\}$ as the optimal Hamiltonian cycle of each cluster by the following definition.

\begin{definition}
Let $\bm{\rho^i} \in \{\bm{\rho_1},\cdots, \bm{\rho_c}\}$ be the $i$-th passed vertice of a Hamiltonian cycle.
Then this Hamiltonian cycle can be defined as $\Pi = \{\bm{\rho^1};\bm{\rho^2};\cdots; \bm{\rho^c}\}$.
\end{definition}

The angle $\alpha_{\bm{\rho^i}}$ of $\bm{\rho^i}$ is used to map the anchors of different clusters on a circle and can be computed by
\begin{eqnarray}
\label{anglef}
\alpha_{\bm{\rho^i}} =
\begin{cases}
0,&      i=1\\
\alpha_{\bm{\rho^{i-1}}}+\\ 
2\pi \frac{dis(\bm{\rho^{i}},\bm{\rho^{i-1}})}{\sum_{j=2}^{c}dis(\bm{\rho^{j}},\bm{\rho^{j-1}})+ dis(\bm{\rho^{c}},\bm{\rho^{1}})}&       \text{otherwise}\\
\end{cases}
\end{eqnarray}
Thirdly, based on $\alpha_{\bm{\rho^i}}$ the position of the anchor of $\bm{\rho^i}$ on a circle can be computed by
\begin{eqnarray}
\label{pvf}
\bm{\mu_{\rho^i}} = [r \times \cos(\alpha_{\bm{\rho^i}}), r \times \sin(\alpha_{\bm{\rho^i}})]
\end{eqnarray}
Finally, the position of $\bm{x_i}$ in the circle can be computed by 
\begin{eqnarray}
\label{pvp}
\bm{\mu_{x_i}} = \sum_{j=1}^{c}\frac{p_{i,j}}{\sum_{k=1}^{c}p_{i,k}}\bm{\mu_{\rho^j}} = \sum_{j=1}^{c}p_{i,j}\bm{\mu_{\rho^j}}
\end{eqnarray}
with $\sum_{k=1}^{c}p_{i,k}=1$.
A Hamiltonian cycle can present the cluster similarities $\{s(\bm{\rho^1},\bm{\rho^2}),\cdots,$ $s(\bm{\rho^{c-1}},\bm{\rho^c}), s(\bm{\rho^1},\bm{\rho^c})\} \in \{s(\bm{\rho_i},\bm{\rho_j})|i = 1,\cdots,c,i<j\}$ 
and based on the following theorem, the optimal Hamiltonian cycle will select high similarities between the clusters.
\begin{theorem}
\label{theorem1}
For the cluster similarities $\{s(\bm{\rho_i},\bm{\rho_j})|i = 1,\cdots,c-1,i<j\}$, the mapping of the optimal Hamiltonian cycle will maximize
\begin{eqnarray}
S_{sam} =\sum_{i=1}^{c-1}s(\bm{\rho^i},\bm{\rho^{i+1}})+s(\bm{\rho^1},\bm{\rho^c})
\end{eqnarray}
\end{theorem}
Therefore, by the optimal Hamiltonian cycle, we can get the global optimal path to make similar clusters close to each other in the circumference of a circle.
In this way, the similarity between $\bm{\rho^{i}}$ and $\bm{\rho^{i+k(k>1)}}$ can be measured by the corresponding geodesic distance on this optimal Hamiltonian cycle.
The algorithm of our mapping is presented in Appendix~\ref{SuppC1}
The theoretical analysis to proof theorem~\ref{theorem1} is in Appendix~\ref{SuppD1}.
The theoretical analysis in Appendix~\ref{SuppD2} illustrates that the optimal Hamiltonian cycle can improve the sample mapping based on the following assumption.
The further away a point is from the center, the more informative its position is, being the point closer to the attributes having the highest values~\cite{angelini2019towards}.

\section{Experimental Results}
In this section, first, we show the visualized results of our HCHC and other visualization methods.
Then, we compare our GLDC with different clustering methods.
Finally, we analyze a dataset of COVID-19 by HCHC.
The experimental setting is shown in Appendix~\ref{SuppE}.
The time cost analysis and case study are shown in Appendix~\ref{SuppF2} and \ref{SuppF4}, respectively.
The parameter analysis is shown in Appendix~\ref{SuppF5}.

\subsection{Visualized Result}
\begin{figure*}
\center
\subfigure[MNIST]{\includegraphics[width=0.87in]{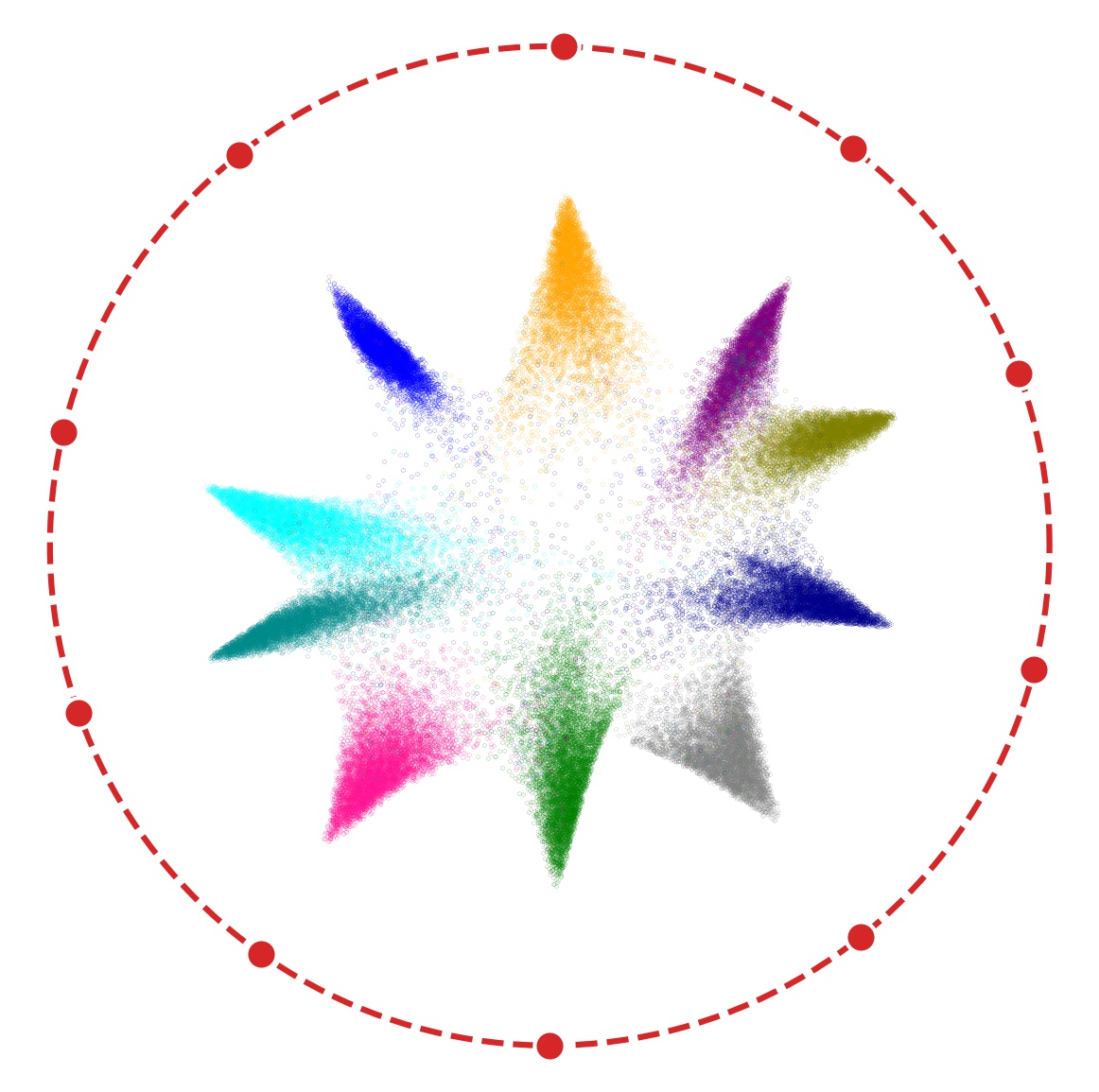}}
\subfigure[Fashion]{\includegraphics[width=0.87in]{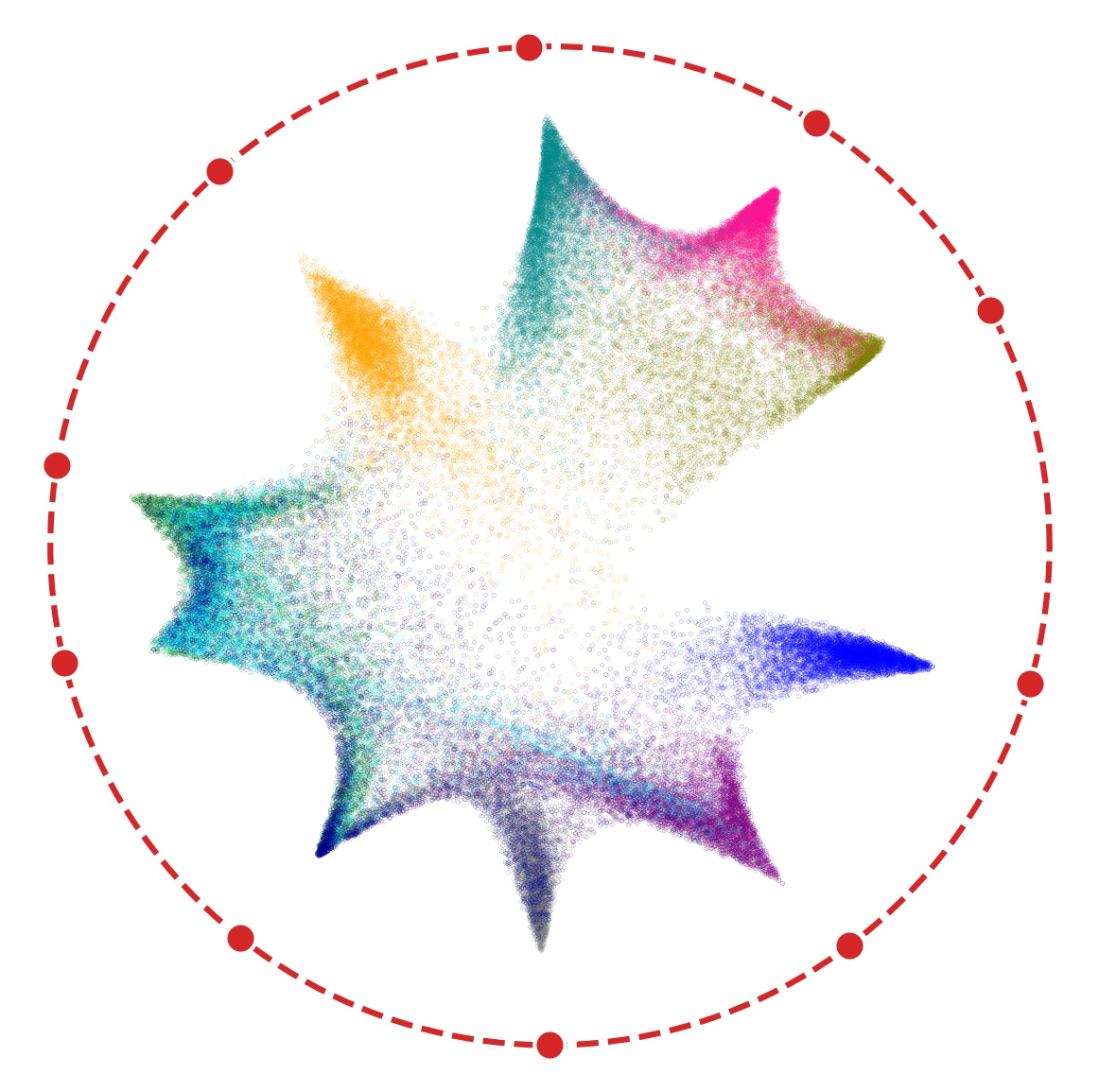}}
\subfigure[USPS]{\includegraphics[width=0.87in]{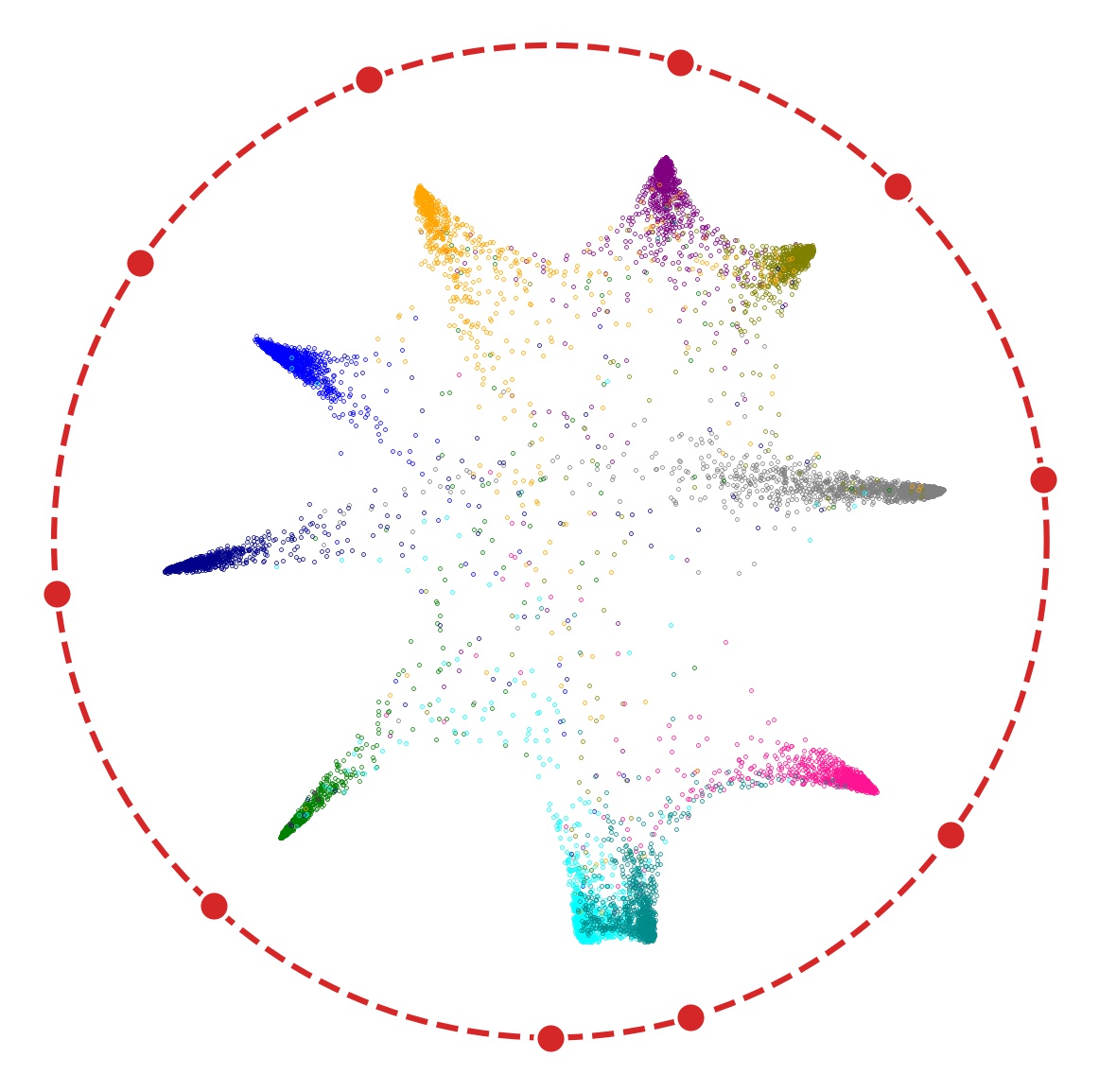}}
\subfigure[Reuters10k]{\includegraphics[width=0.87in]{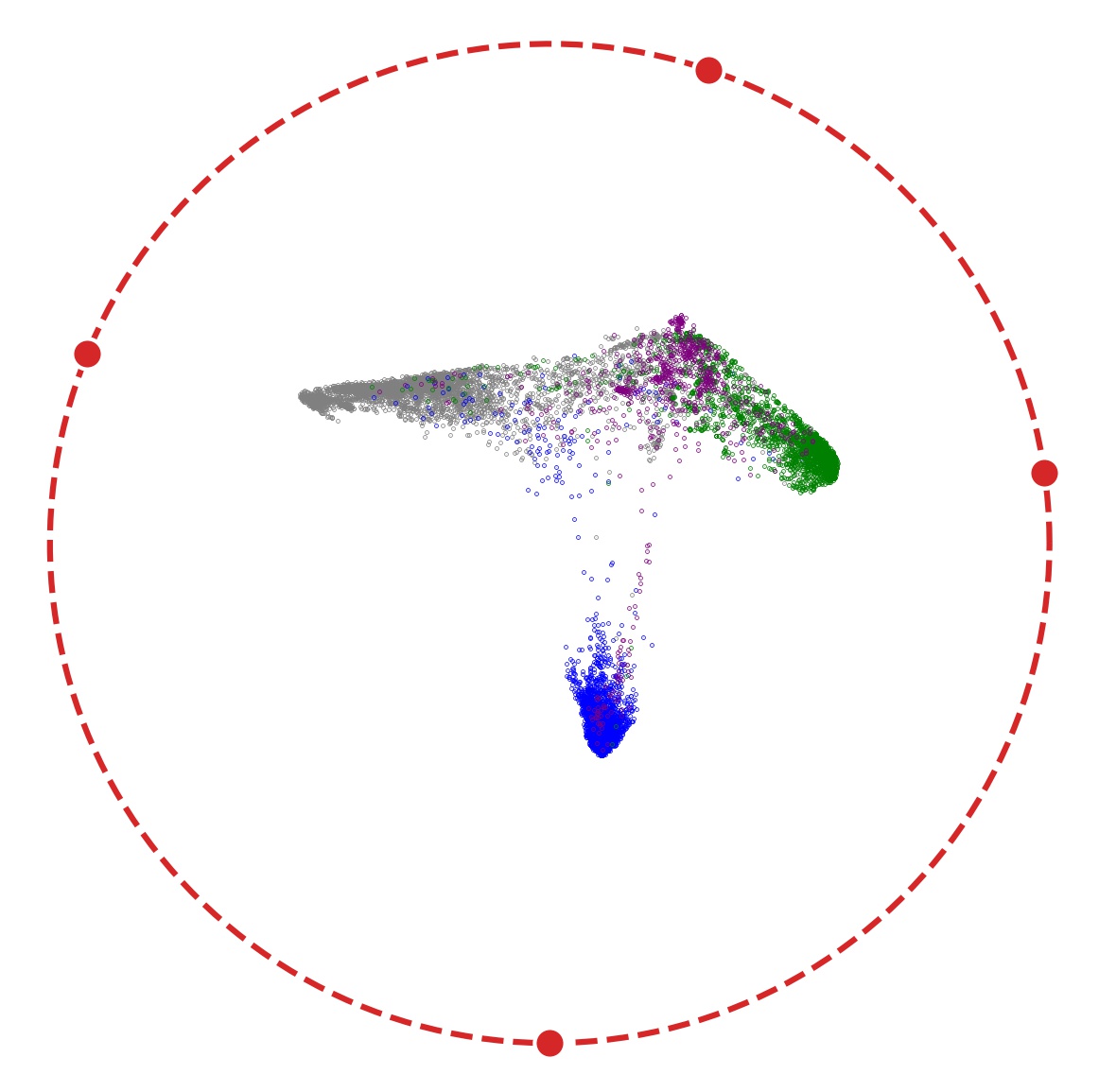}}
\subfigure[HHAR]{\includegraphics[width=0.87in]{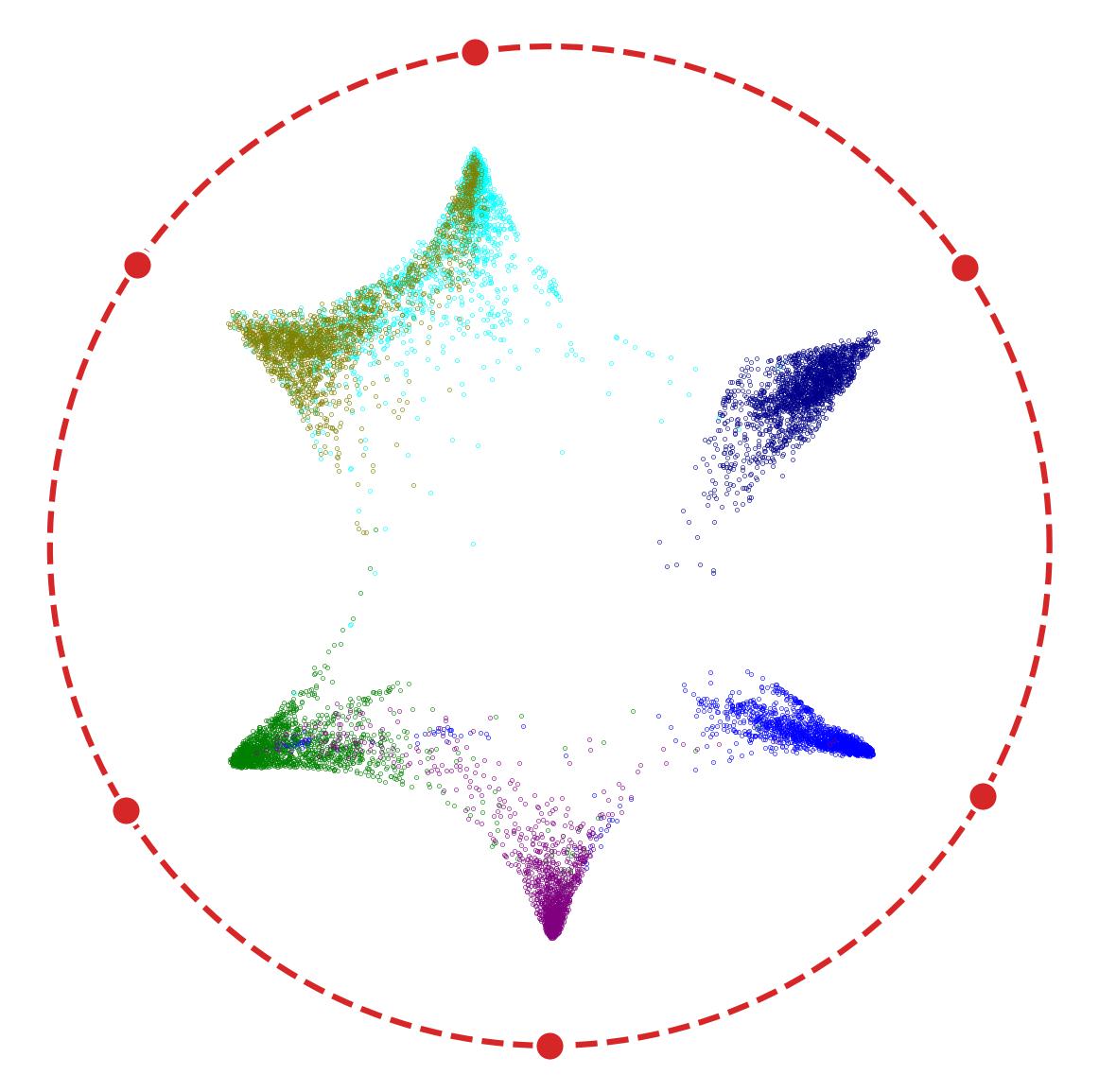}}
\subfigure[Pendigit]{\includegraphics[width=0.87in]{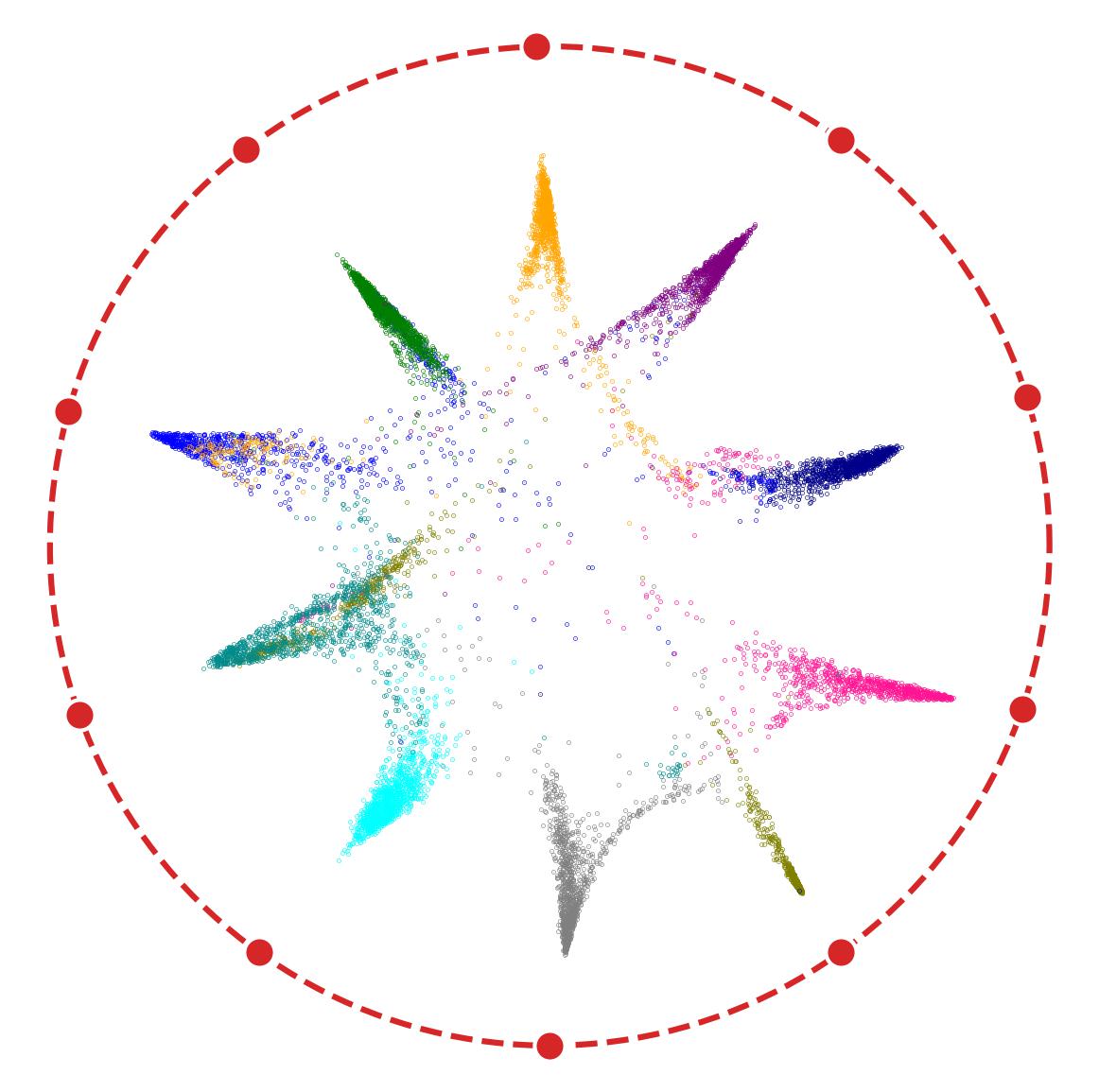}}
\subfigure[BH]{\includegraphics[width=0.87in]{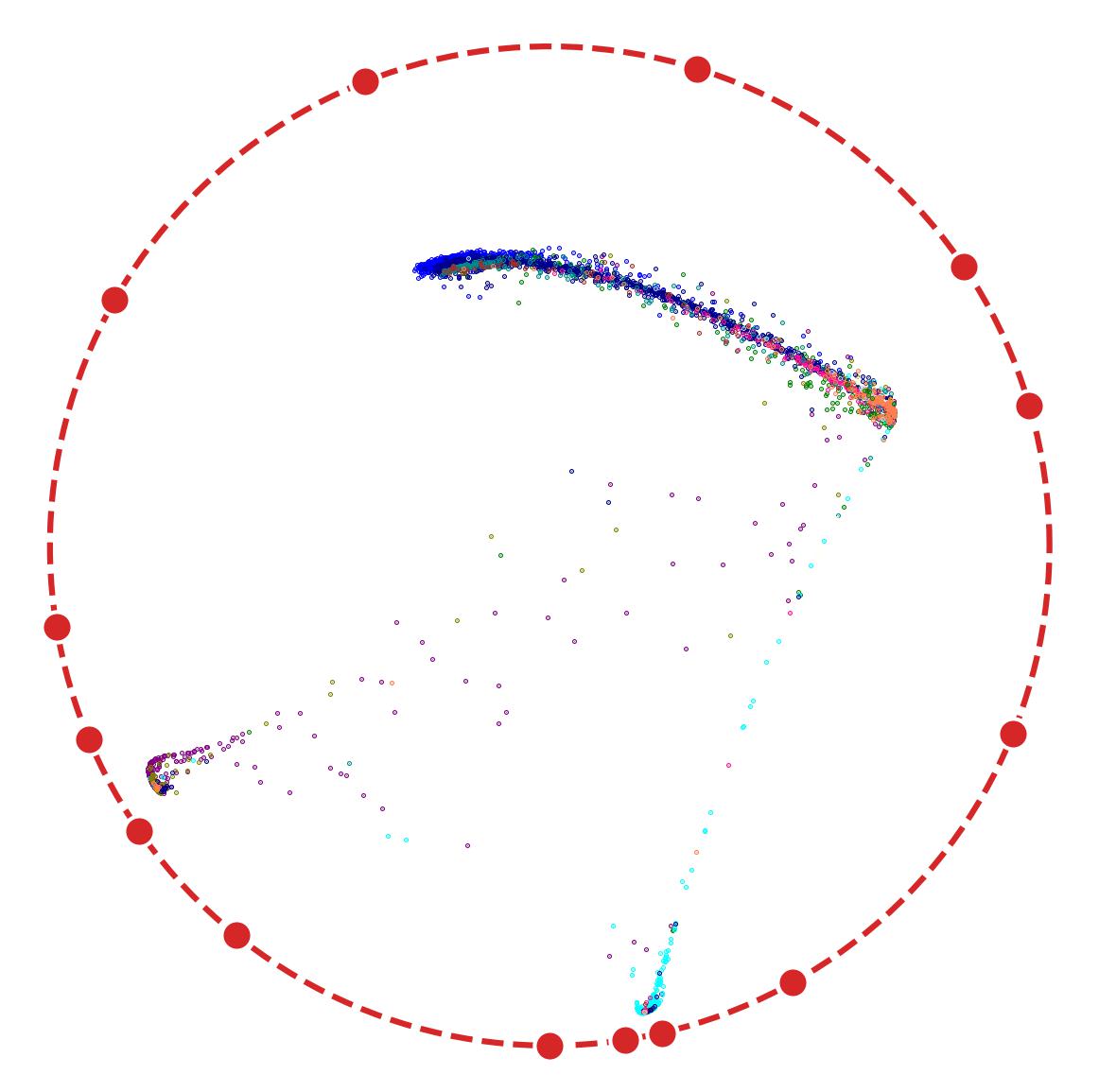}}
\caption{Visualized results on different datasets.}
\label{visall}
\end{figure*}

The visualized results of HCHC with MNIST~\cite{deng2012mnist}, Fashion~\cite{xiao2017fashion}, USPS~\cite{hull1994database}, Reuters10k~\cite{lewis2004rcv1}, HHAR~\cite{stisen2015smart}, Pendigits~\cite{asuncion2007uci}, and BH~\cite{abdelaal2019comparison} are shown in Fig.~\ref{visall}.

\begin{figure}
\center
\subfigure[MNIST]{\includegraphics[width=1.5in]{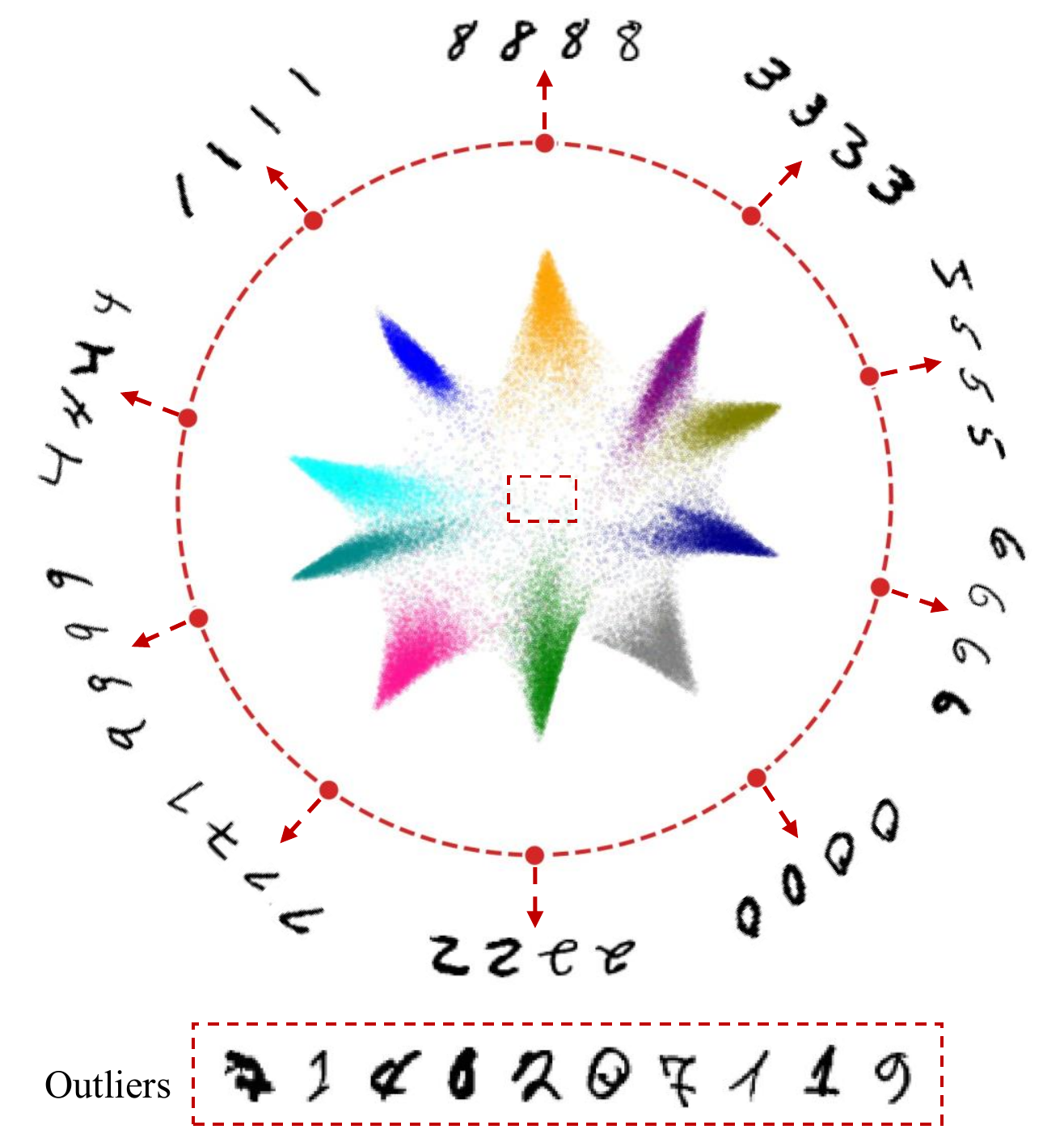}}
\subfigure[Fashion]{\includegraphics[width=1.5in]{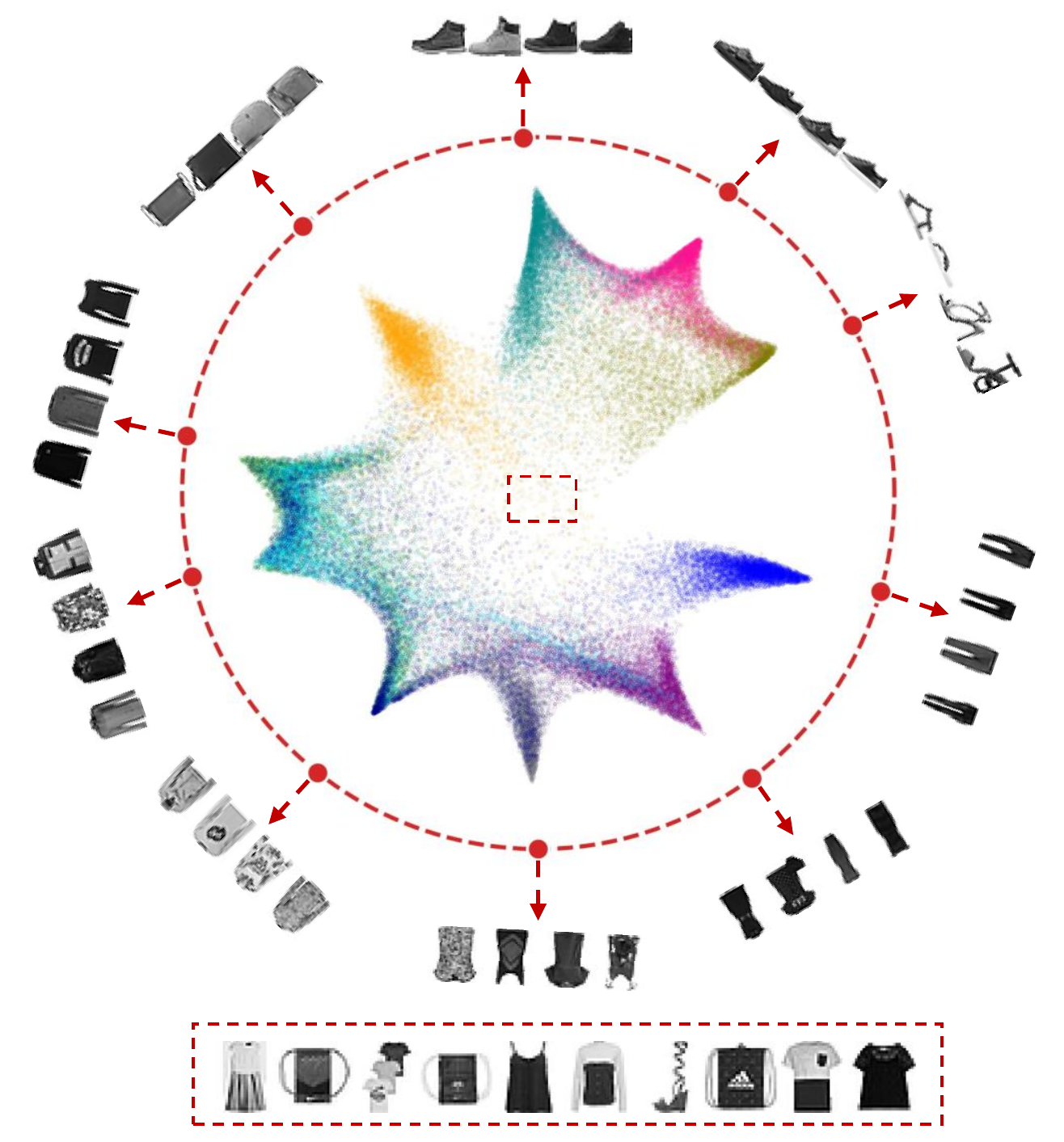}}
\caption{The clusters, samples, labels, and outliers on MNIST and Fashion, respectively.}
\label{MFv}
\end{figure}
\begin{table*}
\scriptsize
\centering
\caption{ACCs and NMIs of different clustering methods.}
\label{Tab03}
\begin{tabular}{ccccccccccccccc}
\toprule
\multirow{2}{*}{Method} & \multicolumn{2}{c}{MNIST} & \multicolumn{2}{c}{Fashion} & \multicolumn{2}{c}{USPS}  & \multicolumn{2}{c}{Reuter10K}  & \multicolumn{2}{c}{HHAR}  & \multicolumn{2}{c}{Pendigits} & \multicolumn{2}{c}{BH}\\
\cmidrule(r){2-3} \cmidrule(r){4-5} \cmidrule(r){6-7} \cmidrule(r){8-9} \cmidrule(r){10-11} \cmidrule(r){12-13} \cmidrule(r){14-15}
&  ACC      &  NMI 
&  ACC      &  NMI 
&  ACC      &  NMI  
&  ACC      &  NMI 
&  ACC      &  NMI
&  ACC      &  NMI
&  ACC      &  NMI   \\
\midrule
$k$-means    &0.532   & 0.500   & 0.474    & 0.512     & 0.668    & 0.601    & 0.559     & 0.375    & 0.599    & 0.588     & 0.666    & 0.681   & 0.492  & 0.561\\
GMM          &0.439   & 0.356   & 0.476    & 0.532     & 0.553    & 0.529    & 0.665     & 0.430    & 0.597    & 0.593     & 0.673   & 0.682   & 0.593   & 0.656\\
SC           &0.680   & 0.759   & 0.551    & 0.630     & 0.712    & 0.656    & 0.658     & 0.401    & 0.538    & 0.741      & 0.724    & 0.784 & 0.554  & 0.601\\
DEC          &0.863   & 0.834   & 0.518    & 0.546     & 0.762    & 0.767    & 0.773     & 0.528    & 0.764    & 0.700       & 0.776    & 0.706   & 0.648   & 0.618\\
IDEC         &0.881   & 0.867   & 0.529    & 0.557     & 0.761    & 0.785    & 0.785     & 0.541    & 0.722    & 0.785        & 0.793    & 0.742   & 0.406   & 0.548\\
DSC          &0.938   & 0.873   & 0.633    & 0.647     & 0.866    & 0.859    & 0.725     & 0.472    & 0.713    & 0.764       & 0.820    & 0.791 & 0.607    & 0.492\\
JULE         &0.964   & 0.913   & 0.563    & 0.608     & 0.950    & 0.913    & -         & -        & -        & -             & -    & -    & -    & - \\
DSCDAN       &0.978   & 0.941   & 0.662    & 0.645     & 0.869    & 0.857    & -         & -        & -        & -             & -    & -      & -    & - \\
N2D          &0.979   & 0.942   & 0.672    & 0.684     & 0.958    & 0.901    & 0.784     & 0.536    & 0.801    & 0.783          & 0.885    & 0.863  & 0.554    & 0.570\\
\midrule
GLDC         &0.979   & 0.941   & 0.715    & 0.691     & 0.910    & 0.862    & 0.834     & 0.629   & 0.878    & 0.821       & 0.867   & 0.830  & 0.704  & 0.655\\
GLDC(w/o $L_a$)   &0.965   & 0.914   & 0.694    & 0.652     & 0.879    & 0.817    & 0.804     & 0.587    & 0.822    & 0.743        & 0.803    & 0.744  & 0.687   & 0.614\\
GLDC(w/o $L_r$)   &0.975   & 0.932   & 0.621    & 0.635     & 0.825    & 0.787    & 0.724     & 0.506    & 0.765    & 0.723        & 0.781    & 0.768   & 0.680    & 0.591\\
\bottomrule
\end{tabular}
\end{table*}
From Fig.~\ref{visall} (a), we can see that the different classes in MNIST are well separated by the clustering in GLDC.
The visualized result of the samples is consistent with their labels.
By HCHC, we can also see the similarities between the different classes such as that the cyan class and the dark cyan class are closer to each other than other pairs of classes.

From Fig.~\ref{visall} (b), we can see that the outliers in Fashion are much more than those in MNIST.
The orange class and blue class can be well separated from the others.
The similarities between the different clusters can also be shown in Fashion by HCHC.
Then we can see that the samples in the dark-blue class, green class, and cyan class are easily mixed in clustering.
This is a piece of important information for us to improve the clustering for Fashion dataset.

From Fig.~\ref{visall} (c), we can see that as the visualized result of MNIST, the different clusters in USPS are well separated.
However, some samples in the cyan class and the dark cyan class are easily mixed.

From Fig.~\ref{visall} (d), we can see that for Reuters10k, the blue class, the green class, and the grey class are well separated by GLDC.
However, the samples in the purple class are mixed with the samples in the green class and the grey class.

From Fig.~\ref{visall} (e), we can see that most of the classes in HHAR are well separated by GLDC.
However, some samples in the cyan class and olive class are easily mixed.

From Fig.~\ref{visall} (f), we can see that most of the classes in Pendigits are also well separated by GLDC.
However, some samples in the dark cyan class and olive class are easily mixed while some samples in the orange class and blue class are also easily mixed.

From Fig.~\ref{visall} (g), we can see that BH is a challenging dataset.
GLDC can correctly cluster most of the samples in the purple class, cycan class, and orange class.

Overall, the above results illustrate that HCHC can well mine the clusters, similarities, and outliers in real-world datasets.
In most cases, the mined result of a dataset can be highly related to the corresponding labels.

Fig.~\ref{MFv} (a) and (b) show the clustering results and the visualizations of Fashion and MNIST by HCHC, respectively.
In this Figure, we show the randomly selected 4 images of each cluster and the recognized outliers.
As we can see, by HCHC the similar clusters are close to each other and the outliers can be well recognized.
For example, as shown in Fig.~\ref{MFv} (a) the shapes of digit numbers “9” and “4” are similar, so their clusters are mapped close to each other.
HCHC also can recognize the illegible handwritten digits as the outliers.
As shown in Fig.~\ref{MFv} (b) for Fashion, HCHC can put the clusters of different types of shoes together, it also can recognize the outliers in the data.

\begin{figure}[h]
\center
\subfigure[HCHC]{\includegraphics[width=1in]{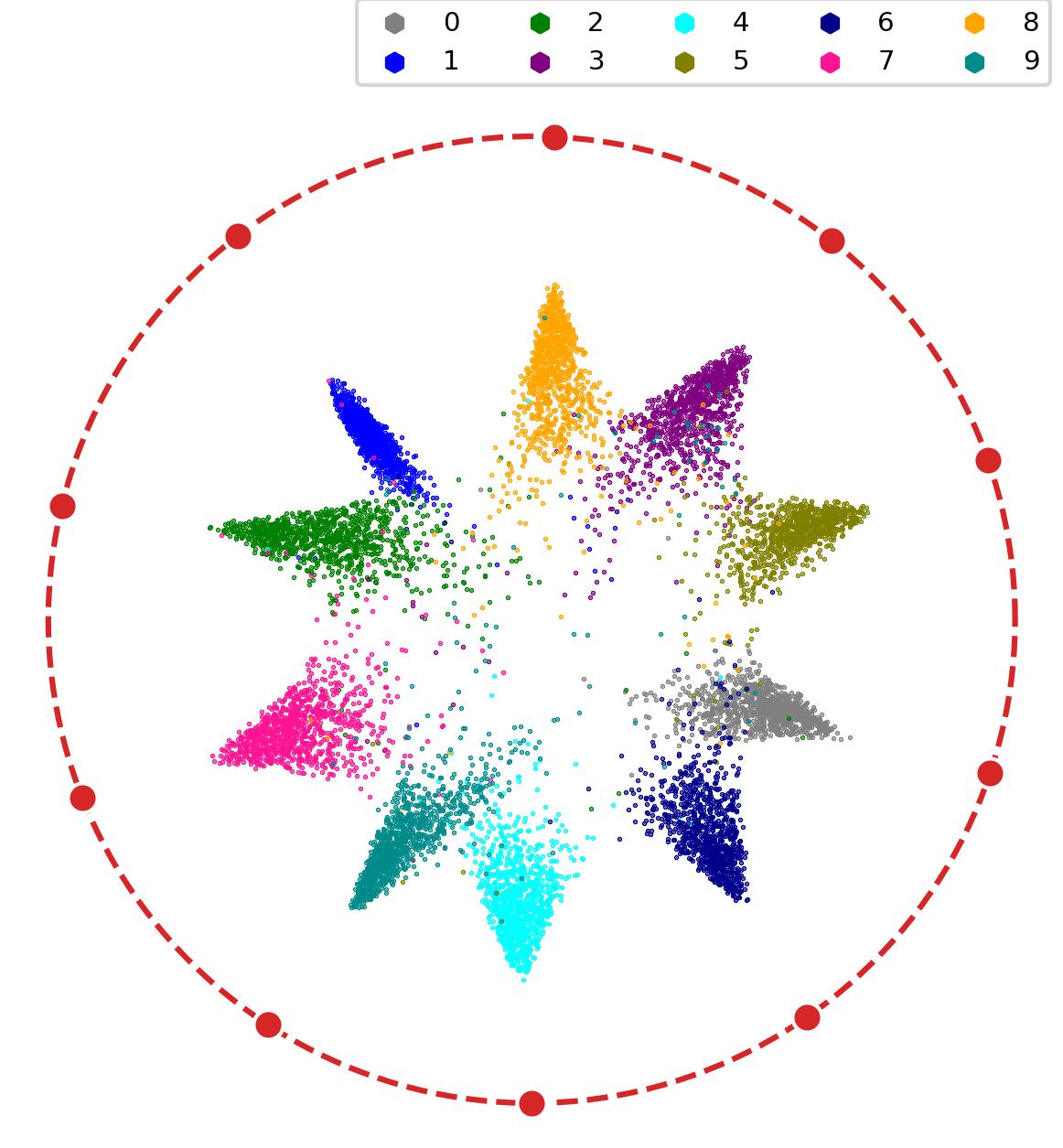}}
\subfigure[MDS]{\includegraphics[width=1in]{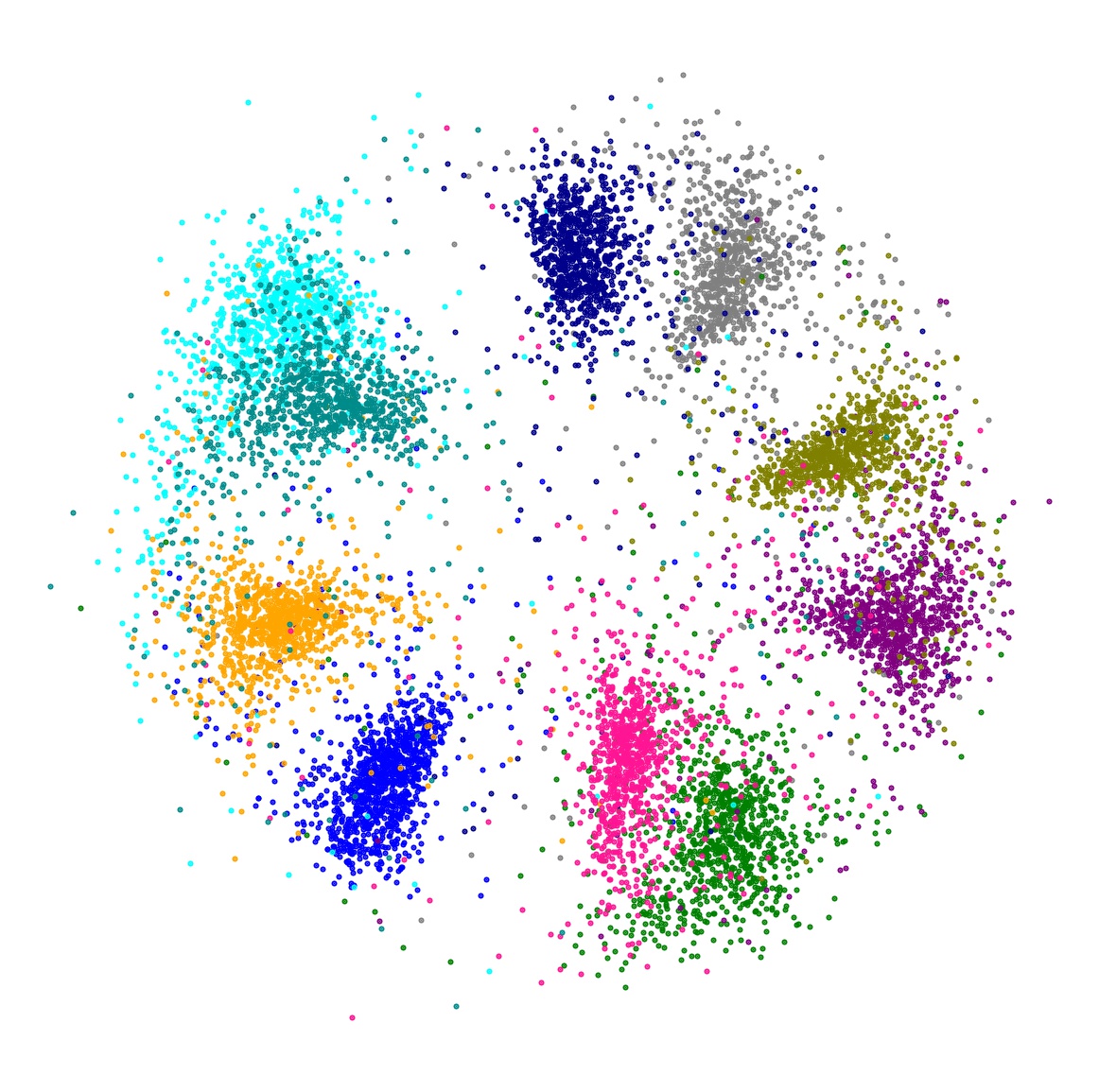}}
\subfigure[PCA]{\includegraphics[width=1in]{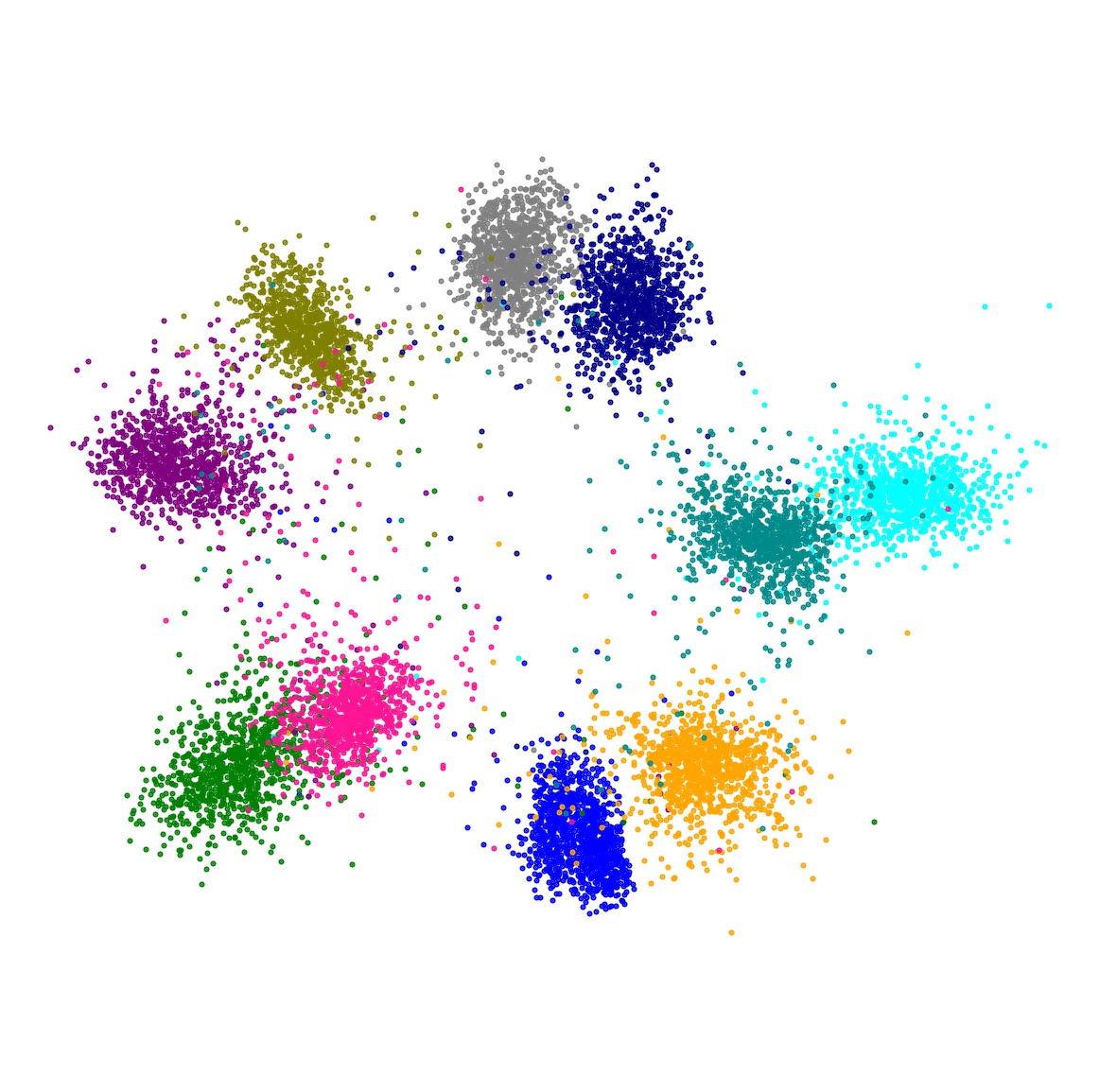}}
\subfigure[Isomap]{\includegraphics[width=1in]{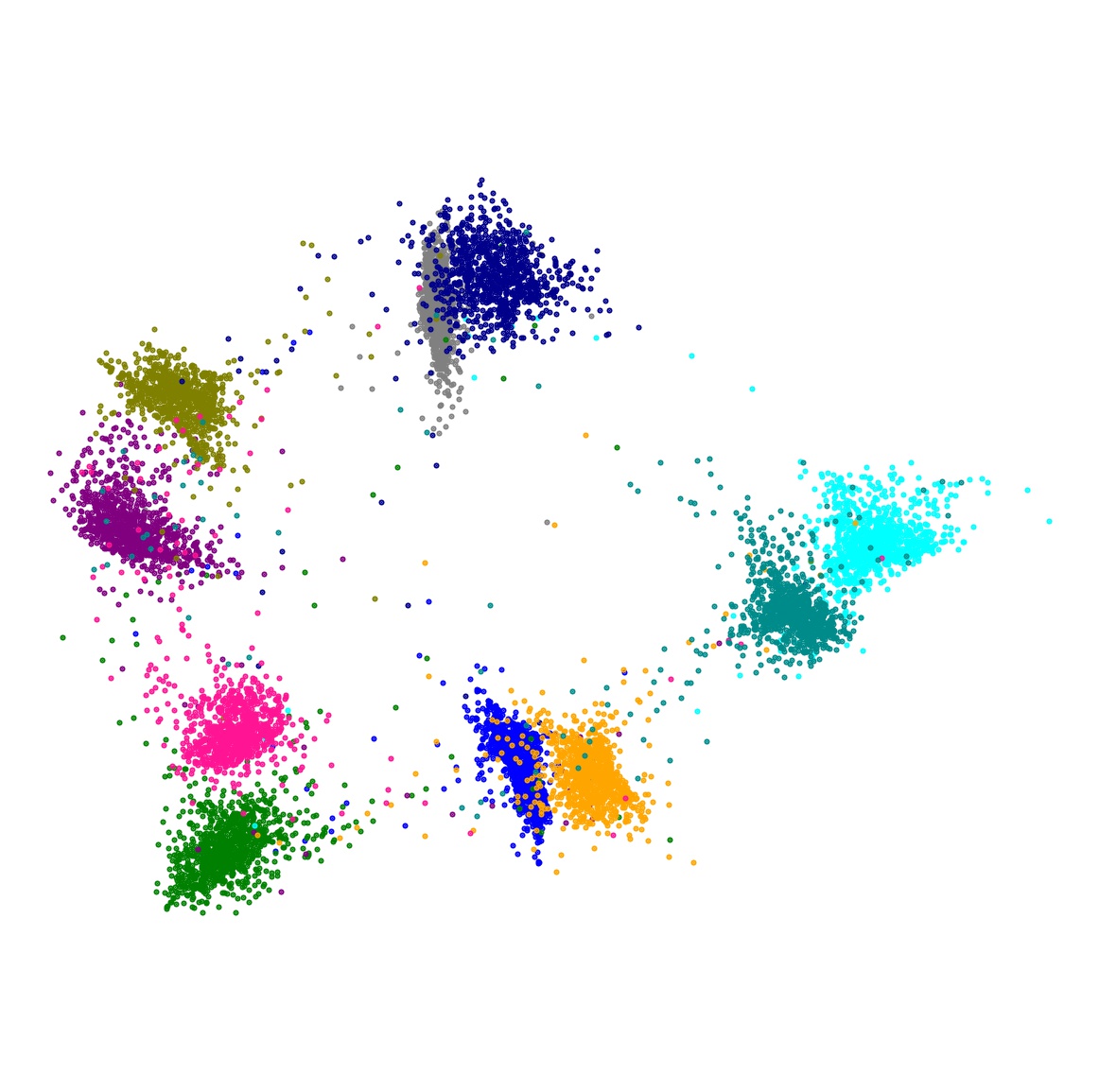}}
\subfigure[$t$-SNE]{\includegraphics[width=1in]{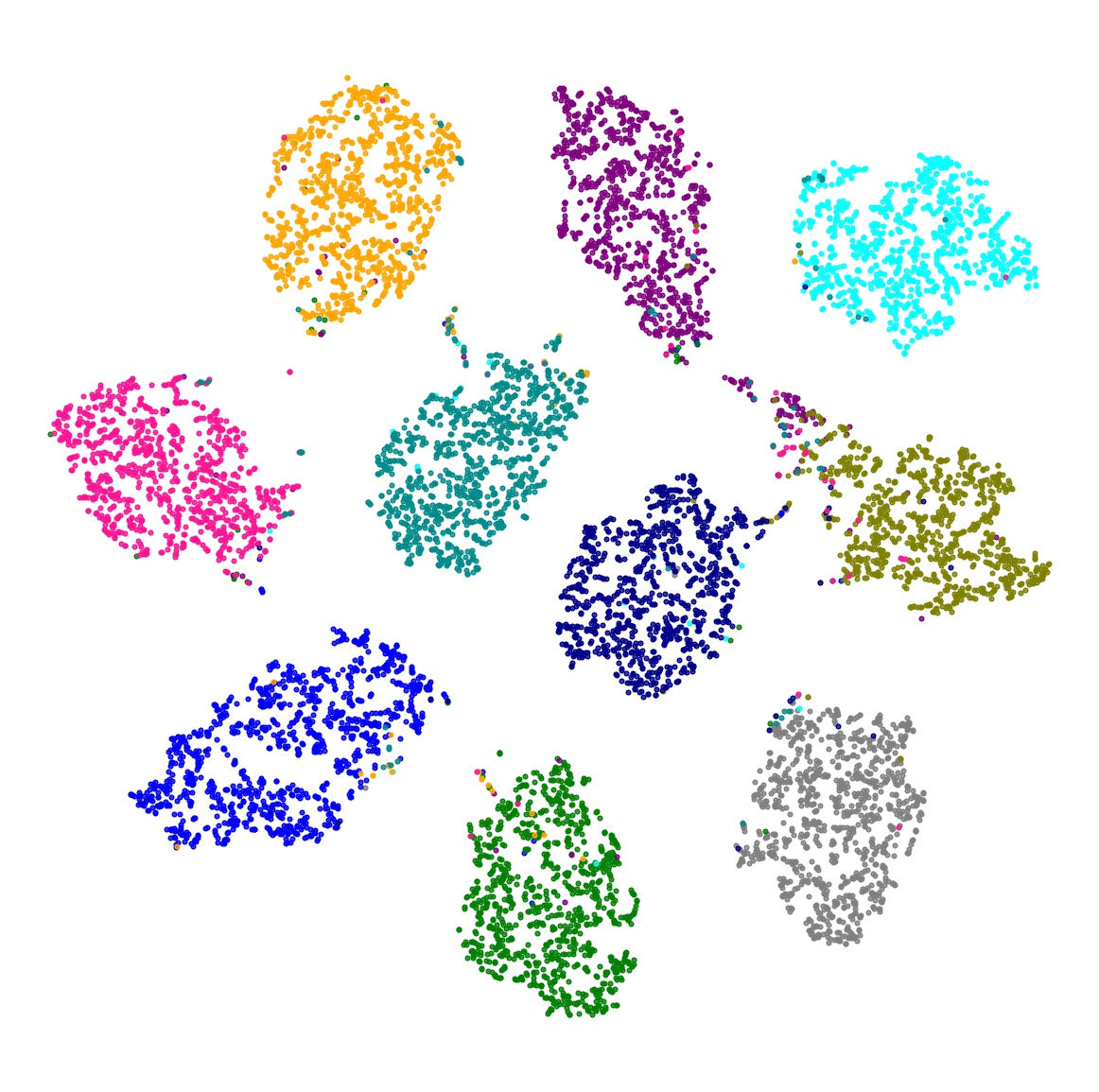}}
\subfigure[UMAP]{\includegraphics[width=1in]{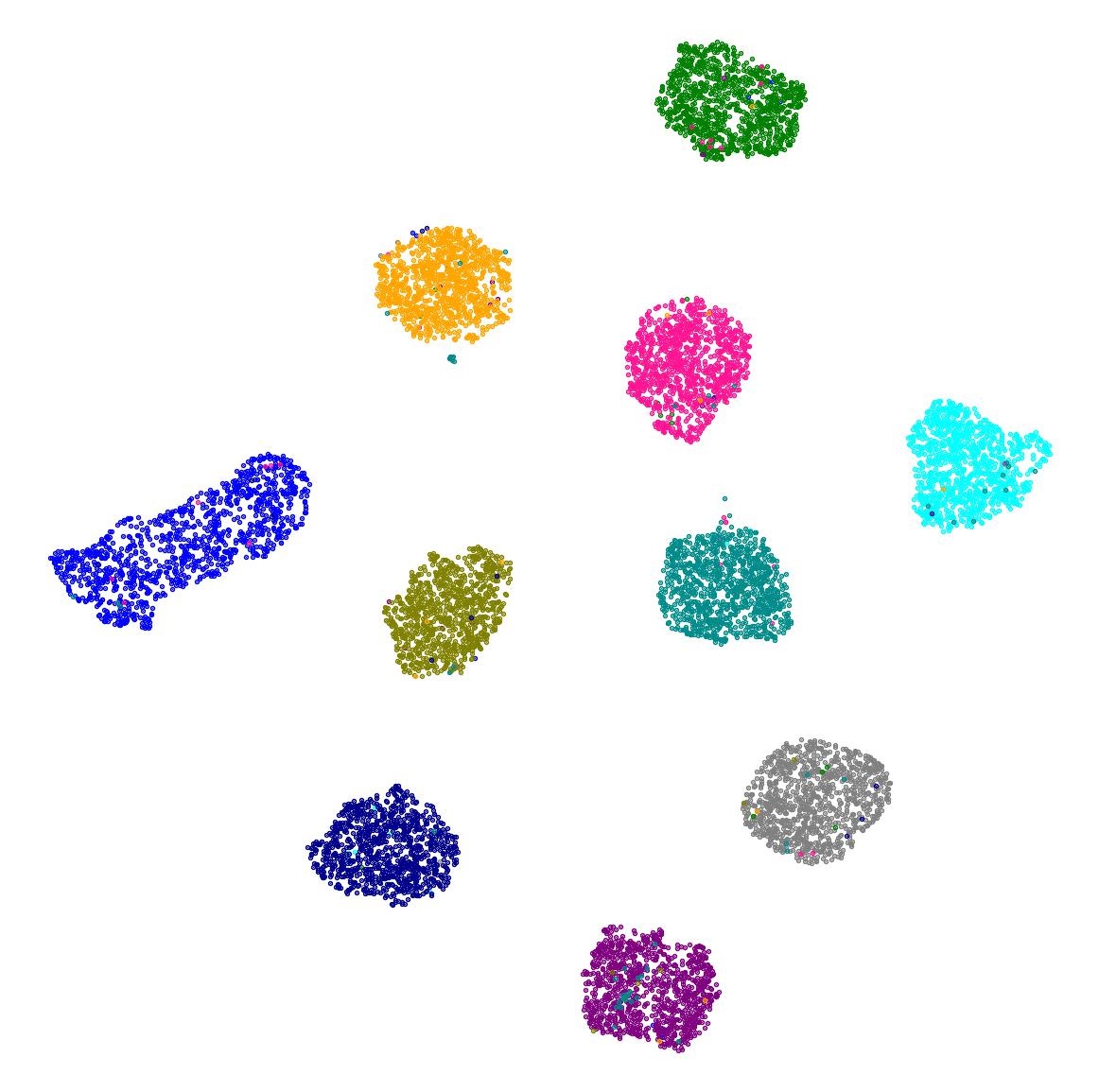}}
\caption{Different visualization methods on MNIST-test.}
\label{casestudyM}
\end{figure}

\begin{figure}[h]
\center
\subfigure[HCHC]{\includegraphics[width=1in]{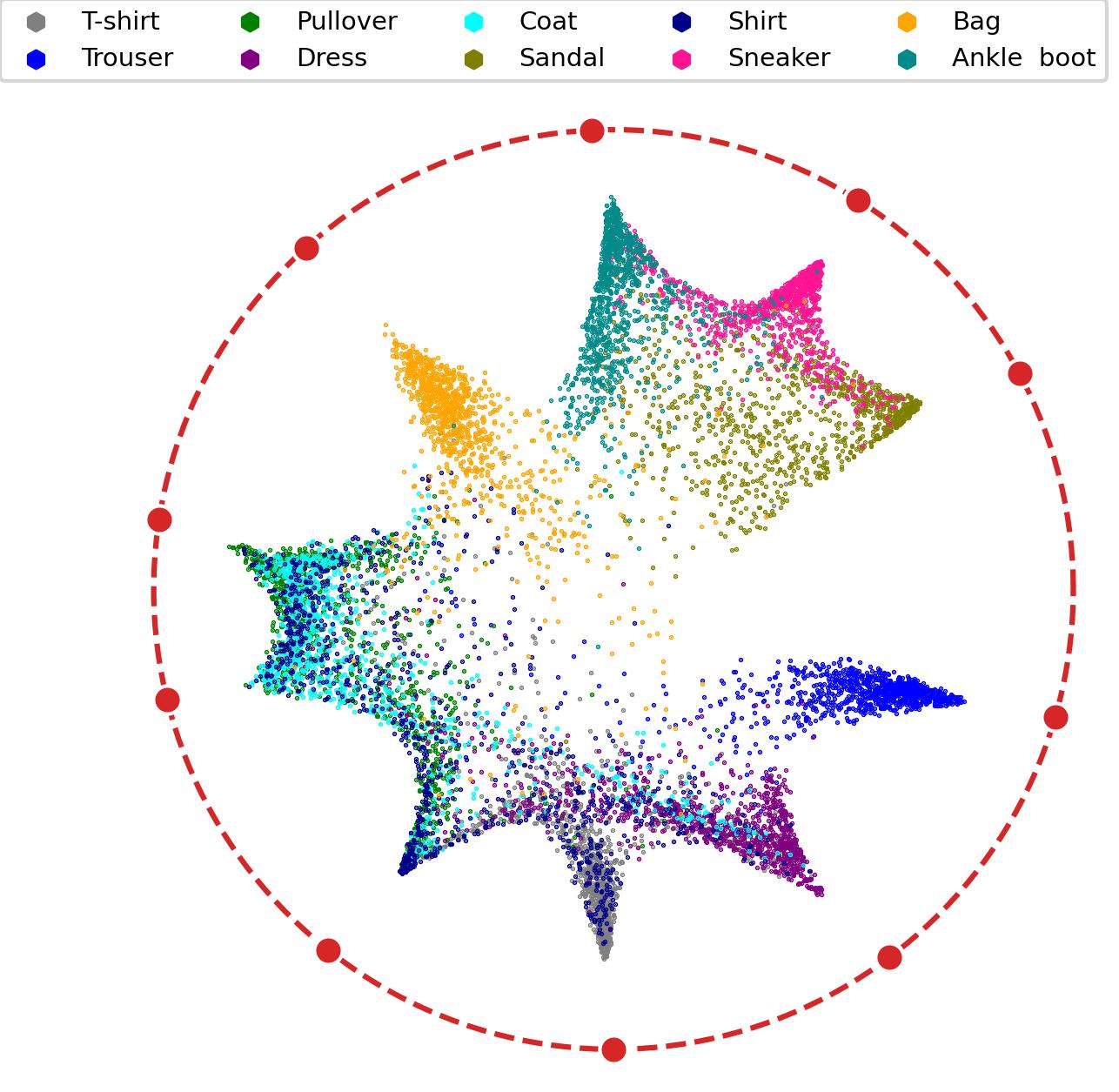}}
\subfigure[MDS]{\includegraphics[width=1in]{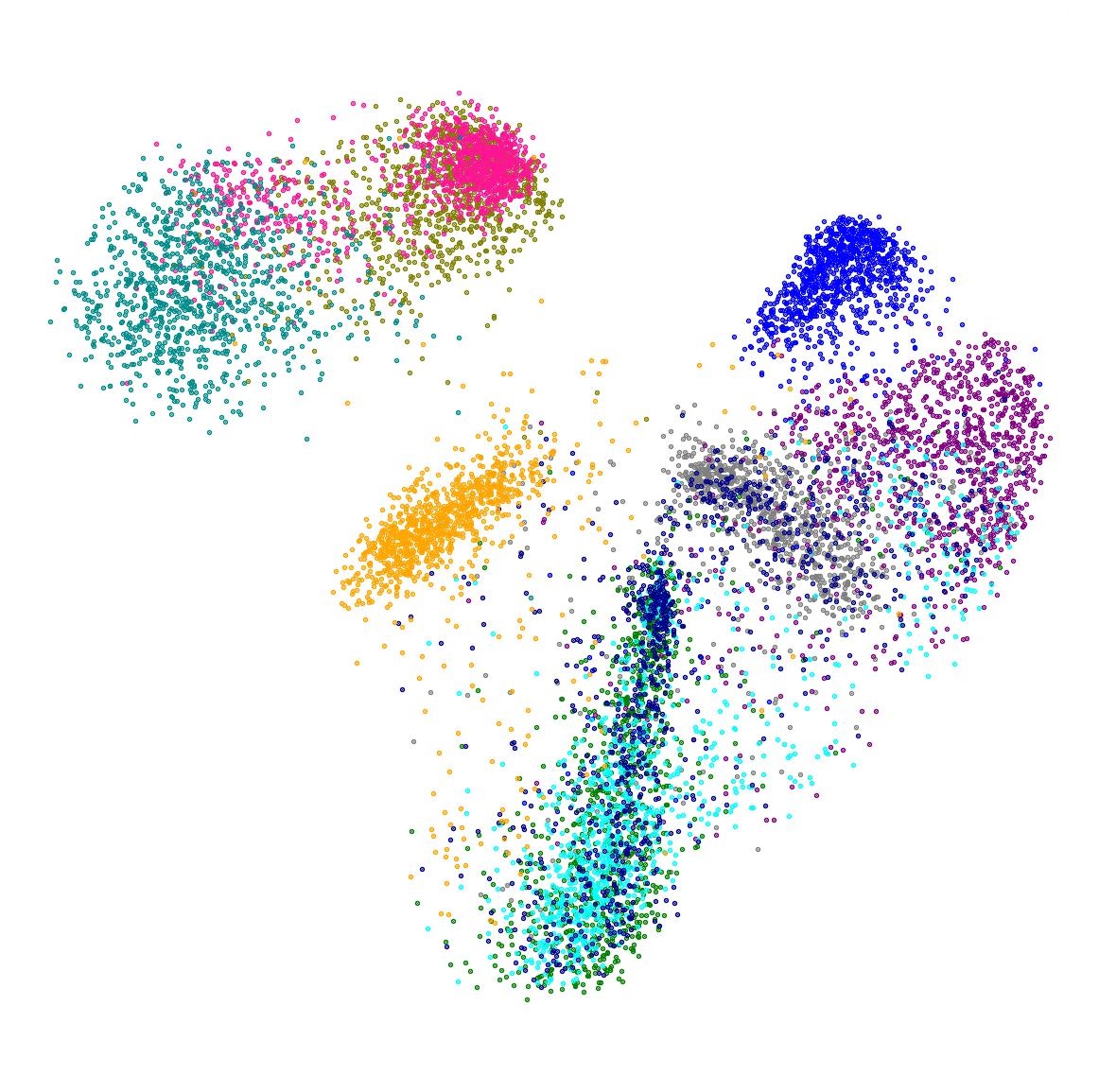}}
\subfigure[PCA]{\includegraphics[width=1in]{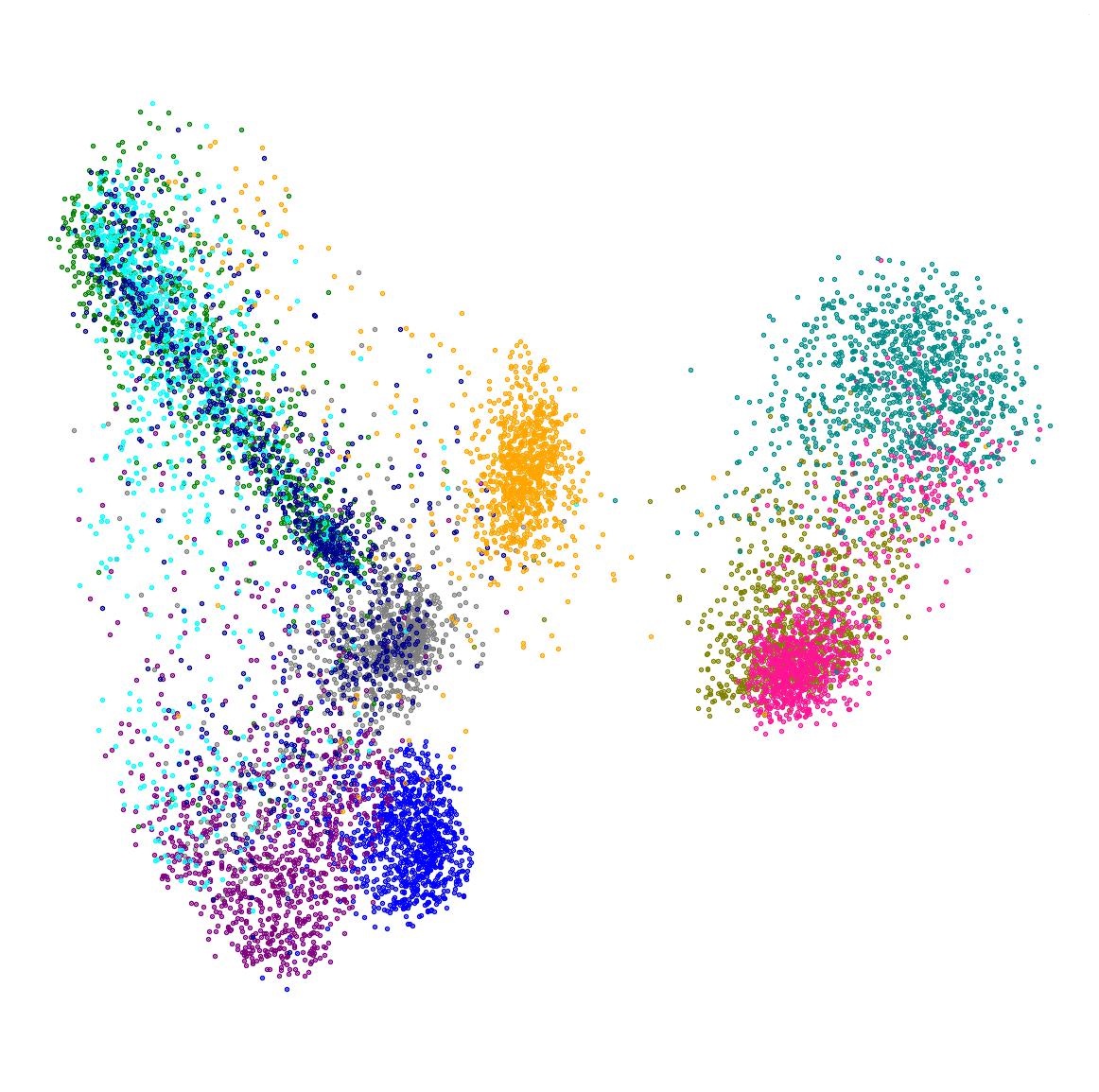}}
\subfigure[Isomap]{\includegraphics[width=1in]{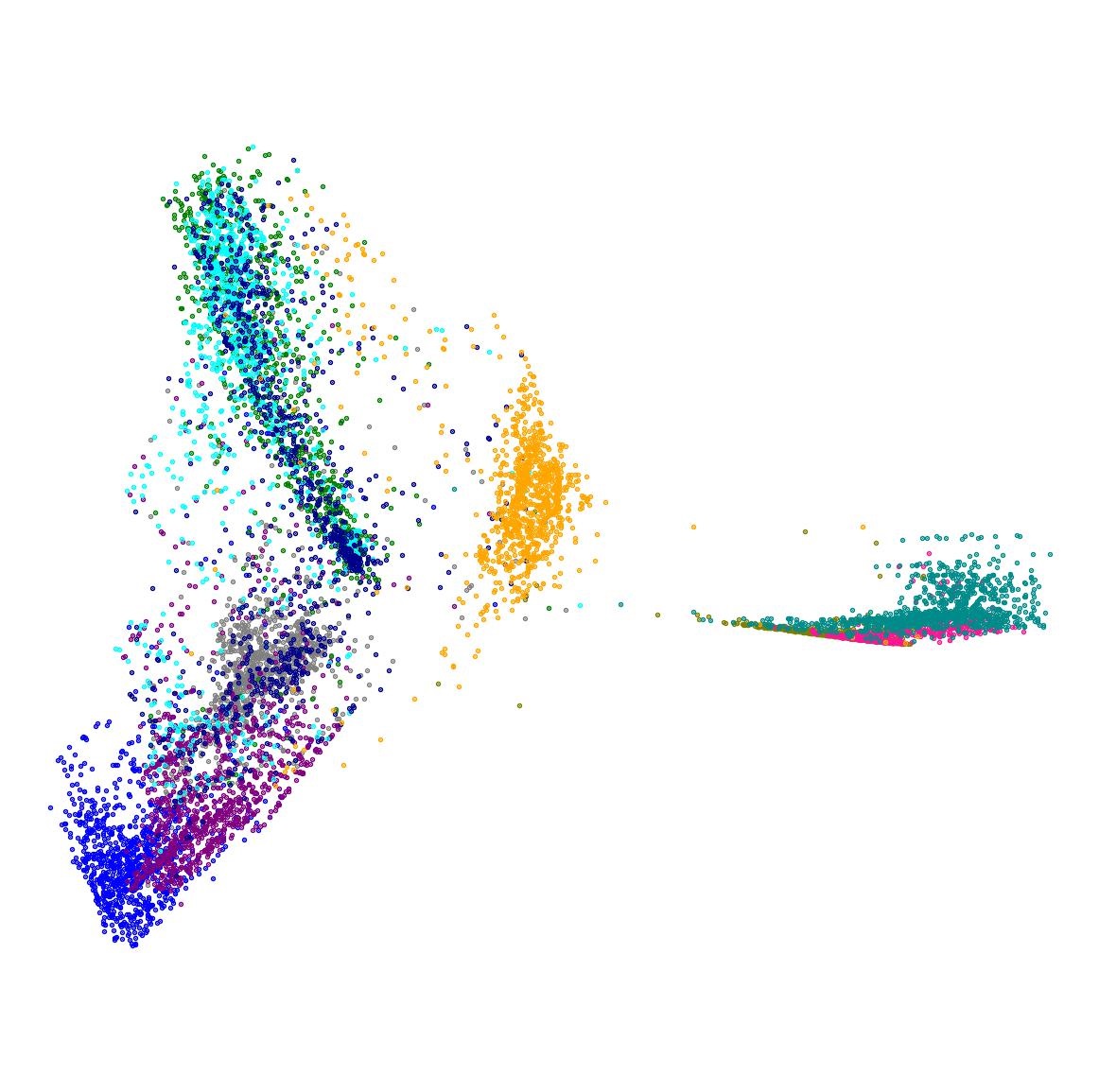}}
\subfigure[$t$-SNE]{\includegraphics[width=1in]{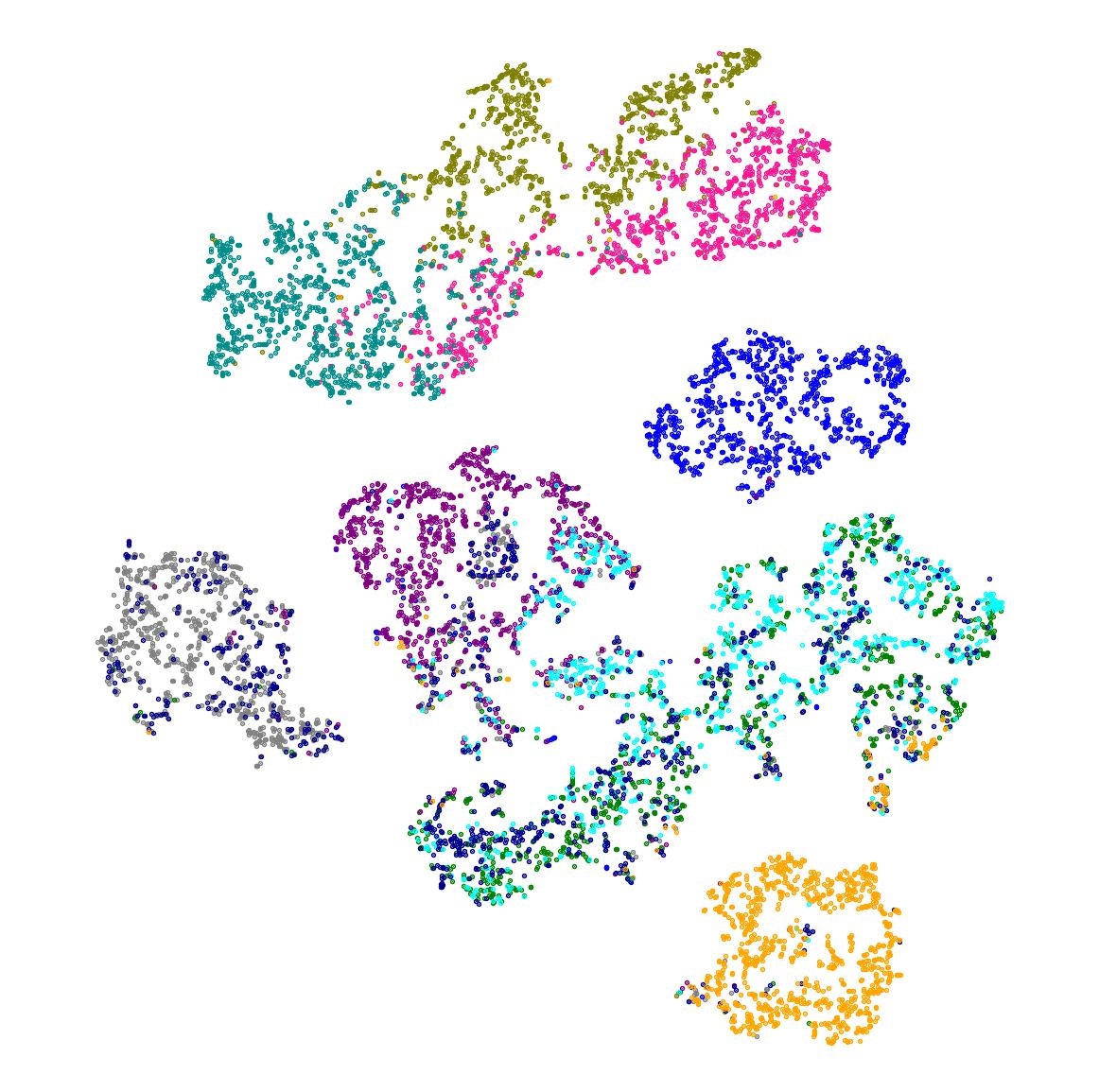}}
\subfigure[UMAP]{\includegraphics[width=1in]{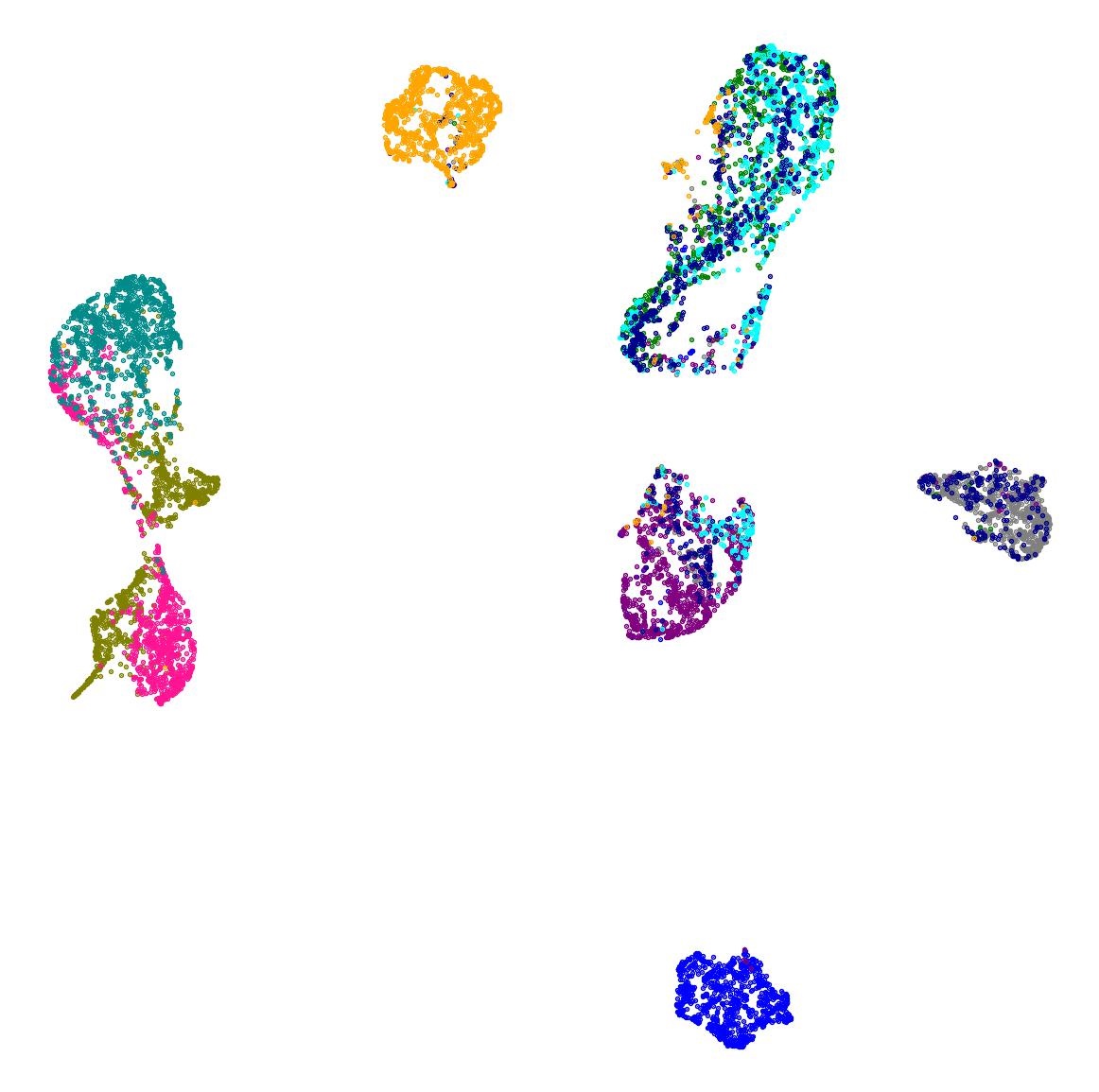}}

\caption{Different visualization methods on Fashion-test.}
\label{casestudyF}
\end{figure}
We compare the different visualization methods on our embedded features of MNIST and Fashion, respectively.
The results are shown in Fig.~\ref{casestudyM} and \ref{casestudyF}.
For MNIST, our HCHC can better visualize the cluster number and index cluster result than MDS, PCA, and Isomap.
Compared with $t$-SNE and UMAP, HCHC can better present the cluster similarities and outliers.
For Fashion, with the help of the anchors, our HCHC can better visualize the number of the clusters and index cluster result than any other visualization methods.
HCHC also can better present the cluster similarities and outliers than $t$-SNE and UMAP for Fashion.
The visualized results of the other datasets are shown in Appendix~\ref{SuppF3}.

\subsection{The Comparison of Different Clustering Methods}

In this subsection, we compared our GLDC with existing clustering methods in both the numerical and visualization cases.
We use ACC and NMI to measure the clustering performance of GLDC with nine existing clustering methods~\cite{kuhn1955hungarian}.
These methods are $k$-means~\cite{macqueen1967some}, GMM~\cite{rasmussen1999infinite}, SC~\cite{shi2000normalized}, DEC~\cite{xie2016unsupervised}, IDEC~\cite{guo2017improved}, DSC~\cite{shaham2018spectralnet}, JULE~\cite{yang2016joint}, DSCDAN~\cite{yang2019deep}, and N2D~\cite{mcconville2021n2d}.
The ACCs and NMIs of different clustering methods are shown in Table~\ref{Tab03}.
As we can see, our GLDC has the best ACCs in clustering tasks with MNIST, Fashion, Reuters-10k, HHAR, and BH.
Our GLDC also has the best NMIs with Fashion, Reuters-10k, and HHAR.
Moreover, we perform the ablation study with $L_a$ and $L_r$ in our overall clustering loss.
As we can see in Table~\ref{Tab03}, both $L_a$ and $L_r$ can well improve the clustering performance of GLDC.

\begin{figure}[h]
\centering
  \includegraphics[width=3.3in]{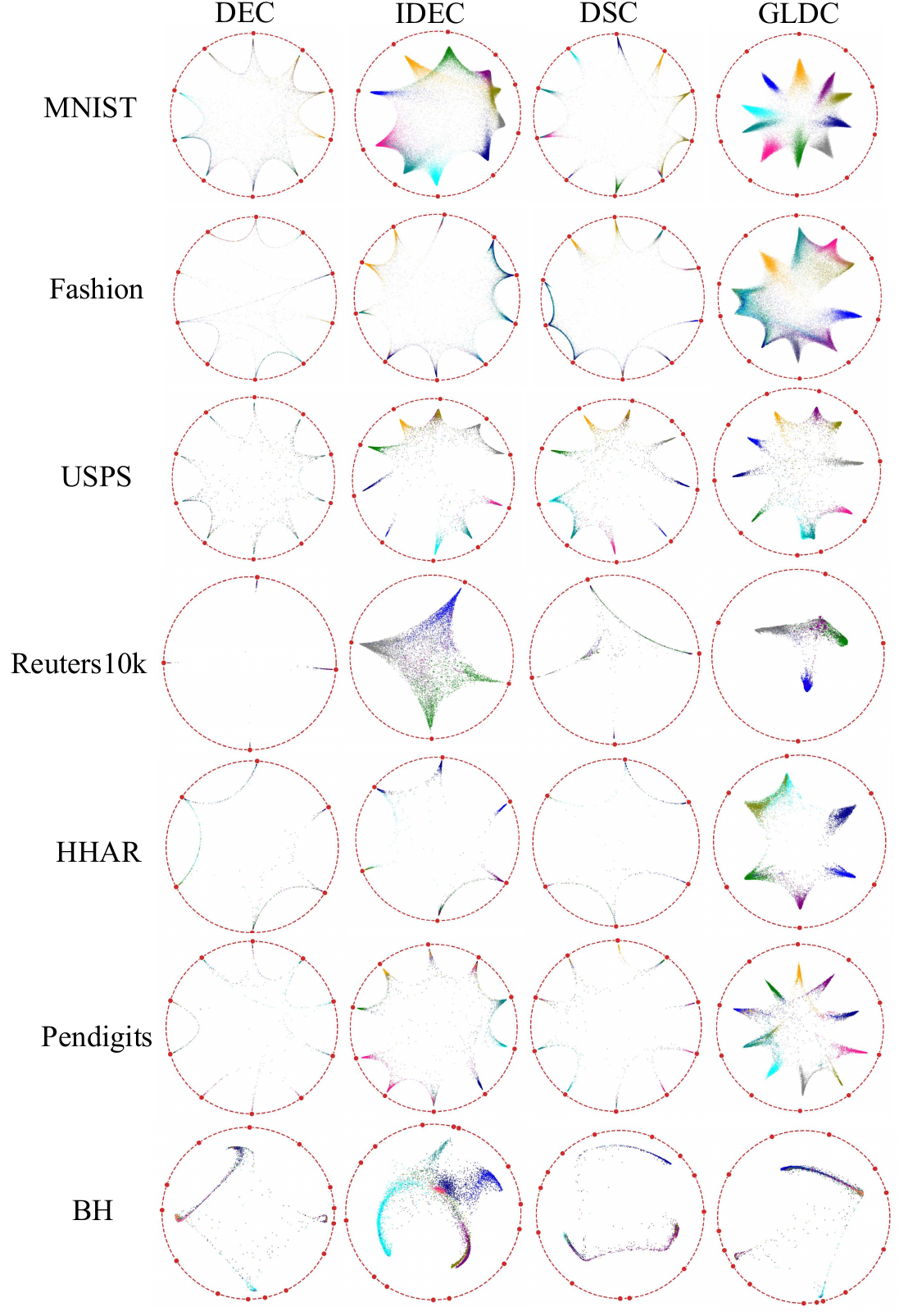}\\
  \caption{Different deep clustering methods on optimal Hamiltonian cycle.}
  \label{vccResult}
\end{figure}

In HCHC, a high clustering performance in ACC or MNI is not enough for satisfactory visualization.
The similarities between the clusters and outliers also need to be presented.
Compared with the existing deep clustering methods, our GLDC can better analyze the similarities between the different clusters by considering the global-structure in data and thus get a better visualized result.
We compare GLDC with three universal deep clustering methods including DEC, IDEC, and DSC in the framework of HCHC.
The results are shown in Fig.~\ref{vccResult}.

As we can see, it is hard to see the similarities between the different clusters by DEC, IDEC, and DSC,
However, our GLDC can well show the similarities between different clusters by considering the global-structure of data in the generated distributions.
For example, by GLDC in HCHC we can find that
(1) in MNIST the blue class is not similar to other classes and thus can be well discovered in the clustering task.
(2) in Reuters-10k, the purple class is similar to the grey class and the green class,
thus these three classes are easily mixed in the clustering task.
To this end, we need to analyze the possibilities of a non-outlier sample belonging to different clusters.
Thus even if a non-outlier sample $\bm{x_i}$ does not belong to cluster $j$, $p_{i,j}$ may not tend to $0$.
In this case, the max value in $\bm{p_i}$ may not tend to $1$ with $\sum_{j=1}^{c}p_{i,j}=1$.
Therefore, the mapped samples may not be very close to the anchor of the corresponding cluster on the red circle.

We also can see the advantages of GLDC in visualized comparisons in Fig.~\ref{vccResult}.
Concretely, we have the following observations.
(1) For the MNIST dataset, DEC can not well recognize the dark cyan class and orange class,
 IDEC can not well recognize the dark cyan class and olive class, DSC cannot cluster the green class very well, however, GLDC can well recognize and cluster all of the classes.
(2) GLDC can well recognize most classes in Fashion, but DEC, IDEC, and DSC cannot do so.
(3) For the USPS dataset, DEC divides the dark cyan class into all the clusters, IDEC mistakenly clusters the purple class and olive class together and cannot well recognize the grey class, DSC also mistakenly cluster the purple class and olive class together, however, GLDC can well recognize and cluster all of the classes.
(4) For the Reuters10k dataset, DEC cannot well recognize most classes, IDEC mistakenly clusters the purple class and blue class together and divide the samples in the green class into two clusters, DSC mistakenly mixes the samples in the green class and purple class, however, GLDC can well recognize most of the classes, i.e., the green class, blue class, and grey class.
(5) For the HHAR dataset, DEC cannot well recognize most classes, IDEC mistakenly clusters the cyan class and olive class together and divides the samples in the dark blue class into two clusters, DSC mistakenly divides the samples in the dark blue class into two clusters, however, GLDC can well recognize most of the classes.
(6) For the Pendigits dataset, DEC, IDEC, and DSC cannot recognize the pink class and dark cyan class, however, GLDC can well recognize these two classes.
(7) For the BH dataset, DEC and GLDC can better cluster the samples in the cyan and purple class than IDEC and DSC.

\subsection{The Result on the COVID-19 Dataset}
The above experiment illustrates the effectiveness of HCHC to discover the knowledge, i.e., clusters, similarities, and outliers, in real-world datasets. 
Then, in this subsection, we use HCHC to learn the temporal features of COVID-19 by a dataset that includes the available COVID-19 daily information of the different states in the USA from 21/1/2020 to 21/1/2022.
The daily information includes daily cases, daily deaths, and so on.
There are 37525 samples in this dataset and their features are detailed in Appendix~\ref{SuppE}.

In Fig.~\ref{cUSA} (a), where the data are labelled by every $6$ months, the samples from the same period can be mapped close to the same cluster anchor.
The results of the data labelled by every $4$ month, every $5$ month, and every $8$ month, are shown in Appendix~\ref{SuppF1}.
The total daily cases in the USA from 21/1/2020 to 21/1/2022 are shown in Fig.~\ref{cUSA} (b).
Combining Fig.~\ref{cUSA} (a) and (b), we can see that after 2020.8 the period of COVID-19 can be around $6$ months, i.e., the number of daily cases increased from autumn to winter and decreased from spring to summer.
This phenomenon may illustrate that there are more people infected with COVID-19 in the low-temperature environment than in the high-temperature environment.
\begin{figure}[]
\center
\subfigure[Labeled by every $6$ months]{\includegraphics[width=1.6in]{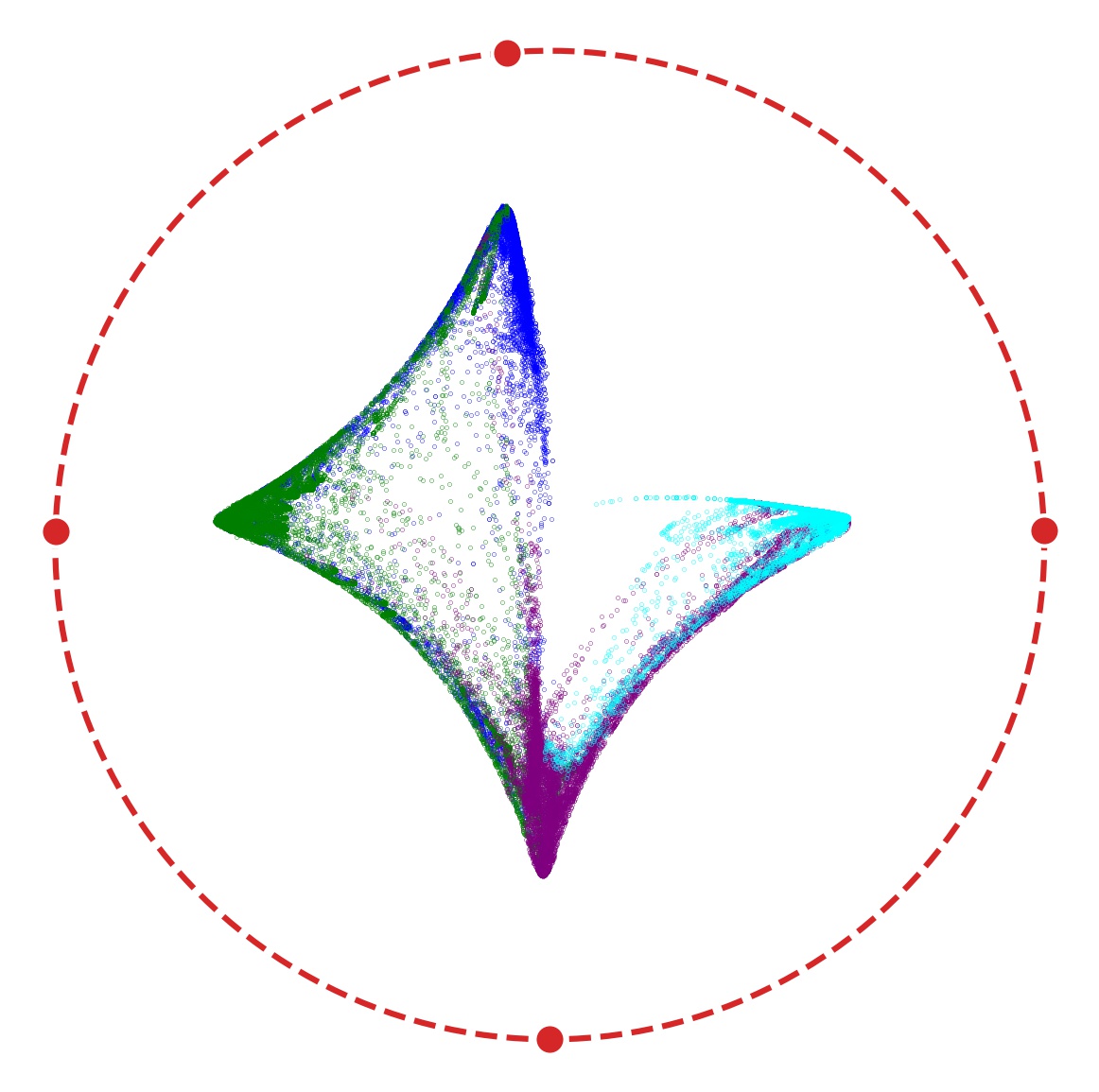}}
\subfigure[Daily case]{\includegraphics[width=1.6in]{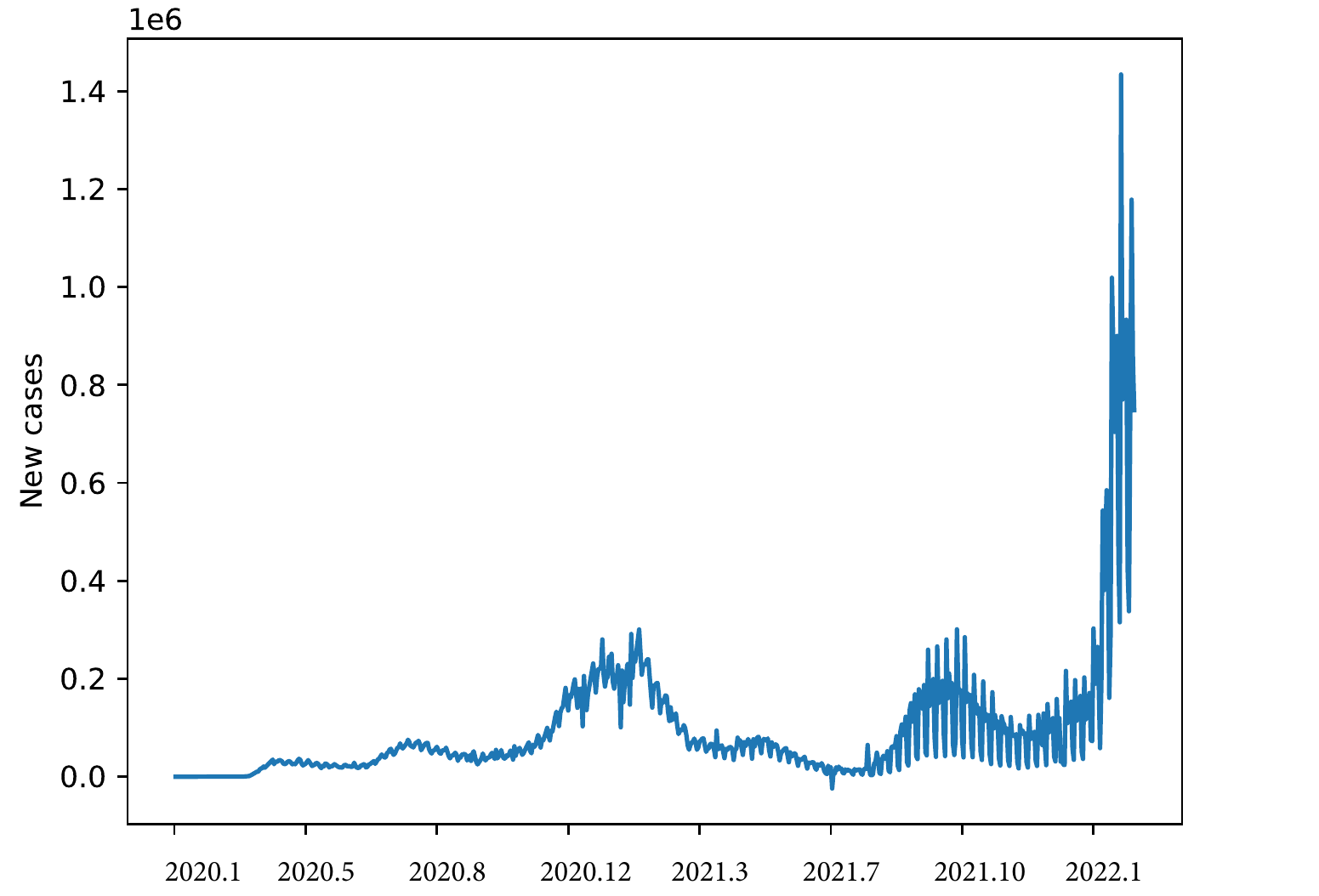}}
\caption{The results of HCHC on the COVID-19 dataset.}
\label{cUSA}
\end{figure}

\section{Conclusion}
This paper proposes HCHC to cluster high-dimensional data, and then visualize the clustering result.
It combines global structure with local structure in one objective function, improving the labels as relative probabilities, to mine the similarities between different clusters while keeping the local structure in each cluster.
Then, the anchors of different clusters are sorted on the optimal Hamiltonian cycle generated by the cluster similarities and mapped on the circumference of a circle.
Finally, a sample with a higher probability of a cluster will be mapped closer to the corresponding anchor.
In this way, HCHC allows us to appreciate three aspects at the same time - (1) cluster recognition, (2) cluster similarities, and (3) outlier recognition.
The experimental result shows the effectiveness of HCHC.
As shown in our experiment, HCHC also can serve as a visualized clustering measure by mapping the labelled samples.
In this way, we can find the reasons for an unsatisfactory clustering result.

In further research, we will investigate how to use HCHC to solve the multi-cluster problem.
Because we rely on an NP-hard optimization step, the optimal Hamiltonian cycle to sort the clusters, it is necessary to give an approximate solution in acceptable time consumption as shown in Appendix~\ref{SuppD1}.
Mapping too many clusters in $2$D space may cause high mapping error.
Therefore we can investigate how to map these clusters in $3$D space.

\section*{Acknowledgements}
The authors thank partial supports from NSF-DMS-2012298, NNSFC-71991471, 11931015, 71974176.

\nocite{langley00}

\bibliography{References}
\bibliographystyle{icml2023}

%%%%%%%%%%%%%%%%%%%%%%%%%%%%%%%%%%%%%%%%%%%%%%%%%%%%%%%%%%%%%%%%%%%%%%%%%%%%%%%
%%%%%%%%%%%%%%%%%%%%%%%%%%%%%%%%%%%%%%%%%%%%%%%%%%%%%%%%%%%%%%%%%%%%%%%%%%%%%%%
% APPENDIX
%%%%%%%%%%%%%%%%%%%%%%%%%%%%%%%%%%%%%%%%%%%%%%%%%%%%%%%%%%%%%%%%%%%%%%%%%%%%%%%
%%%%%%%%%%%%%%%%%%%%%%%%%%%%%%%%%%%%%%%%%%%%%%%%%%%%%%%%%%%%%%%%%%%%%%%%%%%%%%%
\newpage
\appendix
\onecolumn

%\title{The Appendix of High-dimensional Clustering onto Hamiltonian Cycle}
%\maketitle

\section{An example of HCHC}
\label{SuppA}

\begin{figure}[h]
\center
  \includegraphics[width=4in]{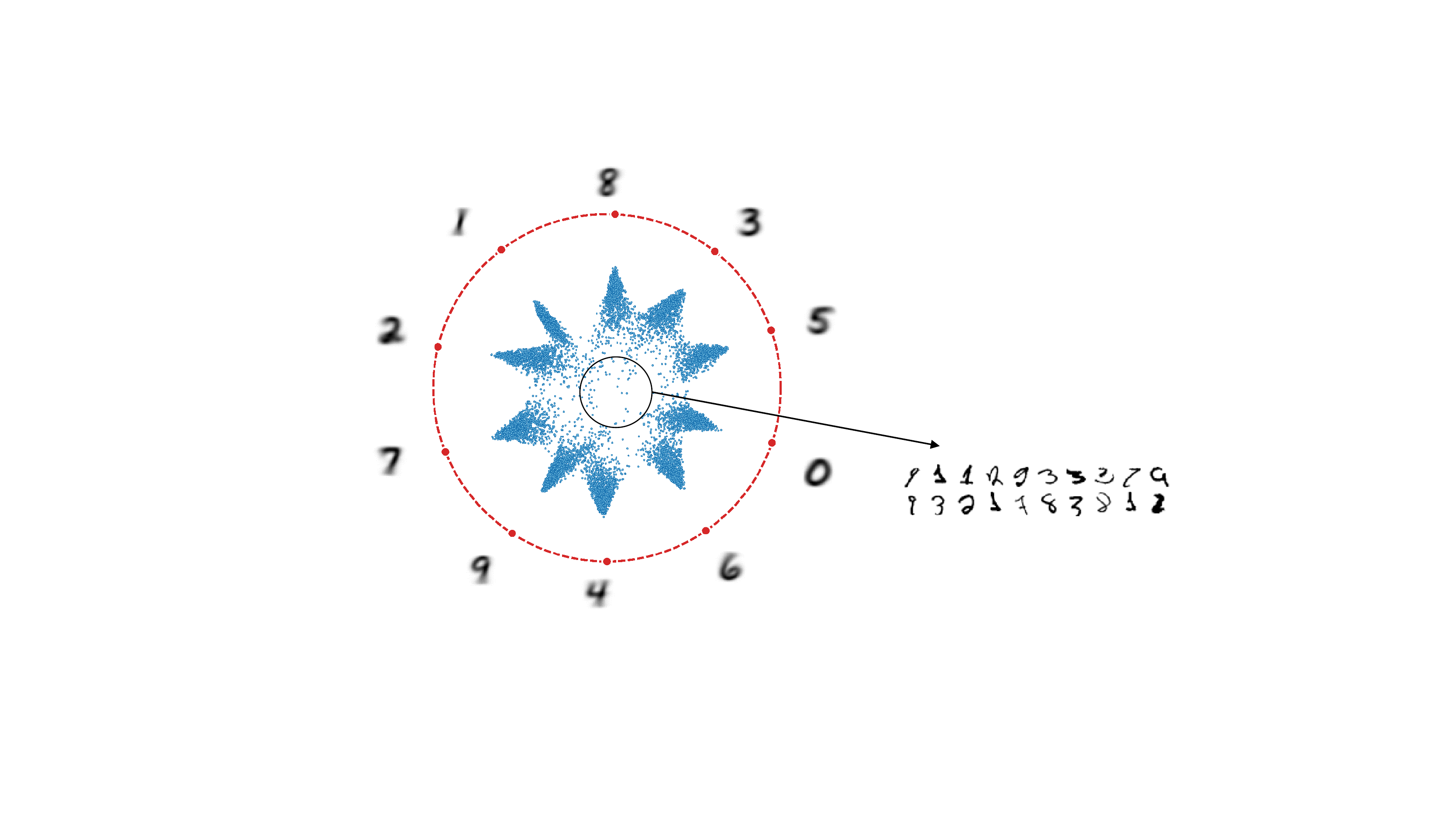}\\
  \caption{An example on MNIST-test}
  \label{exp1}
\end{figure}

Fig.1 is an example to show the visualized result on a handwriting digit number dataset, MNIST-test, by our HCHC.
In this figure, the layout of different clusters is on the circumference of a red circle by the optimal Hamiltonian cycle.
Each red dot on the circumference is the anchor of a cluster, and the blue dots in the circle are mapped samples.
If a mapped sample is close to a red dot, this sample will have a high probability of belonging to the corresponding cluster.
The pictures around the red circle are the means of the different clusters.
As we can see, HCHC well mines and represents the local similarities between the samples and the global similarities between the different clusters.
Concretely, the samples of each class in the MNIST-test can be mapped together, and thus the mean of each cluster can be a corresponding digit number.
The more similar clusters can be mapped closer to each other, e.g., the shapes of digit numbers “9” and “4” are similar, so their clusters are mapped close to each other.
Any sample in the centre of the circle will have a low possibility to belong to any cluster.
Such samples are usually not well-written and can be regarded as outliers.

\section{The comparison between HCHC and Radviz Deluxe}
\label{SuppA1}

To clarify the difference between our new HCHC and the existing Radviz Deluxe, we offer a direct comparison in the following table.
\begin{table}[ht]
\centering
\caption{The Comparison between HCHC and Radviz Deluxe}
\begin{tabular*}{16.5cm}{@{\extracolsep{\fill}}ccccccc}	
\toprule
Method           &  Hamiltonian cycle     &  Deep clustering &  Anchor  &  Visualization  &  Outlier mapping\\
\midrule
  Radviz Deluxe  & Yes                    & No               & Data features                 & Feature values      & Apart from clusters         \\
 HCHC            & Yes                    & Yes              & Cluster labels                &  Clustering result  & Near the centre          \\
\bottomrule
\end{tabular*}
\end{table}

Although both HCHC and Radviz Deluxe map the samples by the Hamiltonian cycle on a circle, there are still four differences between HCHC and Radviz Deluxe. 
1.	HCHC include a new deep clustering method GLDC to mine the clusters, cluster similarities, and outliers in data, however, Radviz Deluxe is only for data visualization without clustering.
2.	The Anchors in HCHC index the cluster labels, however, the anchors in Radviz Deluxe index the data features.
3.	HCHC aims to visualize the mined clustering result in GLDC; however, Radviz Deluxe aims to visualize the feature values of the samples.
4.	HCHC maps the samples by their clustering probability distributions; therefore, the outliers with low probability to any cluster can be mapped near the circle's centre. However, Radviz Deluxe maps the samples by their feature values, so the outliers may be anywhere in the circle.

\section{The Introduction of Radviz}
\label{SuppB}
Radviz defines the polar coordinate of feature $\bm{f_j}$ as 
\begin{eqnarray}
\label{xvf}
\bm{\mu_{f_j}} = [r \times \cos(\alpha_{\bm{f_j}}), r \times \sin(\alpha_{\bm{f_j}})]
\end{eqnarray}
where $\alpha_{\bm{f_j}}$ is the angle corresponding to the position of $\bm{f_j}$ and $r$ is the radius of the circle.
The the position of $\bm{x_i}$ in the circle can be computed by 
\begin{eqnarray}
\label{pvpl}
\bm{\mu_{x_i}} = \sum_{j=1}^{d}\frac{x_{i,j}}{\sum_{k=1}^{d}x_{i,k}}\bm{\mu_{f_j}}
\end{eqnarray}
where $d$ is the number of the features.
\begin{figure}[h]
\centering
  \includegraphics[width=3.5in]{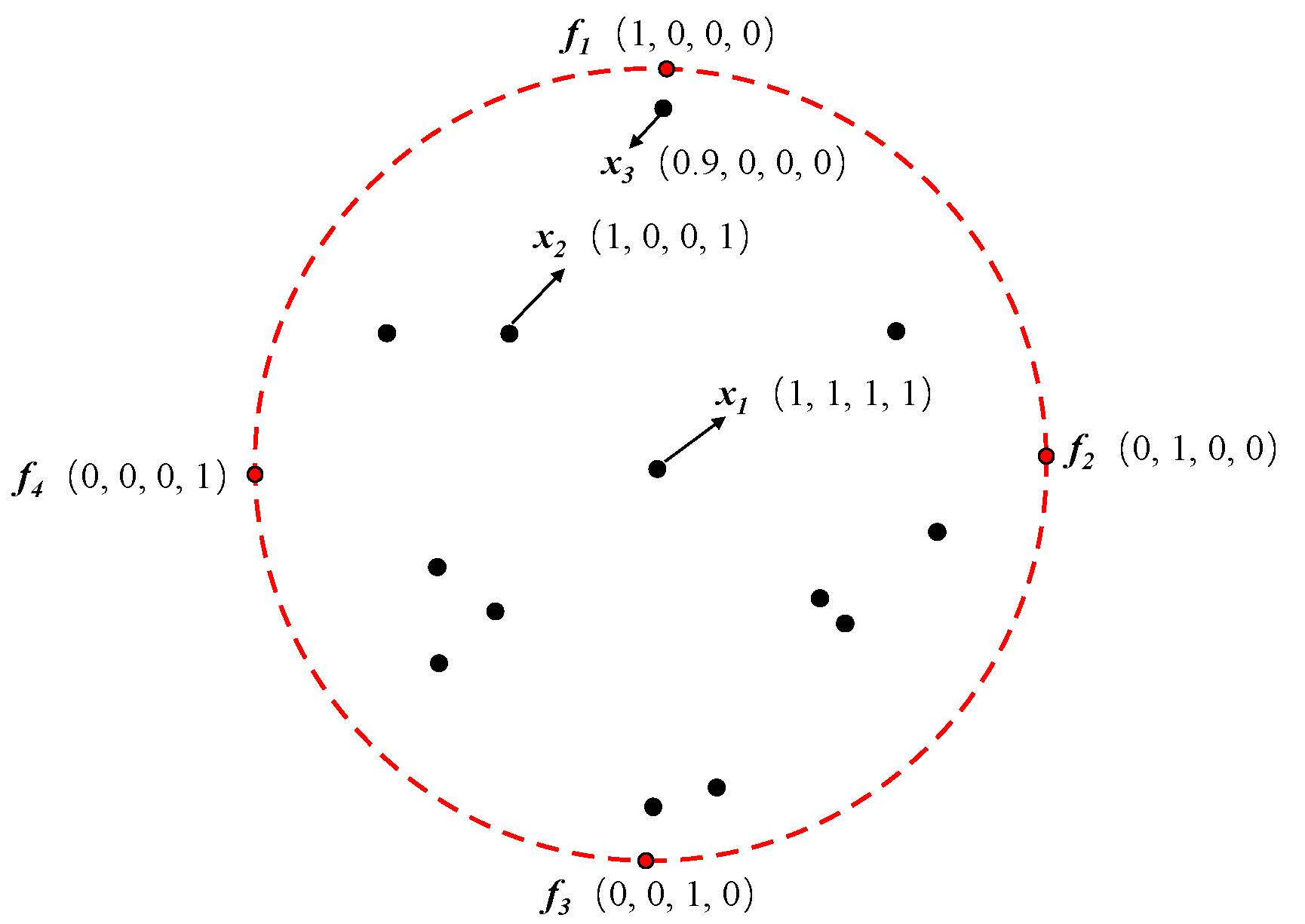}\\
  \caption{An example of RadVis visualization.}
  \label{exp3}
\end{figure}

An example of the Radviz is shown in Fig~\ref{exp3}.
The coordinates represent the values of features $\bm{f_1}$, $\bm{f_2}$, $\bm{f_3}$, and $\bm{f_4}$ for the indicated samples.
$\bm{x_1}$ is at the center of the circle and has coordinates $(1, 1, 1, 1)$.
$\bm{x_2}$ has coordinates $(1, 0, 0, 1)$.
Therefore, the distance between $\bm{x_2}$ and $\bm{f_1}$ is the same as the distance between $\bm{x_2}$ and $\bm{f_4}$.
$\bm{x_3}$ has coordinates $(0.9, 0, 0, 0)$. 
It is located near to anchor the of $\bm{f_1}$.

\section{The Motivation of Our Work}
\label{SuppC}
\subsection{The Limitations in the Presentation of High-dimensional Clustering}
\label{SuppC11}
\begin{figure}[h]
\center
\subfigure[MDS]{\includegraphics[width=1.3in]{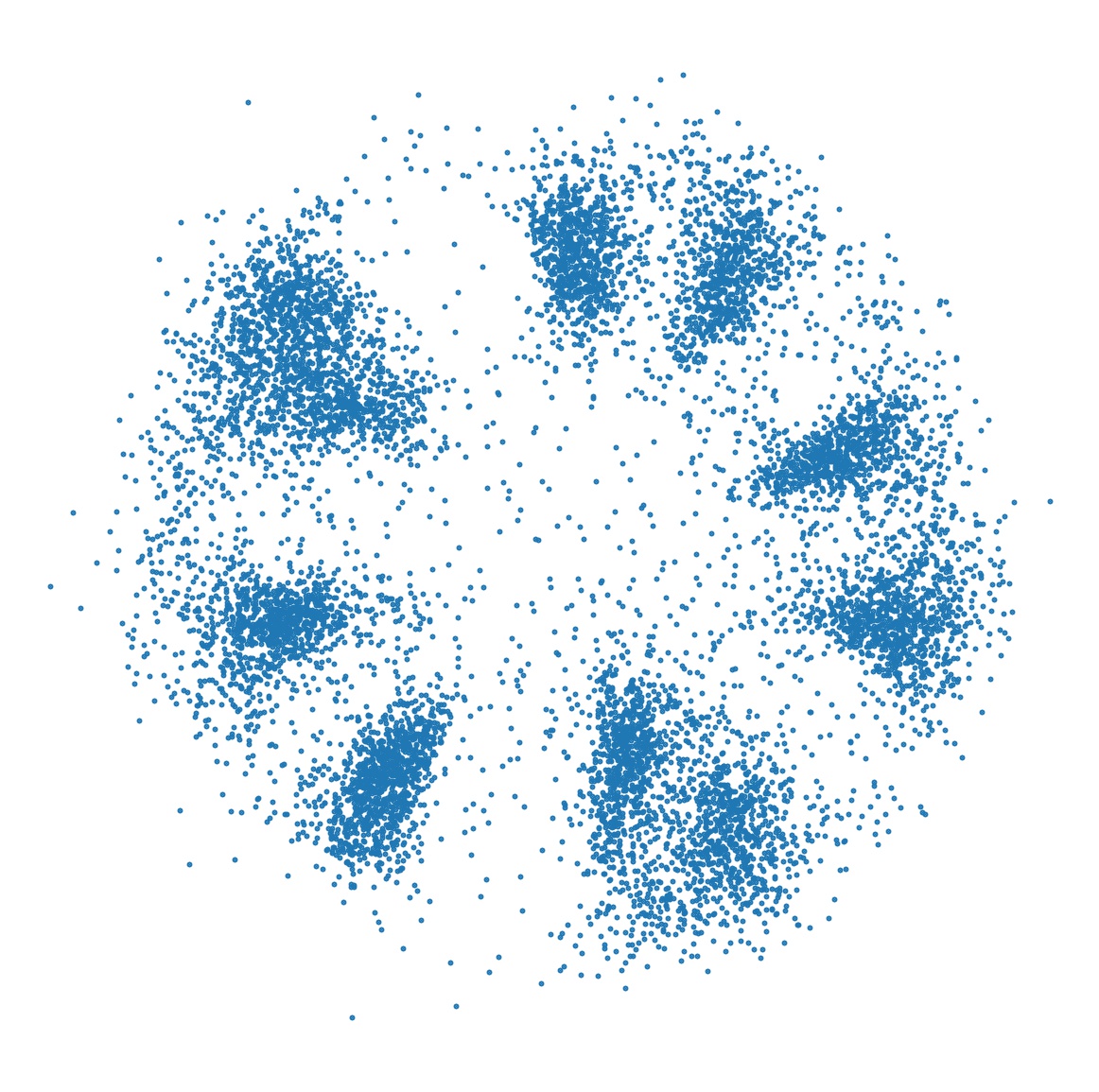}}
\subfigure[PCA]{\includegraphics[width=1.3in]{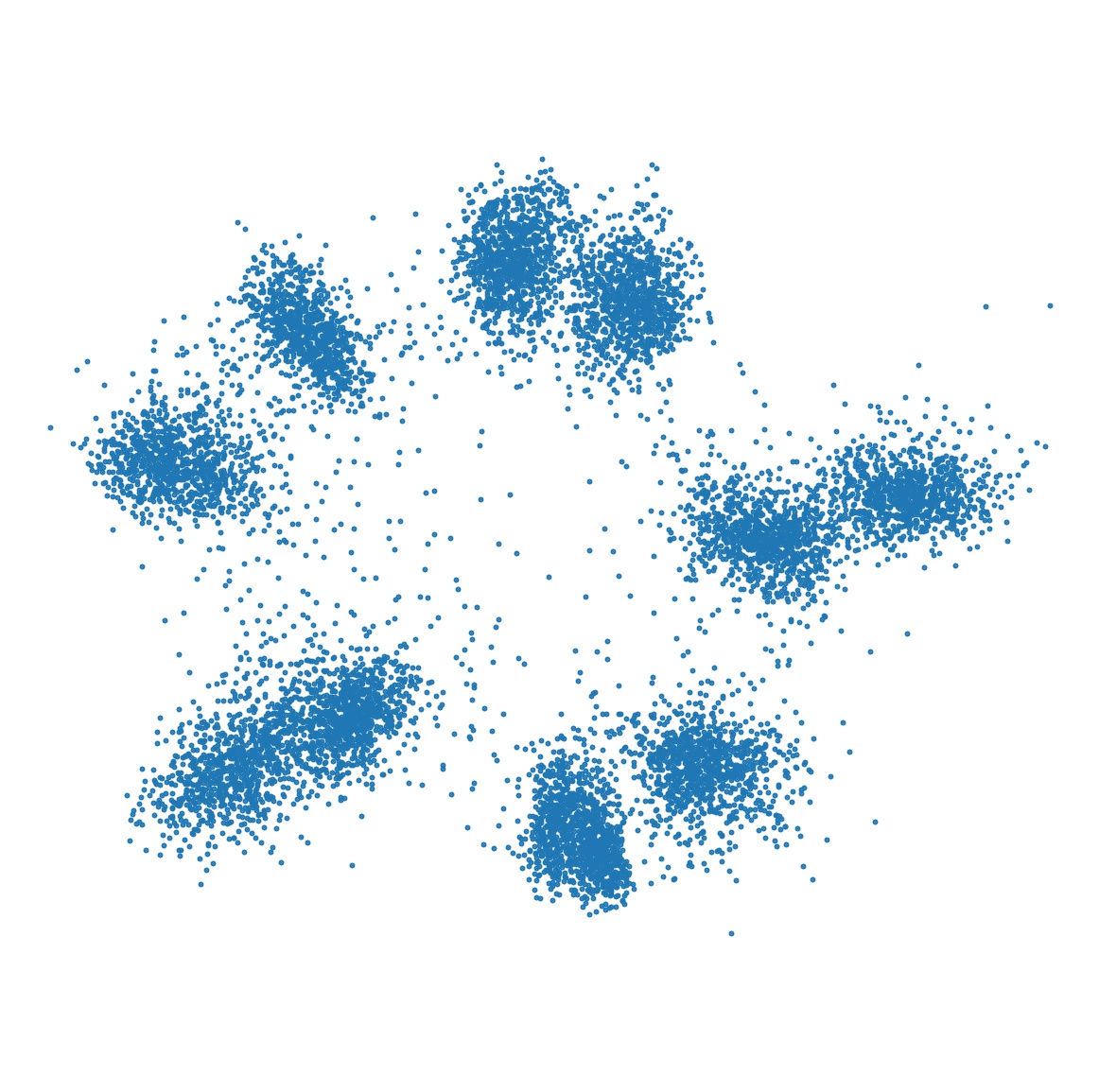}}
\subfigure[Isomap]{\includegraphics[width=1.3in]{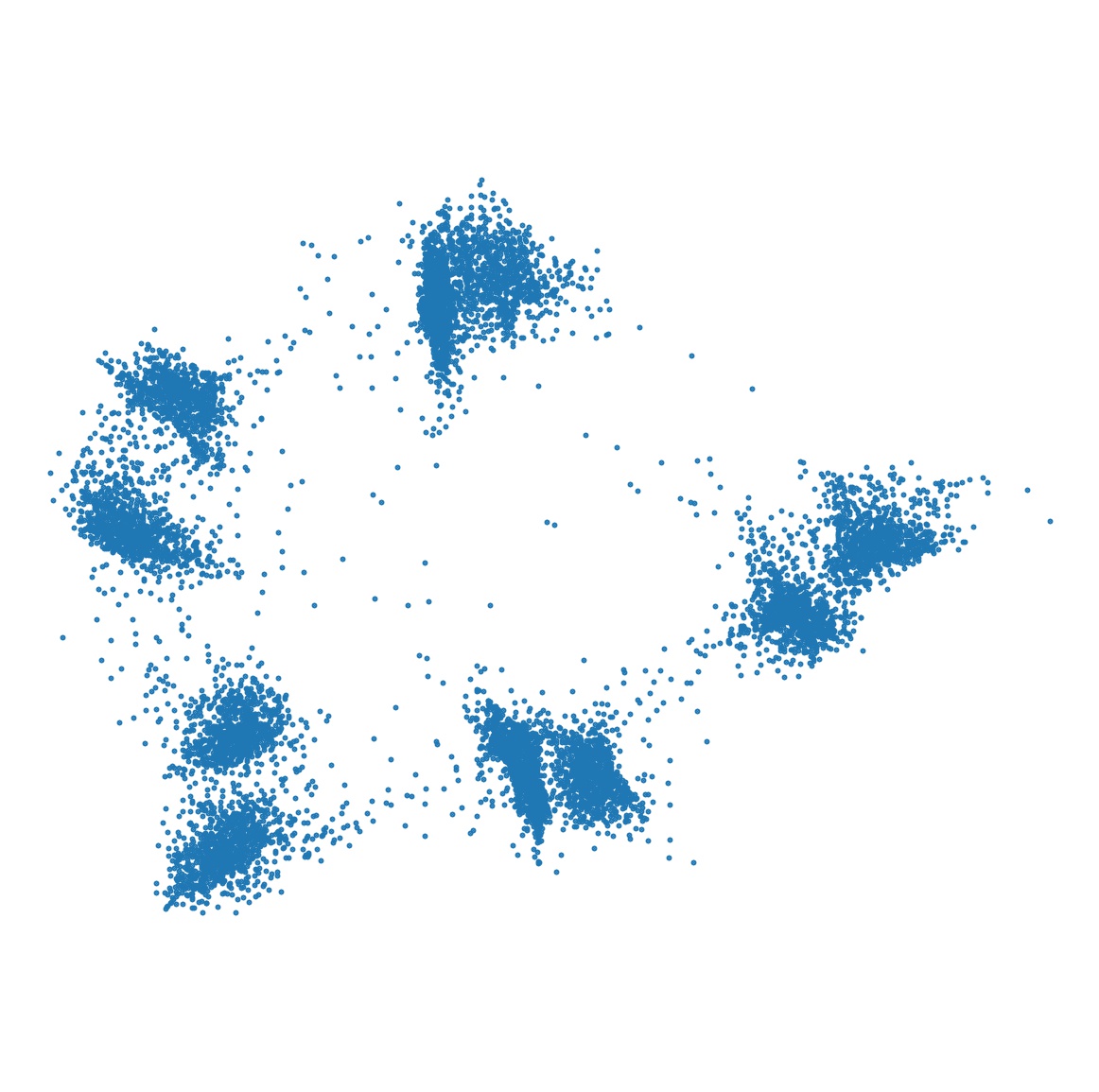}}
\subfigure[$t$-SNE]{\includegraphics[width=1.3in]{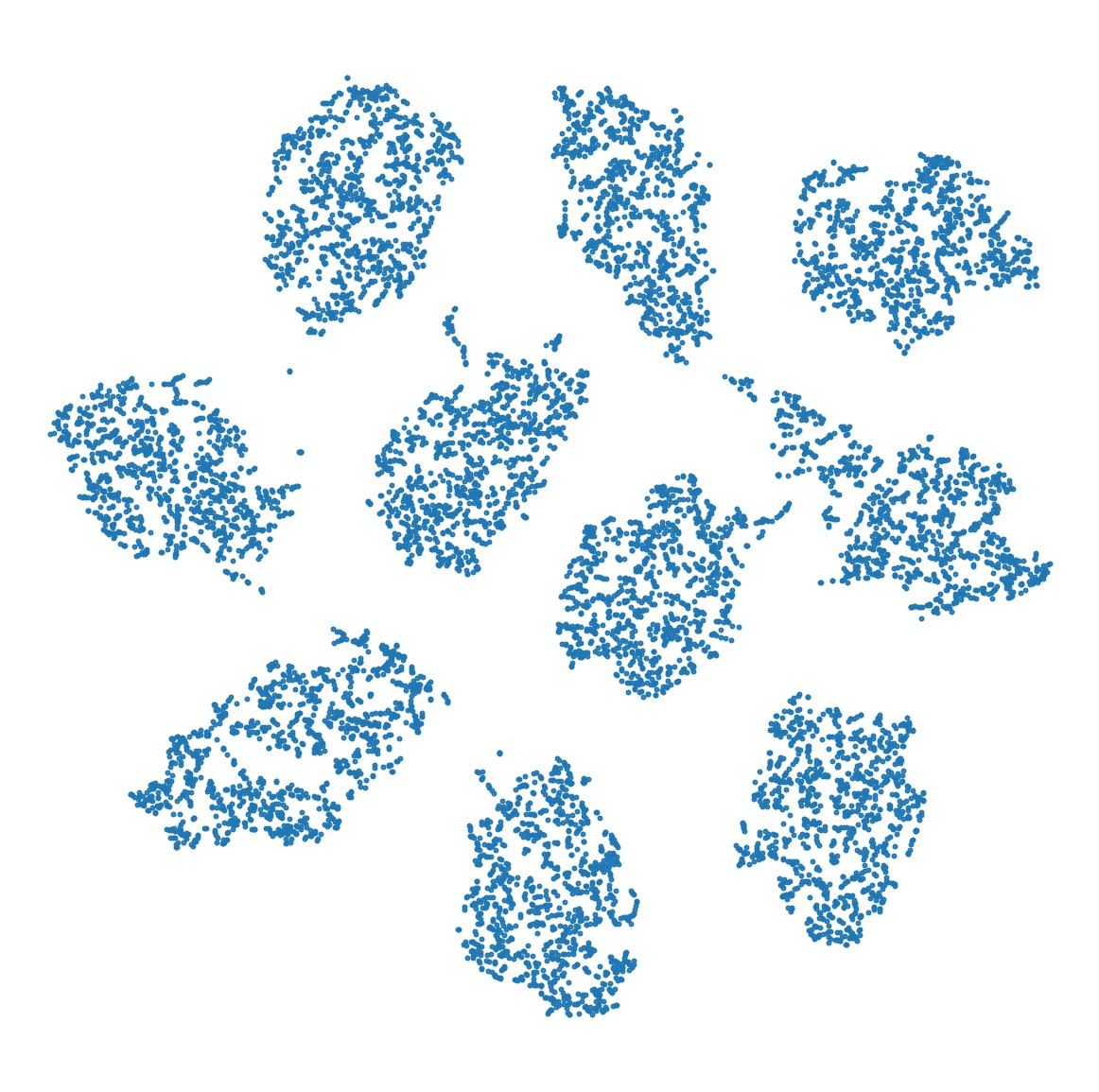}}
\subfigure[UMAP]{\includegraphics[width=1.3in]{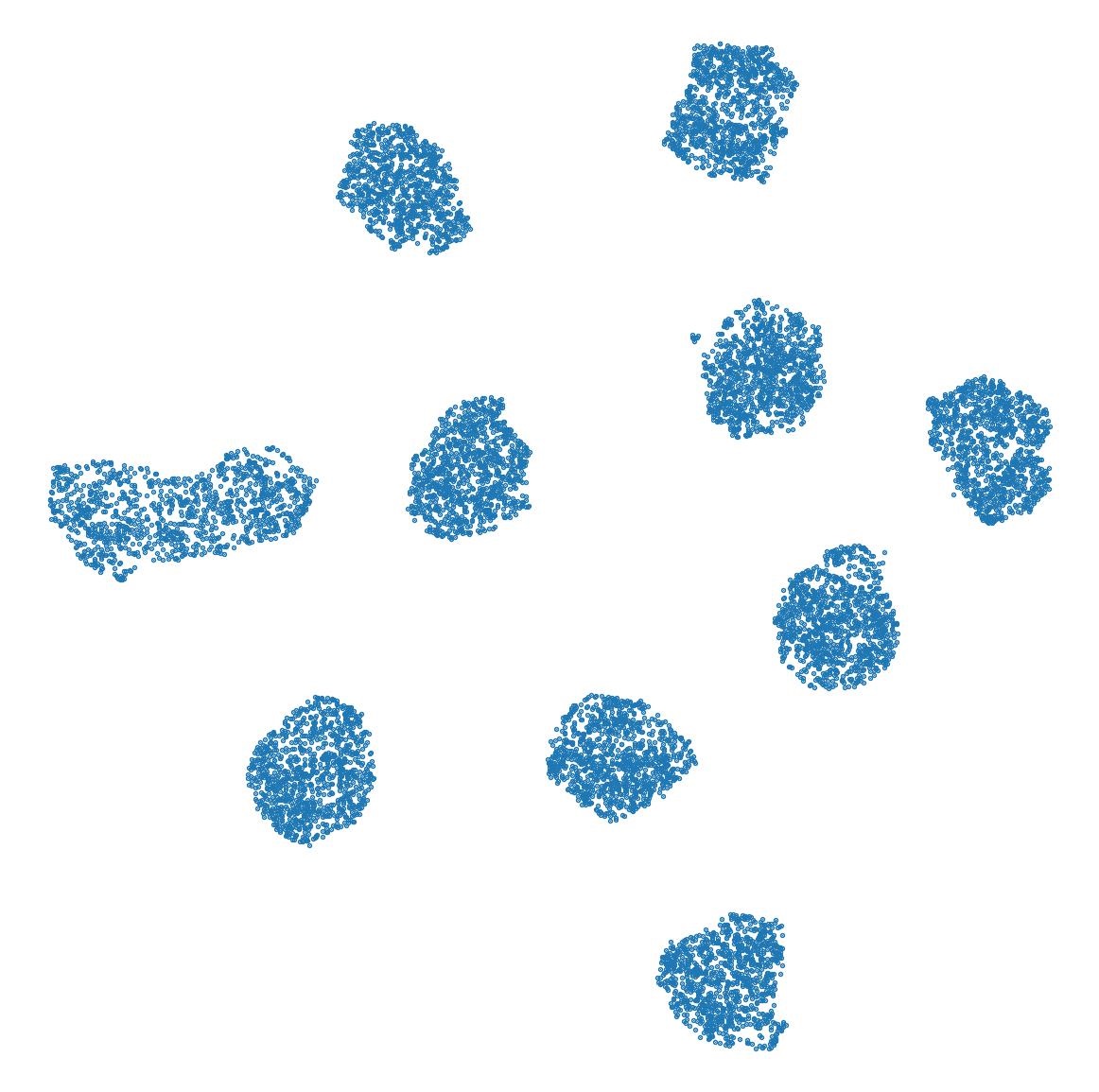}}
\caption{The visualized results of deep features by different visualization methods on MNIST-test.}
\label{vMNIST-test}
\end{figure}

\begin{figure}[h]
\center
\subfigure[MDS]{\includegraphics[width=1.3in]{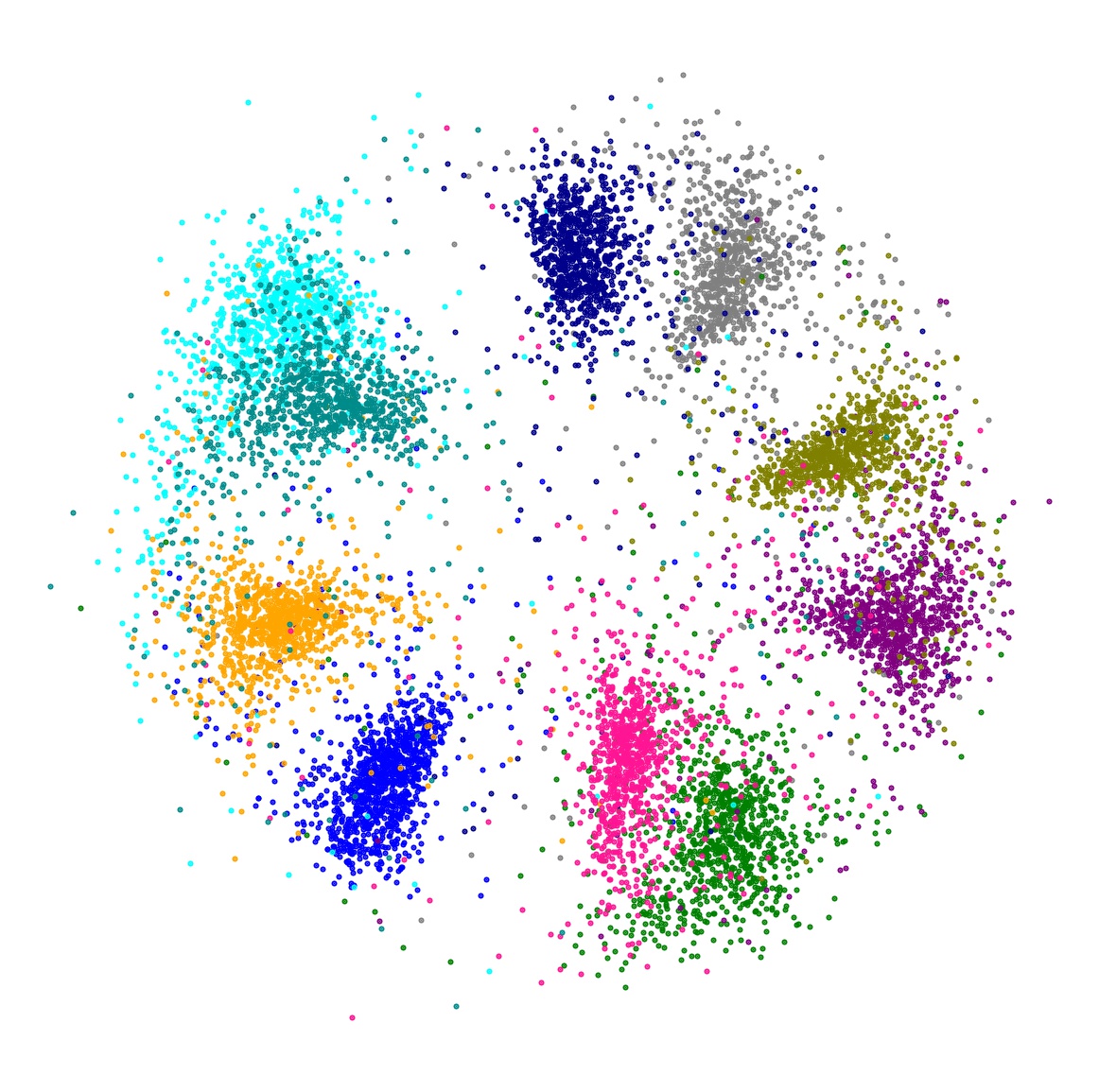}}
\subfigure[PCA]{\includegraphics[width=1.3in]{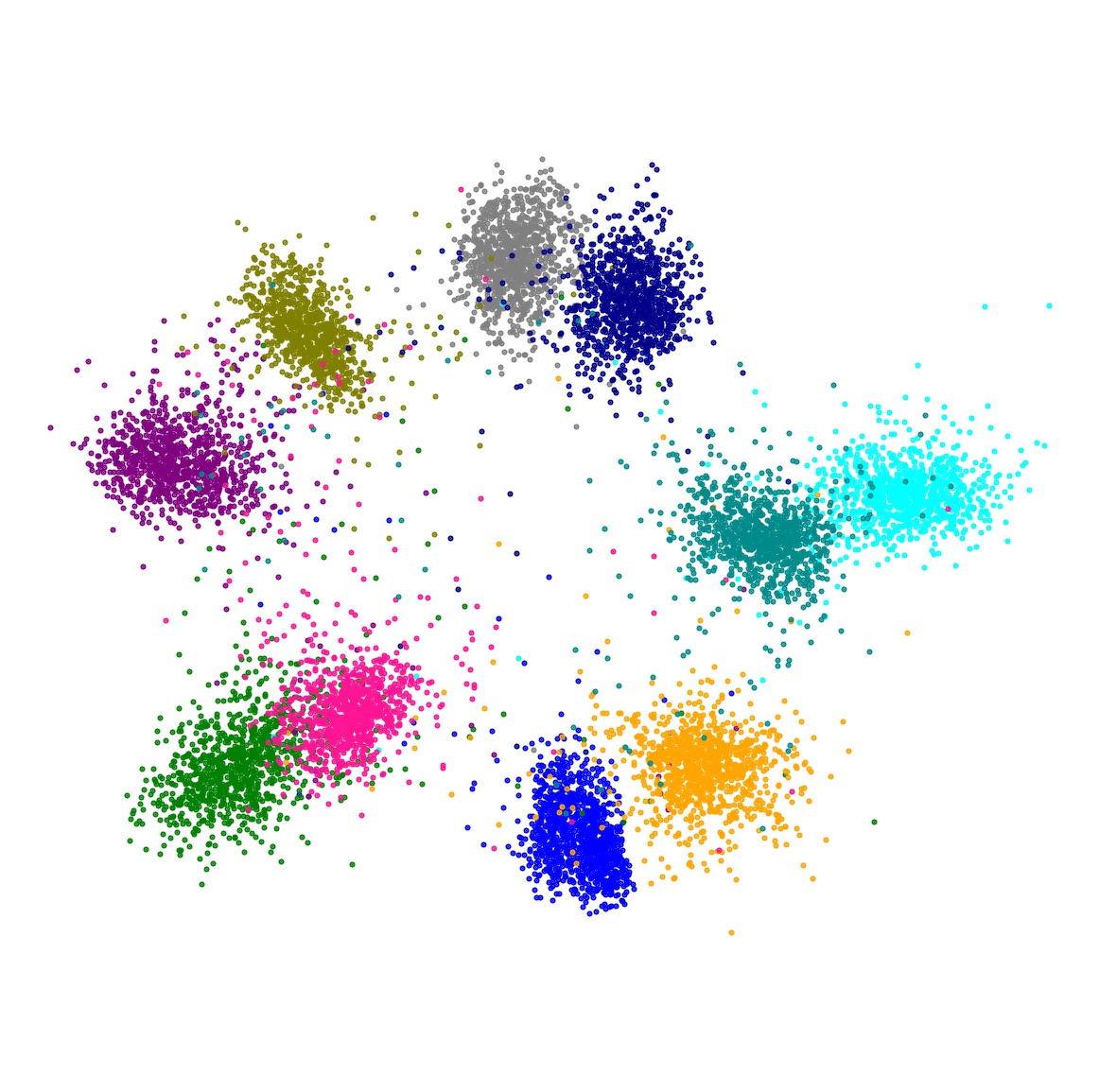}}
\subfigure[Isomap]{\includegraphics[width=1.3in]{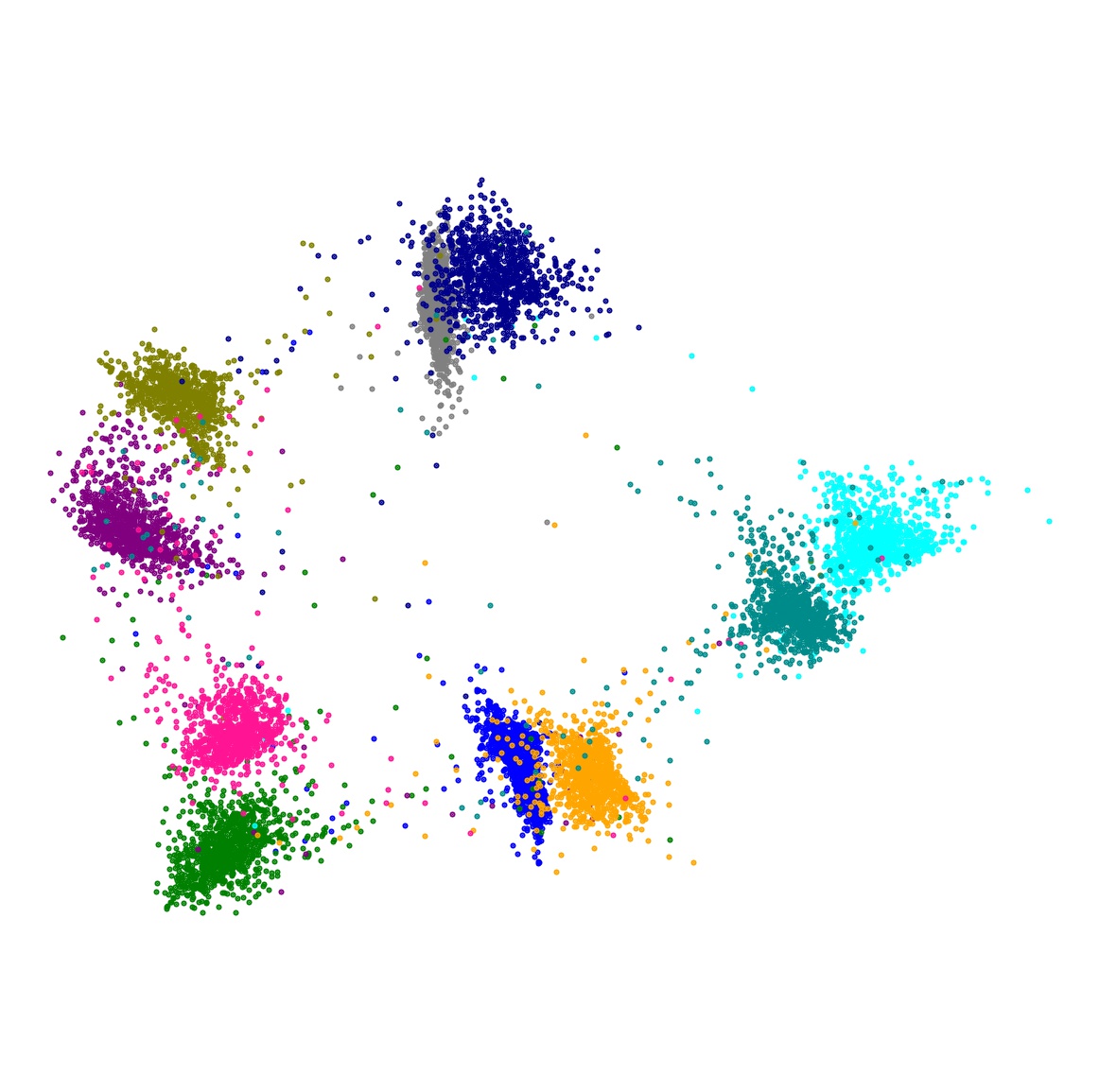}}
\subfigure[$t$-SNE]{\includegraphics[width=1.3in]{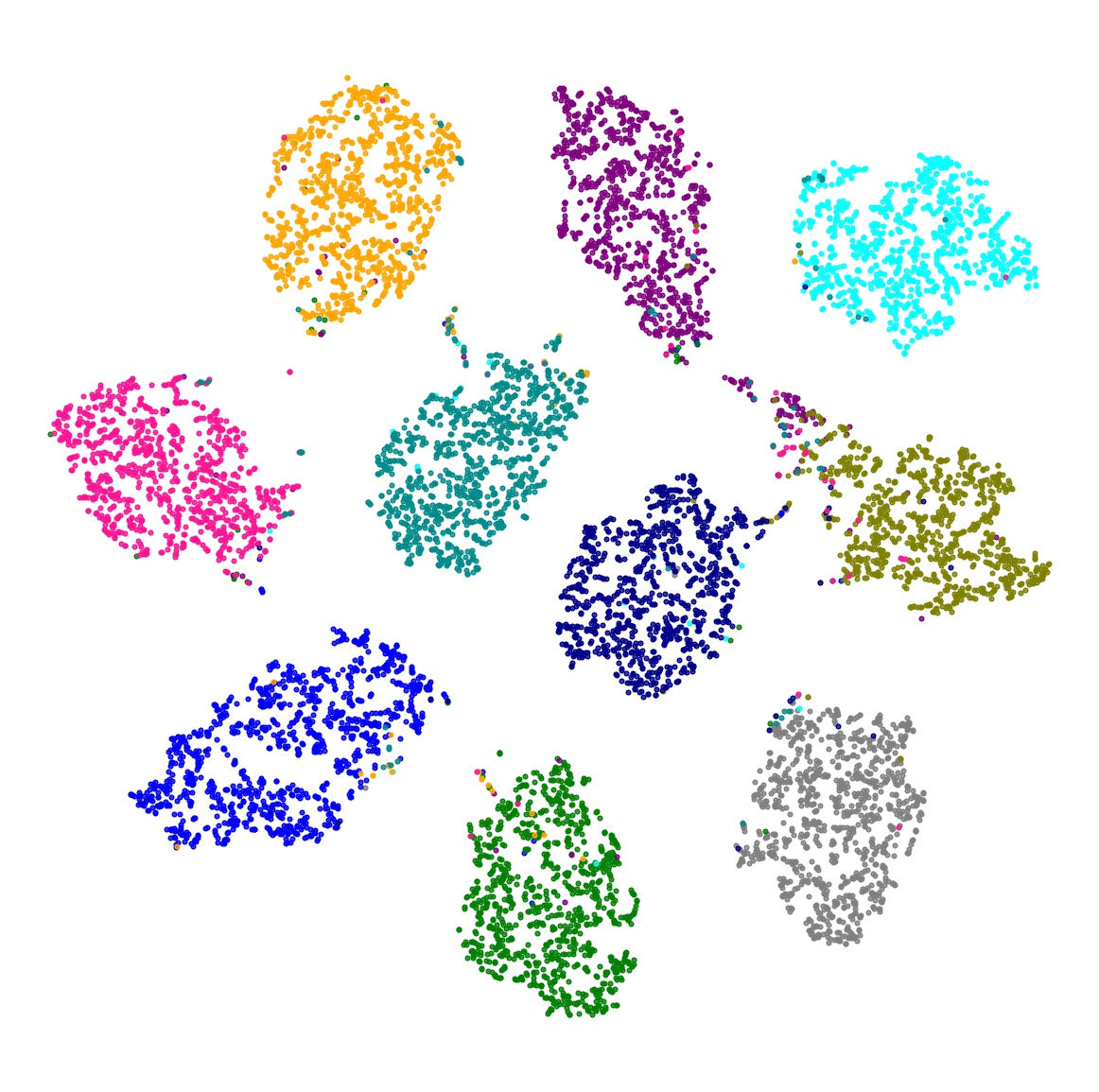}}
\subfigure[UMAP]{\includegraphics[width=1.3in]{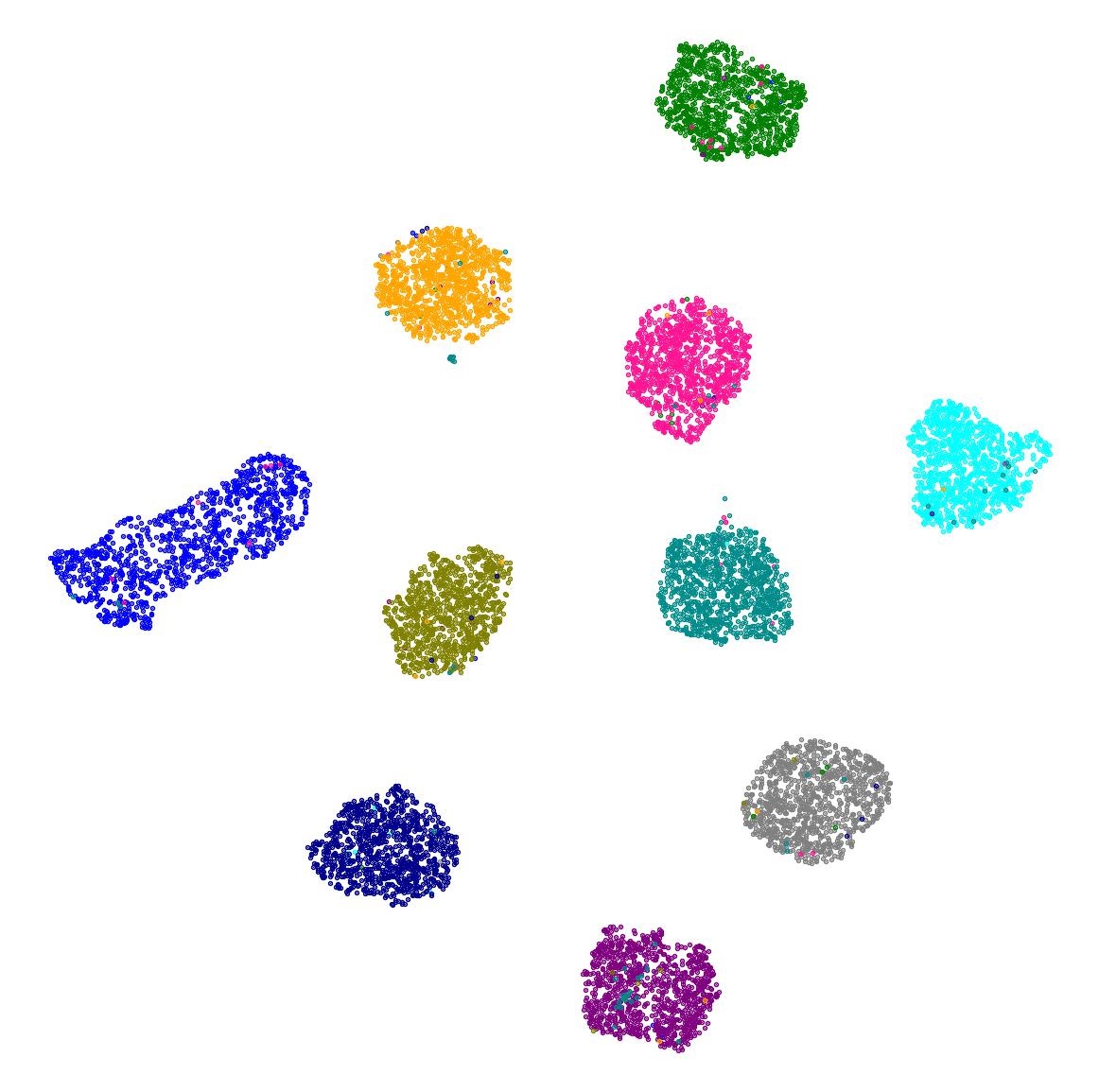}}
\caption{The visualized results of deep features by different visualization methods on MNIST-test. The different colours present the labels of the samples.}
\label{vMNIST-testl}
\end{figure}

For an informative clustering result that is more than pseudo sample labels for high-dimensional data, researchers use some embedding methods, such as MDS and t-SNE, to visualize the sample structures in the embedded space of deep clustering~\cite{zang2022evnet}.
However, the visualized result may be inconsistent with the clustering result.
One reason is that it is hard to visualize all of the distinguishing information of high-dimensional in $2$D space.
The other one is that these methods may have biases in the visualization.
To give a detailed analysis, in Fig~\ref{vMNIST-test} and \ref{vMNIST-testl}, we show the visualized results of five popular embedding methods including MDS~\cite{kruskal1978multidimensional}, PCA~\cite{abdi2010principal}, Isomap~\cite{balasubramanian2002isomap}, $t$-SNE~\cite{van2008visualizing}, and UMAP~\cite{mcinnes2018umap} in the embedded space of our GLDC on MNIST-test.
Note that in this deep clustering, the clustering accuracy (ACC) of MNIST-test is 0.97.
In other words, most distinguishing information of the different classes in MNIST-test is included in the embedded space.

As we can see from Fig.~\ref{vMNIST-test} and \ref{vMNIST-testl}, although most distinguishing information of the different classes is included in the embedded space, we still hard to recognize the pink, green, cyan, and dark cyan classes by MDS and PCA without the colour labels.
This is because MDS and PCA cannot visualize all of the distinguishing information in embedded space with $2$ dimensions.
Isomap can better visualize the dissimilarities between the different classes than MDS and PCA, but we are hard to recognize the grey and dark blue classes by Iosmap.
$t$-SNE and UMAP can produce a better visualization than other methods, but they cannot well capture the global-structure of the data in deep clustering, i.e., $t$-SNE and UMAP cannot visualize similar classes, such as green and pink classes.
Furthermore, $t$-SNE also cannot show the sample distribution, such as the density distribution of different classes, to explain the clustering result.

\begin{figure}
\centering
  \includegraphics[width=3in]{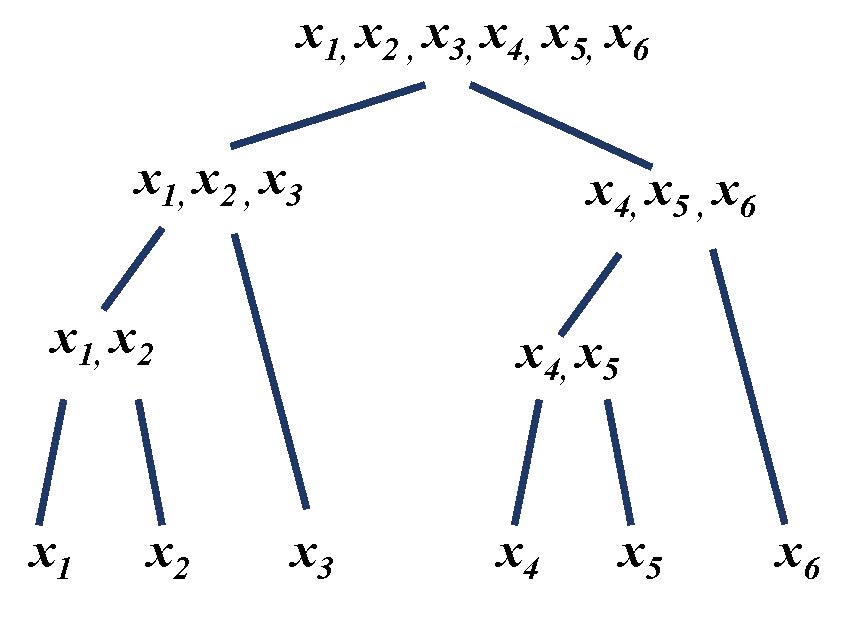}\\
  \caption{An example of dendrogram in hierarchical clustering.} 
  \label{exp2}
\end{figure}

Agglomerative hierarchical clustering combines the samples in data into clusters, those clusters into larger clusters, and so forth, creating a hierarchy of clusters~\cite{rani12013study}.
In this way, some agglomerative hierarchical clustering methods can mine the similarities between clusters and outliers in high-dimensional data and then present the mined knowledge in a dendrogram from the clustering process~\cite{nielsen2016hierarchical,ester1996density}.
Fig.~\ref{exp2} is an example to show the dendrogram with six samples.

As we can see the limitations of the presentations in the dendrogram are as follows.
1) The dendrogram cannot show the in-between samples of the different clusters and the clustering probability distributions, thus the recognized similar clusters and outliers cannot be well explained.
2) It is hard for a dendrogram to show all of the samples by the leaves when the number of samples is large.

\subsection{How to Improve the Visualization of High-dimensional Clustering}
From the above analysis, we should improve the visualization of high-dimensional clustering in the following three goals.
1) There should be an indicator for the cluster of each non-outlier sample and the visualization should be consistent with the clustering result.
2) Besides the clusters, the similarities between different clusters, and outliers also should be analyzed and then visualized.
3) The visualization should be an explanation for the clustering result.
For example, the mapped in-between samples of the different clusters can be used to explain the recognized similar clusters and the clustering probability distributions of different clusters can be used to explain the recognized outliers.

Combining deep clustering with RadViz Deluxe can be a solution for improving the visualization of high-dimensional clustering.
By mapping the extracted clustering probability distributions on the deep clustering in RadViz Deluxe, the cluster of each non-outlier sample can be indexed by its nearest cluster anchor and thus the visualization will be consistent with the clustering result~\cite{zhao2018evaluating}.
When we analyze the similarities between the different clusters in deep clustering, the optimal Hamiltonian cycle generated by these similarities is the global optimal solution for placing the cluster anchors on the circumference of a circle.
In this way, the similarities between the different clusters can be well presented in the visualization.
Finally, each sample at the centre of the circle can be seen as an outlier, because it has a low probability of any cluster and the mapped in-between samples of different clusters can be used to explain the recognized similar clusters.

\section{The Algorithms of HCHC}
\label{SuppC1}
The algorithm of GLDC is summarized in Algorithm~\ref{alg:GLDC}.
\begin{algorithm}[H]
  \caption{GLDC}
\label{alg:GLDC}
\begin{algorithmic}[1]
 \STATE Initialize $\theta$, $\theta'$, $\phi$, $\beta_1$, $\beta_2$, and $\gamma$
 \STATE Initialize a random process $\mathcal{N}$
 \STATE Get $\bm{\widetilde{X}}$ of $\bm{X}$ by \eqref{genaug}
 \FOR {$episode = 0$ to $pretraining-episode_{max}$}
 \FOR {mini-batch in $\bm{X}$}
 \STATE Update $\theta$ and $\theta'$ by minimizing reconstruction loss in \eqref{mresloss}
 \ENDFOR
 \ENDFOR
 \FOR {$episode=0$ to $episode_{max}$}
 \FOR {mini-batch in $\bm{X}$}
 \STATE Get the embedded samples by $\bm{\bm{z_i}} = G_{\theta}(\bm{x_i})$
 \STATE Compute the reconstruction loss by \eqref{mresloss}
 \STATE Construct weighted adjacency matrix $\bm{W^{l}}$ by \eqref{congraph}
 \STATE Get the clustering probability distributions by \eqref{compro}
 \STATE Compute the graph learning loss by \eqref{graphloss}
 \STATE Compute the augmentation learning loss by \eqref{augloss}
 \STATE Update $\theta$ and $\phi$ by minimizing the clustering loss in \eqref{cluloss}
 \ENDFOR
 \ENDFOR
 \STATE Get $\bm{P} = \{\bm{p_1},\bm{p_2}, \cdots, \bm{p_n}\}$.
 \end{algorithmic}
\end{algorithm}\par

The whole algorithm of our mapping is summarized in Algorithm~\ref{alg:HCHC}
\begin{algorithm}[H]
  \caption{HCHC}
\label{alg:HCHC}
\begin{algorithmic}[1]
 \STATE Initialize $\bm{state} = 1$
 \STATE Initialize $\bm{cur} = (1 << c)-1 $
 \STATE Initialize $last = 1$
 \STATE Compute the dissimilarities between each pair of $\bm{\rho_i}$ and $\bm{\rho_j}$ based on \eqref{disfp}
 \WHILE {$\bm{state} < (1 << c)$}
 \FOR {$j = 2$ to $c$}
 \IF {$(\bm{state}\& (1<<j)!= 0)$}
 \FOR {$k = 2$ to $c$}
 \IF {$(\bm{state}\& (1<<k)!= 0)$}
 \STATE Update $dp(\bm{\rho_1},\bm{\rho_j},\bm{state})$ by \eqref{dpfhc}
 \ENDIF
 \ENDFOR
 \ENDIF
 \ENDFOR
 \STATE $\bm{state}$ = $\bm{state} +2$
 \ENDWHILE
 \FOR {$i = c$ to $1$}
 \STATE temp = 1
 \FOR {$j = 1$ to $c$}
 \IF {$((\bm{cur} \& 1 << j) != 0$ and $dp(\bm{\rho_1},\bm{\rho_j},\bm{cur}) + dis(\bm{\rho_{temp}}, \bm{\rho_{last}}) > dp(\bm{\rho_1},\bm{\rho_j},\bm{cur}) + dis(\bm{\rho_{j}}, \bm{\rho_{last}})$}
 \STATE temp = j
 \ENDIF
 \ENDFOR
 \STATE $\bm{\rho^i} = \bm{\rho_{temp}}$
 \STATE $\bm{cur} \oplus = 1 << tem$
 \STATE $last = temp$
 \ENDFOR
 \STATE Compute $\alpha_{\bm{\rho^i}}$ for each $\bm{\rho^i}$ by \eqref{anglef}
 \STATE Compute $\bm{\mu_{\rho^i}}$ for each $\bm{\rho^i}$ by on \eqref{pvf}
 \STATE Compute $\bm{\mu_{x_i}}$ in $\bm{P}$ by on \eqref{pvp}
 \STATE Map each $\bm{\mu_{\rho^i}}$ on a circle
 \STATE Map each $\bm{\mu_{x_i}}$ in the circle
 \end{algorithmic}
\end{algorithm}

\section{The Theoretical Analysis of Hamiltonian Cycle Mapping}
\label{SuppD}

\subsection{Analysis on the Mapping of Cluster Similarity}
\label{SuppD1}
The Hamiltonian cycle problem is formulated by an Irish mathematician, William Rowan Hamilton, to ask whether there is a cycle in a graph $\bm{G}$ passes through every vertex $\bm{v_i}$ exactly once, except the first passed one $\bm{v_1}$
as $\Pi=\{\bm{v^1}, \bm{v^2}, \cdots, \bm{v^m}, \bm{v^1}\}$, where $m$ is the number of the vertices~\cite{dirac1952some}.
If $\bm{G}$ has a Hamiltonian cycle, $\bm{G}$ is Hamiltonian.
Hamiltonian cycle is highly related to a classical problem, the travelling salesman.
This problem is defined as follows~\cite{hoogeveen1991analysis}.
\begin{definition}
Given a complete undirected graph $\bm{G}$ on $m$ vertices and a distance $dis(\bm{v_i},\bm{v_j})$ for each edge between $\bm{v_i}$ and $\bm{v_j}$, find a Hamiltonian cycle of minimum total length.
This Hamiltonian cycle is the optimal Hamiltonian cycle.
\end{definition}
In HCHC, we use Hamiltonian cycle to present the similarities between different clusters.
A Hamiltonian cycle can present the sampled similarities $\{s(\bm{\rho^1},\bm{\rho^2}),\cdots,$ $s(\bm{\rho^{c-1}},\bm{\rho^c}), s(\bm{\rho^1},\bm{\rho^c})\}$ from $\{s(\bm{\rho_i},\bm{\rho_j})|i = 1,\cdots,c,i<j\}$.
Define $t(\bm{\rho_i},\bm{\rho_j})$ as the weighted $s(\bm{\rho_i},\bm{\rho_j})$ as
\begin{eqnarray}
\label{weightf}
t(\bm{\rho_i},\bm{\rho_j}) =T[s(\bm{\rho_i},\bm{\rho_j})]
\end{eqnarray}
where $T(\cdot)$ is the weighting function and for any $s$, $-1<T(s)<1$.
This weight can be computed by different definitions to select the aspect for sampling the similarities.
In this paper, we define $t(\bm{\rho_i},\bm{\rho_j}) = s(\bm{\rho_i},\bm{\rho_j})$.
This definition means that the similarity and its importance are in the direct ratio.
We have the following theorem to get the global optimal mapping of the similarities in the selected aspect.
\begin{theorem}
The mapping of the optimal Hamiltonian cycle will maximize the sum of the weights of the selected similarities as
\begin{eqnarray}
\label{samplingloss}
S_{sam} =\sum_{i=1}^{c-1}t(\bm{\rho^i},\bm{\rho^{i+1}})+t(\bm{\rho^1},\bm{\rho^c})
\end{eqnarray}
\end{theorem}

The proof of the above theorem is as follows.
\begin{proof}
We can define a normalized $dis(\bm{\rho_i},\bm{\rho_j})$ for $\bm{\rho_i},\bm{\rho_j}$ as
\begin{eqnarray}
\label{disf}
dis(\bm{\rho_i},\bm{\rho_j}) = \frac{1-t(\bm{\rho_i},\bm{\rho_j})}{\sum_{i=1}^{c-1}\sum_{j=i+1}^{c}(1-t(\bm{\rho_i},\bm{\rho_j}))}
\end{eqnarray}
Then
\begin{eqnarray}
\label{samplingsumdis}
\sum_{i=1}^{c-1}dis(\bm{\rho^i},\bm{\rho^{i+1}})+dis(\bm{\rho^1},\bm{\rho^c}) = \frac{c-(\sum_{i=1}^{c-1}t(\bm{\rho^i},\bm{\rho^{i+1}})+t(\bm{\rho^1},\bm{\rho^c}))}{\sum_{i=1}^{c-1}\sum_{j=i+1}^{c}(1-t(\bm{\rho_i},\bm{\rho_j}))}
\end{eqnarray}
As we can see, $c$ and $\sum_{i=1}^{c-1}\sum_{j=i+1}^{c}(1-t(\bm{\rho_i},\bm{\rho_j}))$ are two invariant constants, thus $S_{sam}$ can be maximized by
\begin{eqnarray}
\label{samplinglossm}
\arg \min_{\Pi}||\sum_{i=1}^{c}dis(\bm{\rho^i},\bm{\rho^{i+1}})+dis(\bm{\rho^1},\bm{\rho^c})||_2
\end{eqnarray}
Therefore, to get the global optimal mapping of the similarities, this problem can be solved by the optimal Hamiltonian cycle $\Pi^{*}$ of $\bm{G_P}$ with the vector $\bm{\rho_{i}}$ and the distance $dis(\bm{\rho_{i}}, \bm{\rho_{j}})$.
\end{proof}

We can define $T[s(\bm{\rho_i},\bm{\rho_j})]$ as
\begin{eqnarray}
\label{disf}
T(s(\bm{\rho_i},\bm{\rho_j})) = sgn(s(\bm{\rho_i},\bm{\rho_j}))\times |s(\bm{\rho_i},\bm{\rho_j})|^\gamma
\end{eqnarray}
where $sgn(\cdot)$ is a sign function.
\begin{eqnarray}
\label{sgnr}
sgn(s(\bm{\rho_i},\bm{\rho_j})) =
\left\{
             \begin{array}{lr}
              1 & s(\bm{\rho_i},\bm{\rho_j})>0\\
              -1 & \text{otherwise}
             \end{array}
\right.
\end{eqnarray}

If $\gamma = 0$, the weights of all similarities in $\{s(\bm{\rho_i},\bm{\rho_j})|i = 1,\cdots,c,i<j\}$ will be same, in this case $S_{sam}$ can be maximized by a randomly selected Hamiltonian cycle $\Pi$.
If $\gamma \to \infty$, $S_{sam}$ can be maximized by selecting the shortest unselected edge that cannot form the cycle in each iteration, except the last iteration.
In the last iteration, we select the edge that connects the first vertex and the last vertex.
The time complexity of this process is $O(c^3)$, in this complexity, we can get the orders of $1000$ clusters in acceptable time consumption.

\begin{figure}[]
\center
\subfigure[]{\includegraphics[width=2in]{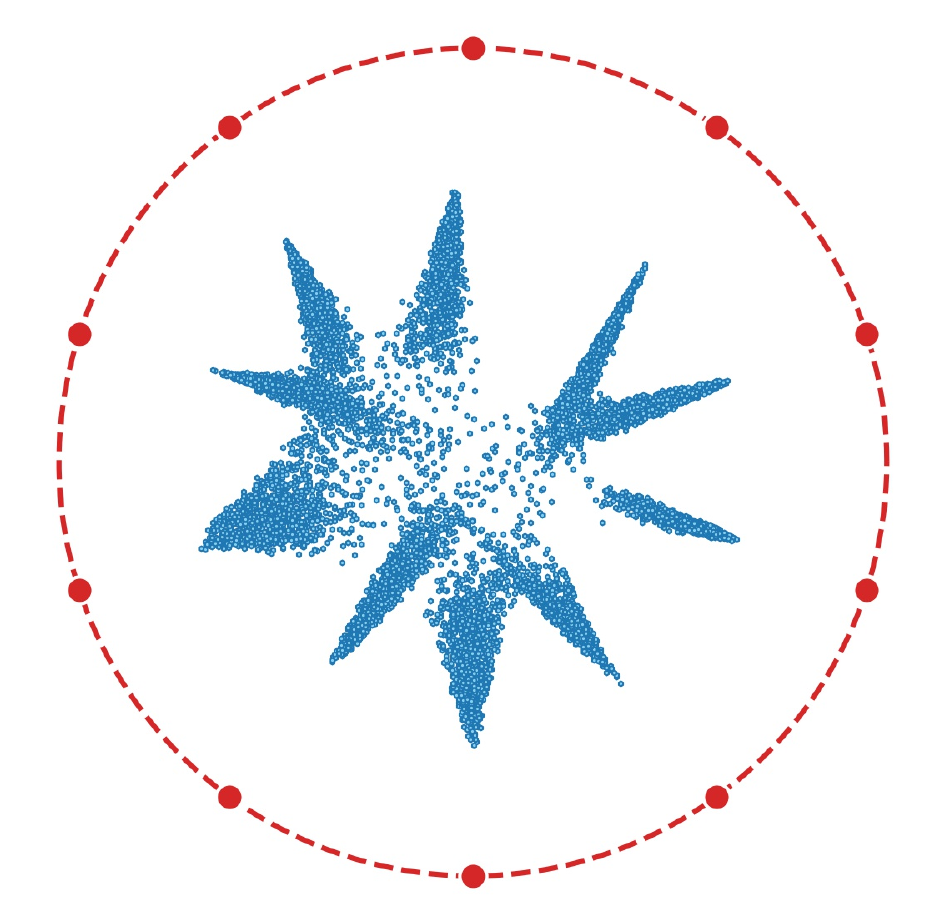}}
\subfigure[]{\includegraphics[width=2in]{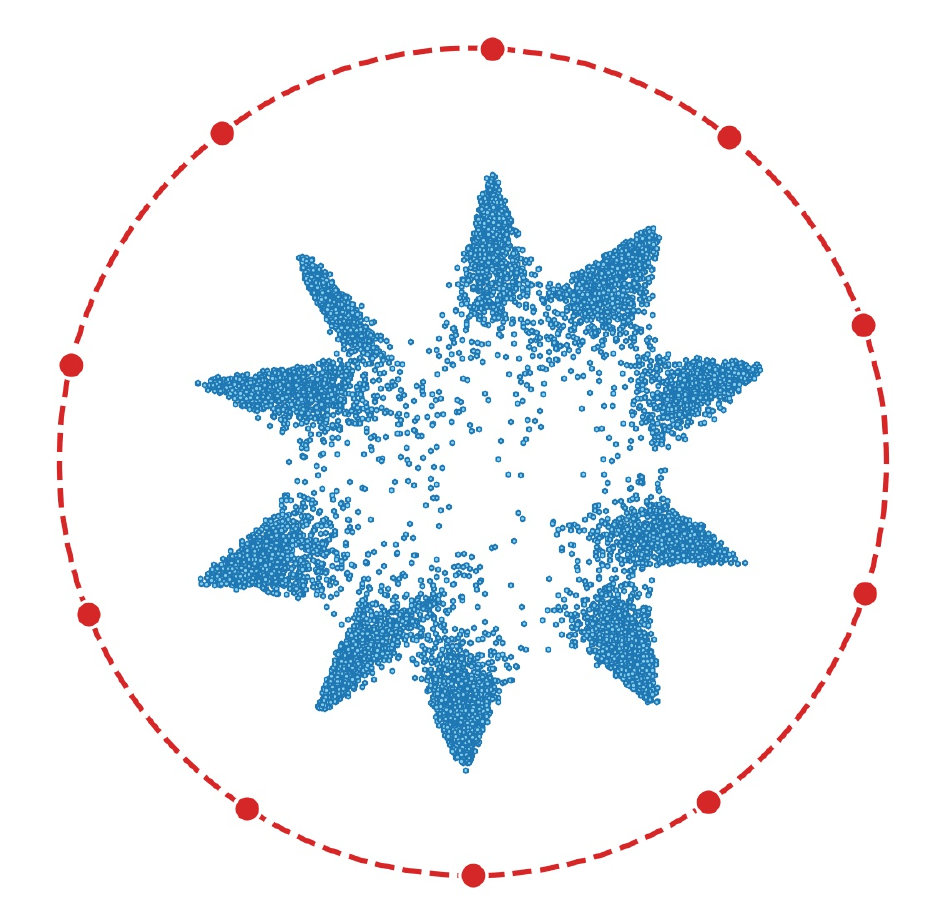}}
\subfigure[]{\includegraphics[width=2in]{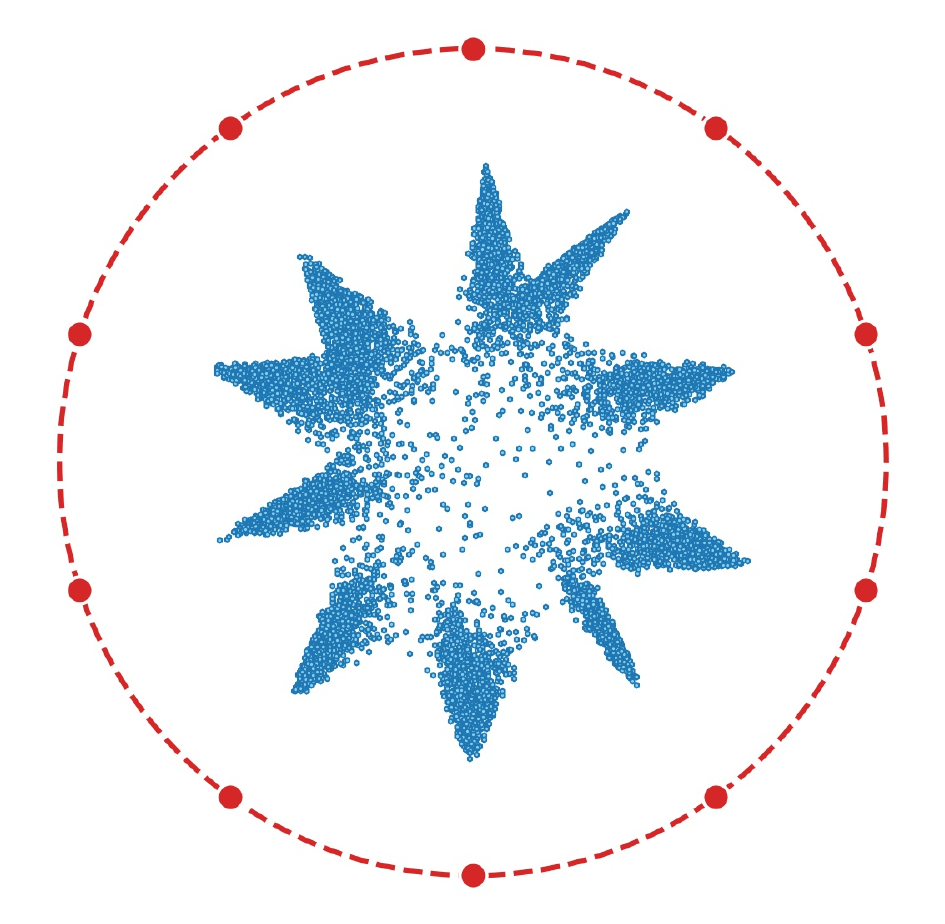}}

\caption{Visualized results with different $\gamma$. (a) Random Hamiltonian cycle with the same similarity, e.g., $\gamma=0$.  (b) $\gamma=1$. (c) $\gamma=1000$.}
\label{vgamma}
\end{figure}
The visualized results with different $\gamma$ on MNIST-test are shown in Fig.~\ref{vgamma}.
As we can see, the similarities of the different clusters in MNIST-test can be well presented when $\gamma=1$ but cannot be well presented by the randomly selected Hamiltonian cycle.
When $\gamma$ is very large, these similarities can be presented with acceptable time consumption.

\subsection{Analysis on the Mapping Performance}
\label{SuppD2}
The main idea to improve the mapping of Radviz is that the further away a point is from the center, the more informative its position is, being the point closer to the attributes having the highest values~\cite{angelini2019towards}.
Thus, on the circumference of a circle, the objective function to improve mapping by the layout of the anchors can be defined as
\begin{eqnarray}
\label{errorm}
\arg \max_{\Pi = \{\bm{\rho^1};\cdots; \bm{\rho^c}\}} \sum_{i=1}^{n}||\sum_{j=1}^{c}p_{i,j}\mu_{\bm{\rho^j}}||_2^2
\end{eqnarray}
This objective function aims to make the points far away from the center.
Define $r=1$, then we have
\begin{eqnarray}
&&\sum_{i=1}^{n}||\sum_{j=1}^{c}p_{i,j}\mu_{\bm{\rho^j}}||_2^2\\ \nonumber
&=&\sum_{i=1}^{n}||\sum_{j=1}^{c}p_{i,j}cos(\alpha_{\bm{\rho^j}}), \sum_{j=1}^{c}p_{i,j}sin(\alpha_{\bm{\rho^j}})||_2^2 \\ \nonumber
&=&\sum_{i=1}^{n}\sum_{j=1}^{c}p_{i,j}^2(cos(\alpha_{\bm{\rho^j}})^2 +sin(\alpha_{\bm{\rho^j}})^2)\\ \nonumber
&+& 2\sum_{i=1}^{n}\sum_{j=1}^{c-1}\sum_{k=j+1}^{c}p_{i,j}p_{i,k}(cos(\alpha_{\bm{\rho^j}})cos(\alpha_{\bm{\rho^k}})+sin(\alpha_{\bm{\rho^j}})sin(\alpha_{\bm{\rho^k}})) \\ \nonumber
&=& \sum_{i=1}^{n}\sum_{j=1}^{c}p_{i,j}^2(cos(\alpha_{\bm{\rho^j}})^2 +sin(\alpha_{\bm{\rho^j}})^2)+ 2\sum_{i=1}^{n}\sum_{j=1}^{c-1}\sum_{k=j+1}^{c}p_{i,j}p_{i,k}(cos(\alpha_{\bm{\rho^k}}-\alpha_{\bm{\rho^j}}))\\ \nonumber
&=& \sum_{i=1}^{n}\sum_{j=1}^{c}p_{i,j}^2+ 2\sum_{i=1}^{n}\sum_{j=1}^{c-1}\sum_{k=j+1}^{c}p_{i,j}p_{i,k}cos(\alpha_{\bm{\rho^k}}-\alpha_{\bm{\rho^j}}) \\ \nonumber
&=& \sum_{i=1}^{n}\sum_{j=1}^{c}p_{i,j}^2+\sum_{j=1}^{c-1}\sum_{k=j+1}^{c}(\bm{\rho^j})^\intercal \bm{\rho^k}cos(\alpha_{\bm{\rho^k}}-\alpha_{\bm{\rho^j}})
\end{eqnarray}
Thus the objective function~(\ref{errorm}) can be maximized by 
\begin{eqnarray}
\label{errormc}
\arg \max_{\Pi = \{\bm{\rho^1};\cdots; \bm{\rho^c}\}} \sum_{j=1}^{c-1}\sum_{k=j+1}^{c}(\bm{\rho^j})^\intercal \bm{\rho^k}cos(\alpha_{\bm{\rho^k}}-\alpha_{\bm{\rho^j}})
\end{eqnarray}
Based on Pearson correlation coefficient, the normalized objective function can be defined by as 
\begin{eqnarray}
\label{errormcc}
\arg \max_{\Pi = \{\bm{\rho^1};\cdots; \bm{\rho^c}\}} \sum_{j=1}^{c-1}\sum_{k=j+1}^{c}\frac{(\bm{\rho^j}-\overline{\bm{\rho^j}})^\intercal (\bm{\rho^k}-\overline{\bm{\rho^k}})}{|\bm{\rho^j}-\overline{\bm{\rho^j}}||\bm{\rho^k}-\overline{\bm{\rho^k}}|}cos(\alpha_{\bm{\rho^k}}-\alpha_{\bm{\rho^j}})
\end{eqnarray}
To solve this problem, we should put the anchors of similar clusters together.
From the analysis in Appendix~\ref{SuppD1}, we can see that the optimal Hamiltonian cycle $\Pi^{*}$ is an approximate solution of the objective function~(\ref{errormcc}).

\section{Experiment Setting}
\label{SuppE}
We use seven datasets including MNIST~\cite{deng2012mnist}, Fashion~\cite{xiao2017fashion}, USPS~\cite{hull1994database}, Reuters10k~\cite{lewis2004rcv1}, HHAR~\cite{stisen2015smart}, Pendigits~\cite{asuncion2007uci}, and BH~\cite{abdelaal2019comparison} to illustrates the effectiveness of HCHC.
The details of the datasets are shown in Table~\ref{DataDescription}.

There are some parameters that need to be tuned.
The initial $\beta_1$ is set as $5$.
As we can see, with the increase of the iteration numbers in training, the magnitudes of $L_r$ and $L_a$ will become smaller and smaller.
Thus a discount factor $\gamma$ is used to tune the magnitude of $\beta_1$ in every iteration $t$ as
\begin{eqnarray}
\label{discountb}
\beta_1 = \gamma^t \beta_1
\end{eqnarray}
where $\gamma$ is set as 0.8.
$\beta_2$ is set as $10$.
$\sigma^2$ is set from $\{0.05,0.1, 0.2\}$.
$\xi$ is set from $\{0.005,0.05, 0.1, 0.2\}$.
$k$ is set from $\{3,4, 5, 30\}$.
The batch size is set as $128$.
We use Adam optimizer in our training and the learning rate is set as $0.002$.
The autoencoder is composed of eight layers with dimensions $D$-500- 500-2000-5-2000-500-500-$D$, where $D$ is the dimension of the input samples.

\begin{table}[ht]
\centering
\caption{Data description.}
\label{DataDescription}
\begin{tabular*}{8.5cm}{@{\extracolsep{\fill}}ccccc}	
  \hline
DID& Dataset      &  Instances & Features &  Classes  \\
 \hline
 1 & MNIST         & 70000               & 784                 & 10              \\
 2 & Fashion            & 70000               & 784                 & 10              \\
 3 & USPS              & 9298                & 256                & 10              \\
 4 & Reuters10k               & 10000                & 2000                & 4            \\
 5 & HHAR                    &10299                 & 561                &6     \\
 6 & Pendigits               & 10992                & 16                & 10           \\
 7 & BH                      & 8569                 & 17499               & 14           \\
 \hline
\end{tabular*}
\end{table}

We also compare our GLDC with nine clustering methods including $k$-means~\cite{macqueen1967some}, GMM~\cite{rasmussen1999infinite}, SC~\cite{shi2000normalized}, DEC~\cite{xie2016unsupervised}, IDEC~\cite{guo2017improved}, DSC~\cite{shaham2018spectralnet}, JULE~\cite{yang2016joint}, DSCDAN~\cite{yang2019deep}, N2D~\cite{mcconville2021n2d}.
The details of the compared methods are as follows.\\
1. $k$-means works by computing centres of the different clusters and cluster assignments iteratively by the Euclidean distance~\cite{macqueen1967some}.\\
2. GMM works by computing the centres of the different clusters and cluster assignments iteratively by the Gaussian model~\cite{rasmussen1999infinite}.\\
3. SC learns a map that embeds input data points into the eigenspace of their associated Laplacian matrix and then clusters them by $k$-means~\cite{shi2000normalized}.\\
4. DEC simultaneously learns feature representations and cluster assignments by deep neural networks~\cite{xie2016unsupervised}.\\
5. IDEC improves DEC by integrating the clustering loss and autoencoder reconstruction loss~\cite{guo2017improved}.\\
6. DSC learns a map that embeds input data points into the eigenspace of their associated Laplacian matrix and then clusters them, in the deep neural network~\cite{shaham2018spectralnet}.\\
7. JULE consists of a multinomial logistic regression function stacked on top of a multi-layer convolutional autoencoder in deep clustering~\cite{yang2016joint}.\\
8. DSCDAN discriminatively performs feature embedding and spectral clustering by CNN for image clustering~\cite{yang2019deep}.\\
9. N2D replaces the clustering layer with a manifold learning technique on the autoencoder representations~\cite{mcconville2021n2d}.

We use ACC and NMI to measure the clustering performance of GLDC with seven existing clustering methods~\cite{kuhn1955hungarian}.
The definition of ACC is as follows.
Denote $\bm{a}=\{a_{1}, a_{2}, \cdots , a_{n}\}$ as the clustering results and $\bm{b}=\{b_{1}, b_{2}, \cdots , b_{n}\}$ as the ground truth label of $X$.
ACC is defined as:
\begin{eqnarray}
ACC=\frac{\sum^{n}_{i=1} \delta (a_{i},map(b_{i}))}{n}
\end{eqnarray}
where $\delta (a,b) = 1$, if $a = b$ and $\delta (a,b) = 0$, otherwise.
$map(b_{i})$ is the best mapping function that permutes clustering labels to match the given truth labels using the Kuhn-Munkres algorithm.
The larger ACC is, the better the clustering result is.
The NMI is defined as
\begin{eqnarray}
NMI(\bm{b},\bm{a})=\frac{MI(\bm{b},\bm{a})}{\sqrt{(H(\bm{b}) H(\bm{a}))}}
\end{eqnarray}
where $H(\bm{b})$ and $H(\bm{a})$ are the entropies of $\bm{b}$ and $\bm{a}$.
$MI(\bm{b},\bm{a})$ is the mutual information metric of $\bm{b}$ and $\bm{a}$.
The larger NMI is, the better the clustering result is.

We use HCHC to learn the temporal features of COVID-19 by a dataset that includes the available COVID-19 daily information of the different states in the USA from 21/1/2020 to 21/1/2022.
There are 37525 samples in this dataset and their features are detailed in Table~\ref{features}.
%There are 37525 samples in this dataset and their attributes include (1) daily cases, (2)  daily cases per $100K$ capita, (3) daily deaths, (4) daily deaths per $100K$ capita, (5) the percentage of total cases / total deaths, 
%(6) the percentage of total cases / population, (7) weekly cases, (8) the change of weekly cases, (9) the cases in $28$ days, and (10) the change of the cases in $28$ days.
\begin{table*}
\centering
\footnotesize
\caption{The mapping colours and corresponding time periods (day/month/year).}
\label{mapcolor}
\begin{tabular*}{15cm}{@{\extracolsep{\fill}}ccccc}	
  \hline
 Sample color & (a) & (b) & (c) & (d) \\
 \hline
        blue          &22/1/20 to 21/9/20    &22/1/20 to 21/7/20         &22/1/20 to 21/6/20      & 22/1/20 to 21/5/20  \\
	    green       &22/9/20 to 21/5/21    &22/7/20 to 21/1/21         &22/6/20 to 21/11/20     &22/5/20 to 21/9/20 \\
       purple         &22/5/21 to 21/1/22    &22/1/21 to 21/7/21        &22/11/20 to 21/4/21     & 22/9/20 to 22/1/21 \\
        cyan          &-                     &22/7/21 to 21/1/22        &22/4/21  to 21/9/21     &22/1/21 to 21/5/21   \\
	   olive         &-                     &-                          &22/9/21  to 21/1/22     &22/5/21 to 21/9/21 \\
       dark-blue      &-                     &-                          &-                        &22/9/21 to 21/1/22 \\
 
\hline
\end{tabular*}
\end{table*}
\begin{table*}[h]
\centering
\caption{The features of the COVID-19 dataset.}
\label{features}
\begin{tabular*}{8cm}{@{\extracolsep{\fill}}cc}
  \hline
 Feature                                 & Brief explanation \\
 \hline
  \#1                                 & daily cases                     \\
  \#2                                 & daily cases per $100K$ capita\\
  \#3                                 & daily deaths   \\
  \#4                                 & daily deaths per $100K$ capita \\
  \#5                                 & the percentage of total cases / total deaths                          \\
  \#6                                 & the percentage of total cases / population                   \\
  \#7                                 & weekly cases                        \\
  \#8                                 & the change of weekly cases                          \\
 \#9                                 & the cases in $28$ days \\
  \#10                                & the change of the cases in $28$ days \\
  \hline
\end{tabular*}
\end{table*}

\section{Supplementary Experiment}
\label{SuppF}
\subsection{Supplementary Experiment for Covid-19 Dataset}
\label{SuppF1}
\begin{figure}[]
\center
\subfigure[Labeled by every $8$ months]{\includegraphics[width=1.5in]{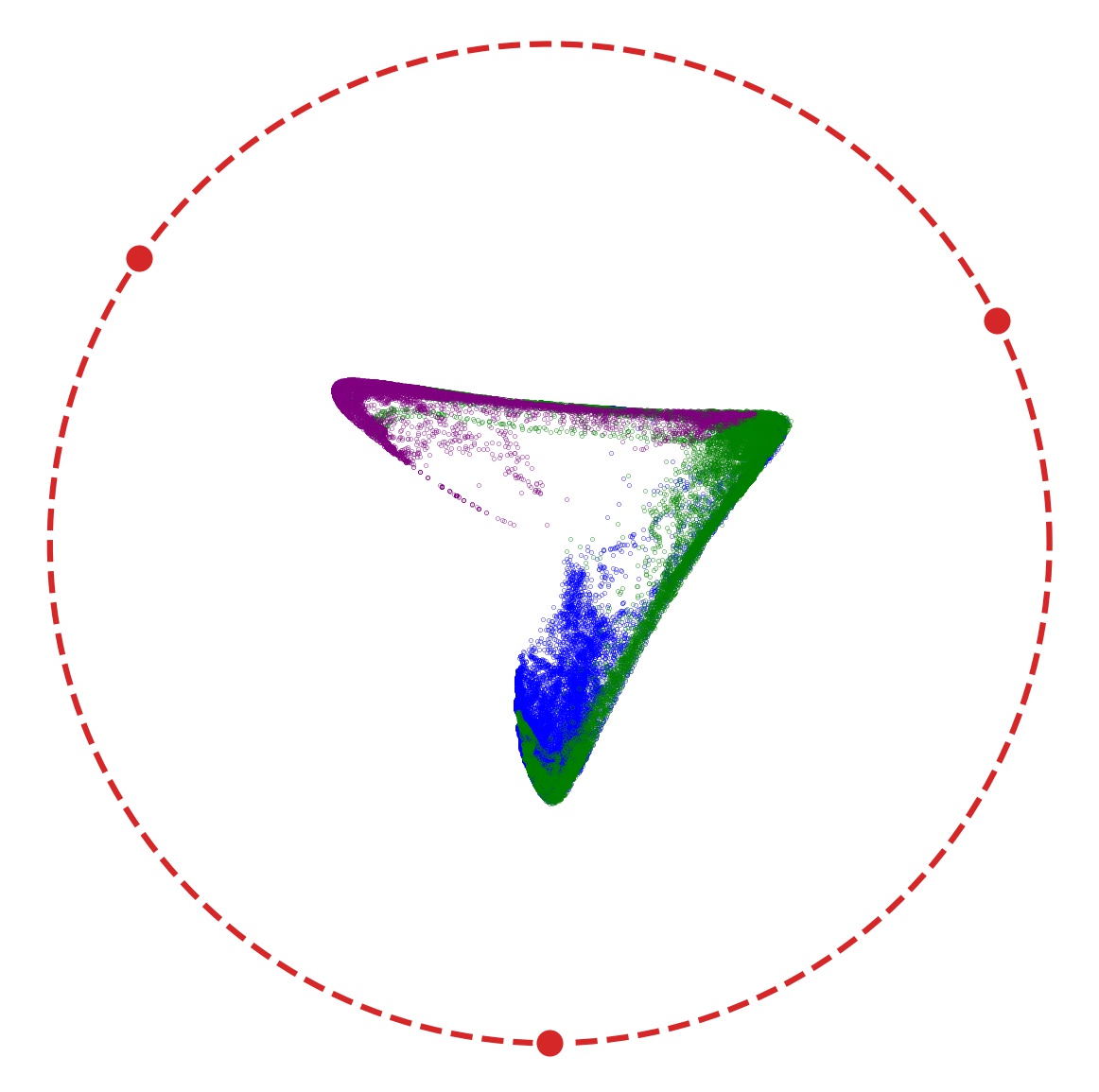}}
\subfigure[Labeled by every $6$ months]{\includegraphics[width=1.5in]{USA4color.jpg}}
\subfigure[Labeled by every $5$ months]{\includegraphics[width=1.5in]{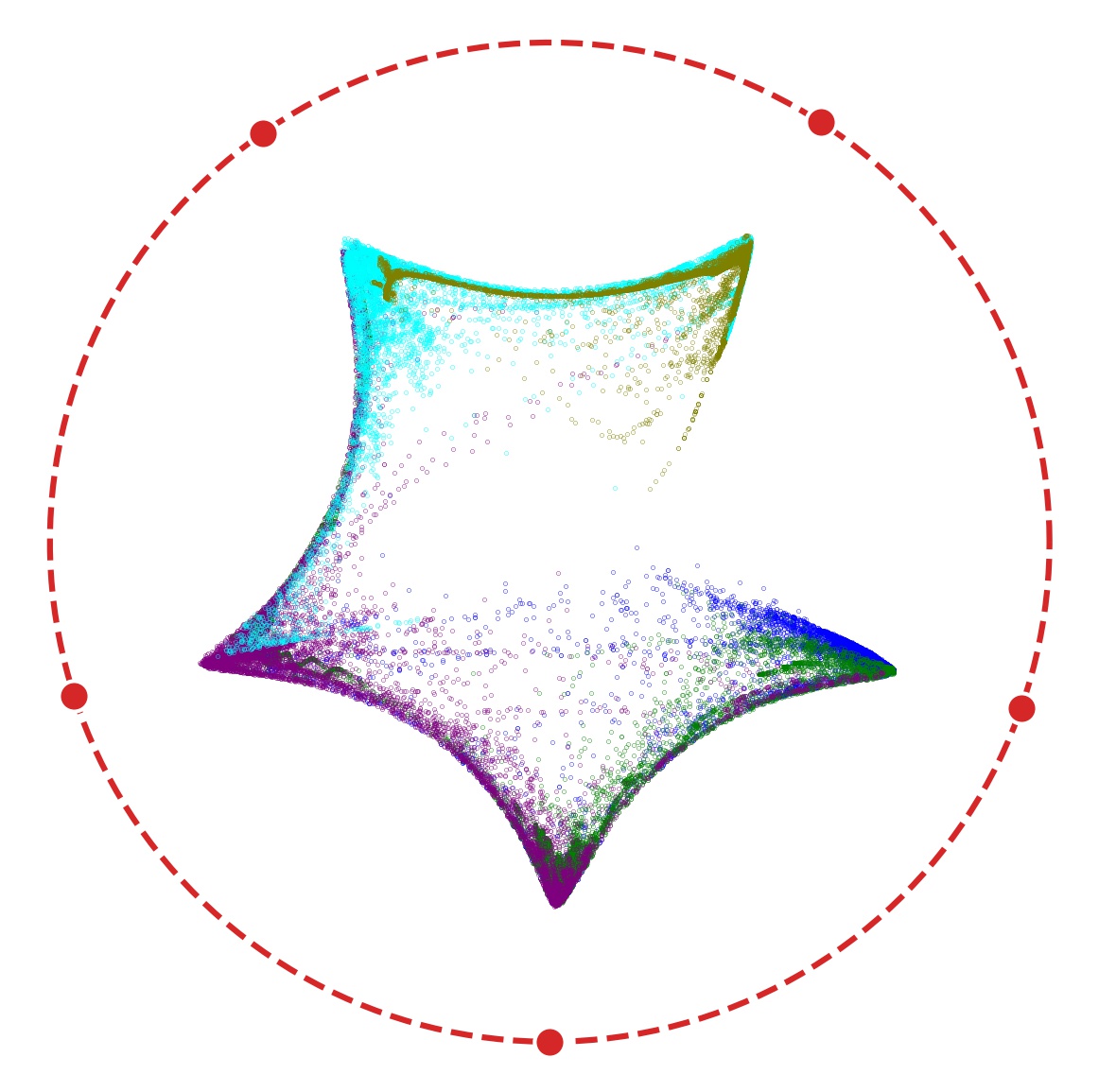}}
\subfigure[Labeled by every $4$ months]{\includegraphics[width=1.5in]{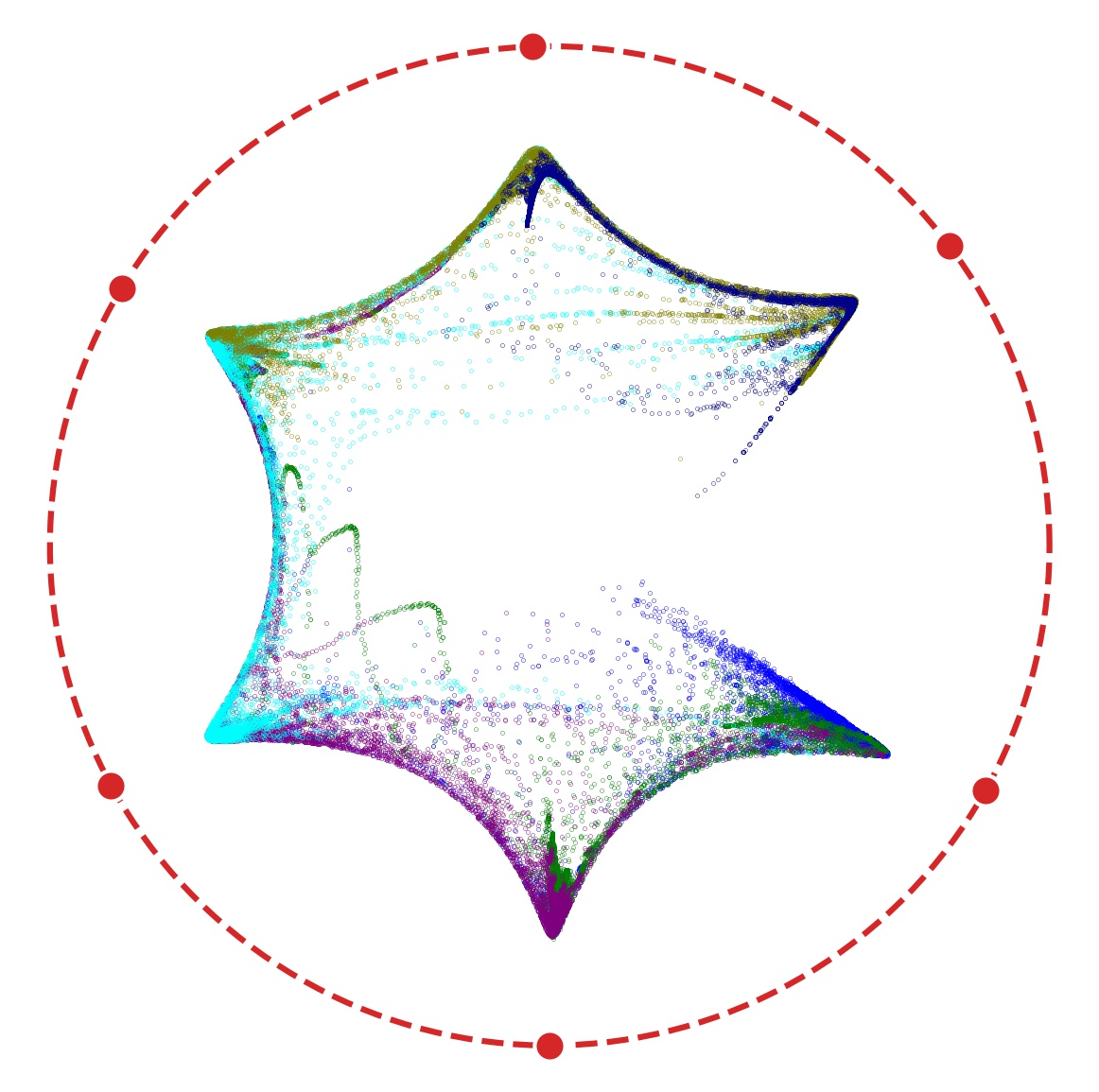}}
\caption{The results of HCHC on the COVID-19 dataset.}
\label{scUSA}
\end{figure}
In Fig.~\ref{scUSA} (a) where the data are labeled by every $8$ months, the samples from 22/9/20 to 21/5/21 are divided into two clusters;
In Fig.~\ref{scUSA} (c) where the data are labeled by every $5$ months, the samples from 22/1/20 to 21/6/20 and 22/6/20 to 21/11/20 are clustered in one cluster, the samples from 22/11/20 to 21/4/21 are divided into two clusters,
and the samples from 22/4/21 to 21/9/21 are also divided into two clusters;
In Fig.~\ref{scUSA} (d) where the data are labeled by every $4$ months, except the samples which from 22/1/20 to 21/5/20 and 22/5/20 to 21/9/20 are clustered into one cluster, the samples from any other period are divided into two clusters.
Whereas, in Fig.~\ref{scUSA} (b) where the data are labeled by every $6$ months, the samples from the same period can be mapped close to the same cluster anchor.

\subsection{Time Complexity Analysis}
\label{SuppF2}
\begin{figure}
\centering
  \includegraphics[width=3.5in]{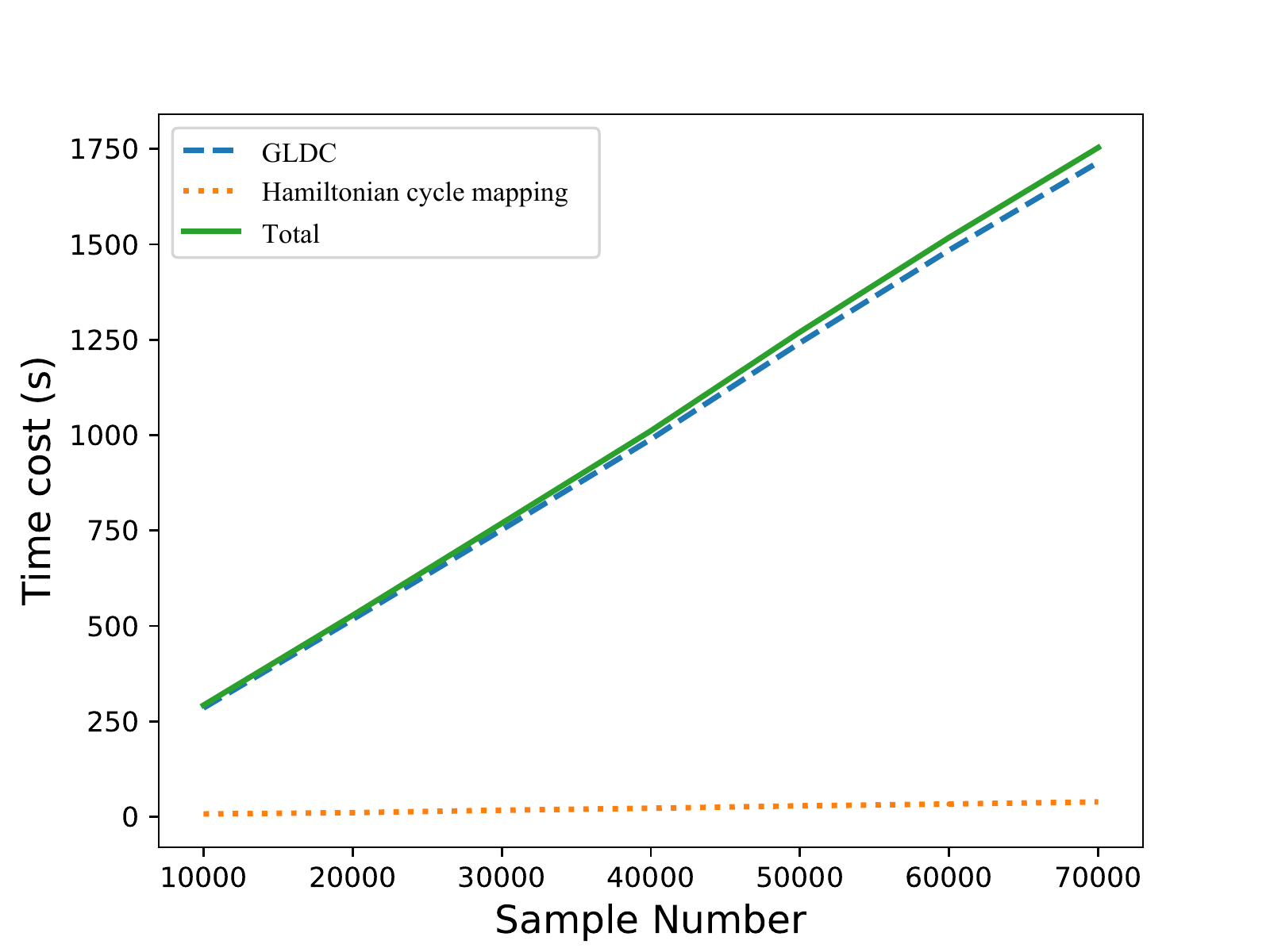}\\
  \caption{The time cost of HCHC.}
  \label{timecost}
\end{figure}

The time cost of our HCHC is linear to the number of the samples in data,
thus it can effectively visualize the clustering results in big data.
Fig.~\ref{timecost} shows the time cost of HCHC on MNIST with different numbers of the samples.
The epoch number is set as $200$.
We can find that the time cost of GLDC is more than the time cost of Hamiltonian cycle mapping on MNIST dataset.
%Therefore one way to reduce the time cost of HCHC is to reduce the number of samples for training the network.
%Fig.~\ref{reducet} shows the visualized results of training the network by only 10000 samples in MNIST and Fashion, respectively.
%The visualized results in this figure are similar to the visualized results of training the network by the whole datasets.
%\begin{figure}
%\center
%\subfigure[]{\includegraphics[width=2in]{MNISTT.jpg}}
%\quad
%\subfigure[]{\includegraphics[width=2in]{FashionT.jpg}}

%\caption{The visualized result of training the network by part of the data. (a) MNIST. (b) Fashion.}
%\label{reducet}
%\end{figure}
\subsection{Supplementary Experiment for Different Visualization Methods}
\label{SuppF3}
\begin{figure}[]
\center
\includegraphics[width=6.5in]{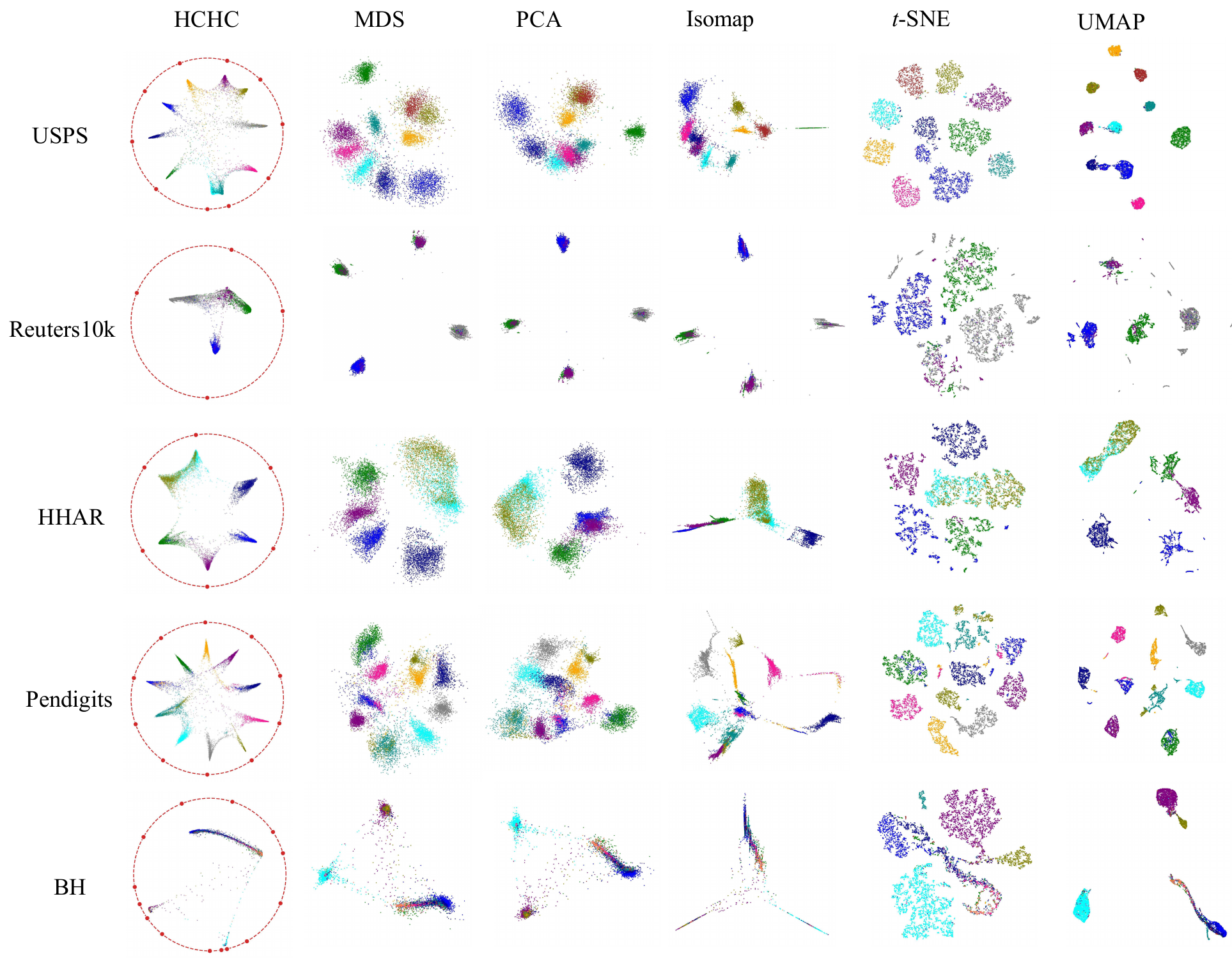}

\caption{The visualized results of the different visualization methods. Samples are coloured by their labels.}
\label{casestudyU}
\end{figure}

Here we show the visualized results of HCHC, MDS, PCA, Isomap, $t$-SNE, and UMAP on the datasets of USPS, Reuters10k, HHAR, Pendigits, and BH.
The results are shown in Fig.~\ref{casestudyU}.

From Fig.~\ref{casestudyU}, we can see that for USPS, HCHC can better better present the cluster number and index the clutering results than MDS, PCA, and Isomap.
It also can better present the cluster similarities and outliers than $t$-SNE and UMAP for USPS and BH.
For HHAR, Pendigit, and BH, with the help of anchors, HCHC can better present the cluster number and index the clustering results than any other visualization method.
HCHC also can better present the cluster similarities than $t$-SNE for HHAR and better present outliers than $t$-SNE and UMAP for Pendigit.
For these two datasets, MDS, PCA, and Isomap can better present the cluster similarities and outliers than $t$-SNE and UMAP, but $t$-SNE and UMAP can better present the cluster structure than MDS, PCA, and Isomap.
Although HCHC, PCA, and Isomap can well present the clustering result of Reuters10k, only HCHC can well present the similarities between the purple class, grey class, and green class.

\subsection{Case Study}
\label{SuppF4}
\begin{table*}[]
\centering
\footnotesize
\caption{Expert comparison of different visualization methods on MNIST.}
\label{caseuserM}
\begin{tabular*}{8cm}{@{\extracolsep{\fill}}ccccc}	
  \hline
Method & Task (a) & Task (b) & Task (c) & Average \\
 \hline
        MDS           &3.3    &4.2     &4.3     & 3.9  \\
	    PCA           &3.6    &3.9     &4.5     &4 \\
       Isomap         &3.9    &4.4     &4.1   & 4.1 \\
        $t$-SNE       &4.5    &2.1     &2.5      &3.0   \\
	   UMAP           &4.9    &3.2     &1.5    &3.2\\
       HCHC           &4.5    &4.6     &4.5    &4.5 \\
 
\hline
\end{tabular*}
\end{table*}
\begin{table*}[]
\centering
\footnotesize
\caption{Expert comparison of different visualization methods on Fashion.}
\label{caseuserF}
\begin{tabular*}{8cm}{@{\extracolsep{\fill}}ccccc}	
  \hline
 Method & Task (a) & Task (b) & Task (c) & Average \\
 \hline
        MDS           &2.1    &4.6     &4.3     & 3.7  \\
	    PCA           &2.4    &4.6     &4.3     &3.8 \\
       Isomap         &1.8    &4.3     &4.1  & 3.4 \\
        $t$-SNE       &3.0    &4.0     &3.5      &3.5   \\
	   UMAP           &3.3    &3.5     &2.1   &3.0\\
       HCHC           &4.0    &4.4     &4.4    &4.3 \\
 
\hline
\end{tabular*}
\end{table*}
This subsection summarizes the feedback from the ten experts on different visualization methods for clustering.
Five of the experts major in clustering and five of the experts major in visualization.
We design the following three tasks to evaluate the quality of different visualization methods.
Task (a) is to verify the ability of each visualization method to present the number of clusters and correctly identify which cluster a sample belongs to. 
Task (b) is to verify the ability of each visualization method to present the similarities between the samples from different classes. 
Task (c) is to verify the ability of each visualization method to present the outliers. 
Each task is scored on a scale from 0 to 5.
The results of six visualization methods including HCHC, MDS, PCA, Isomap, $t$-SNE, and UMAP are used to visualize the clustering results of MNIST-test and Fashion-test in Fig.~\ref{casestudyM} and \ref{casestudyF}, respectively.
Then based on the visualization results, the ten experts are asked to score these visualization methods on our three designed tasks.

The average scores of the ten experts in different tasks for MNIST-test are shown in Table~\ref{caseuserM}.
For the results of MNIST-test, as we can see, HCHC can well perform all of our three designed tasks and thus get the highest average score.
However, MDS, PCA, and Isomap cannot well perform task (a).
For example, by MDS and PCA, we are hard to correctly identify the pink, green, cyan, and dark cyan clusters.
By Isomap, we are hard to correctly identify grey and dark blue clusters.
$t$-SNE and UMAP cannot well perform tasks (b) and (c).
For example, the shapes of handwriting digit numbers ``9” and ``4” are similar, but $t$-SNE and UMAP cannot map the clusters of these two digit numbers close to each other.

The average scores of in different tasks for Fashion-test are shown in Table~\ref{caseuserF}.
For the results of Fashion-test, as we can see, HCHC also can well perform all of our three designed tasks and thus get the highest average score.
For task (a) With the help of the anchors, our HCHC can better visualized the cluster number and index cluster result than any other visualization methods.
For task (b) HCHC, MDS, PCA, and Isomap can well present between the samples of shirt and the samples of dress, but $t$-SNE and UMAP cannot do so.
We are also hard to see the outliers by $t$-SNE and UMAP.

\subsection{Parameter Analysis}
\label{SuppF5}
\begin{figure}[]
\center
\subfigure[MNIST-$\beta_1$]{\includegraphics[width=2.5in]{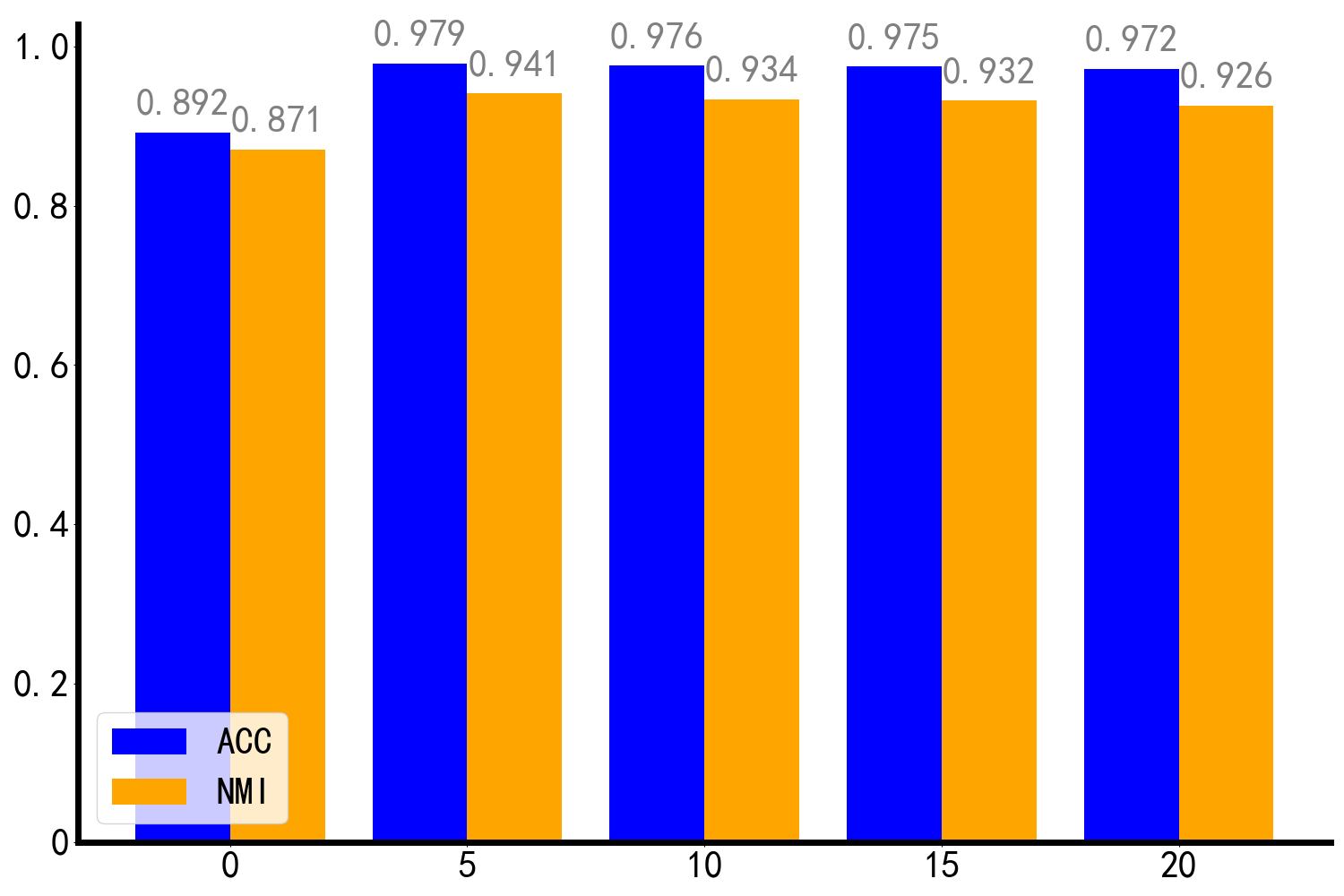}}
\subfigure[MNIST-$\beta_2$]{\includegraphics[width=2.5in]{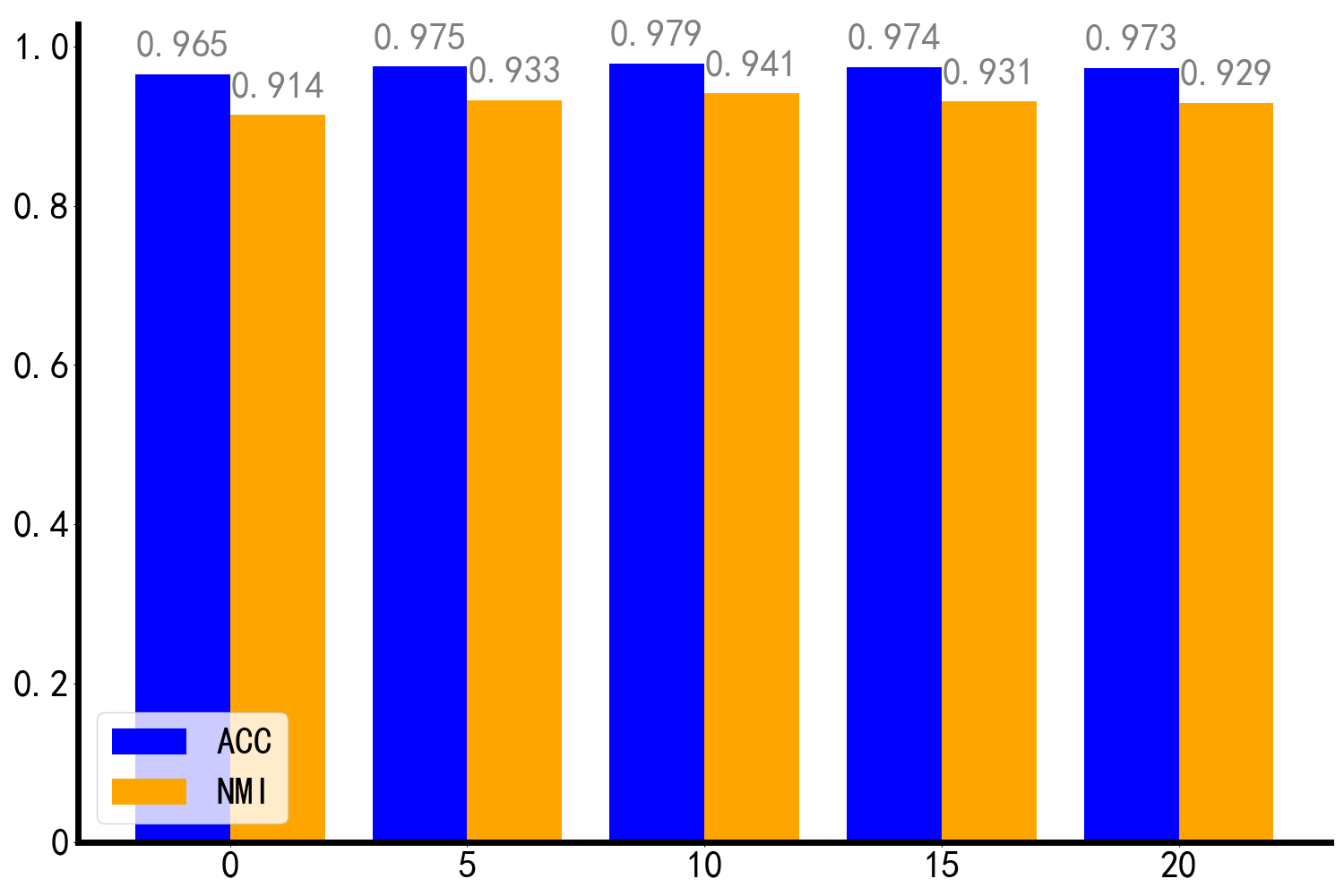}}
\subfigure[HHAR-$\beta_1$]{\includegraphics[width=2.5in]{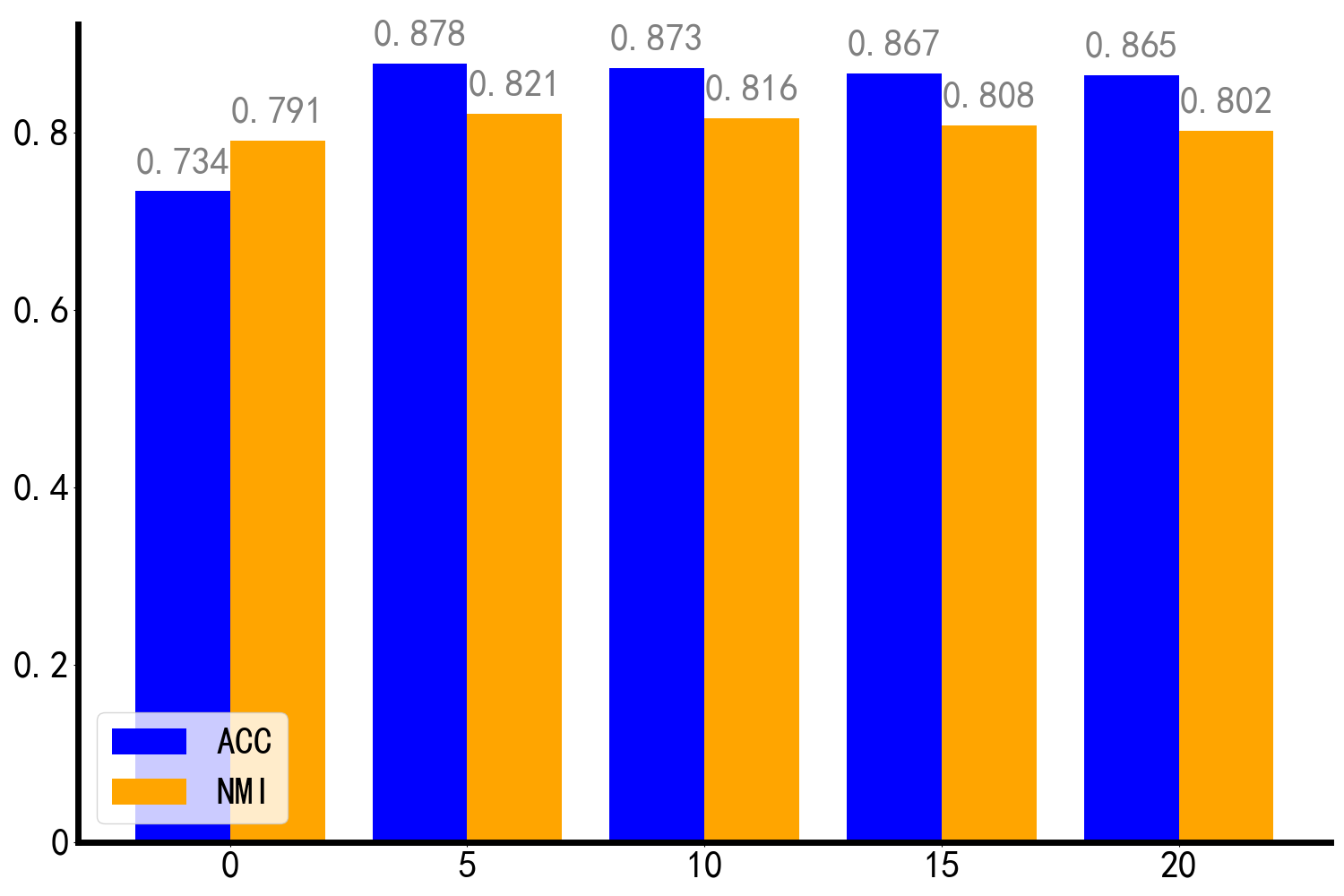}}
\subfigure[HHAR-$\beta_2$]{\includegraphics[width=2.5in]{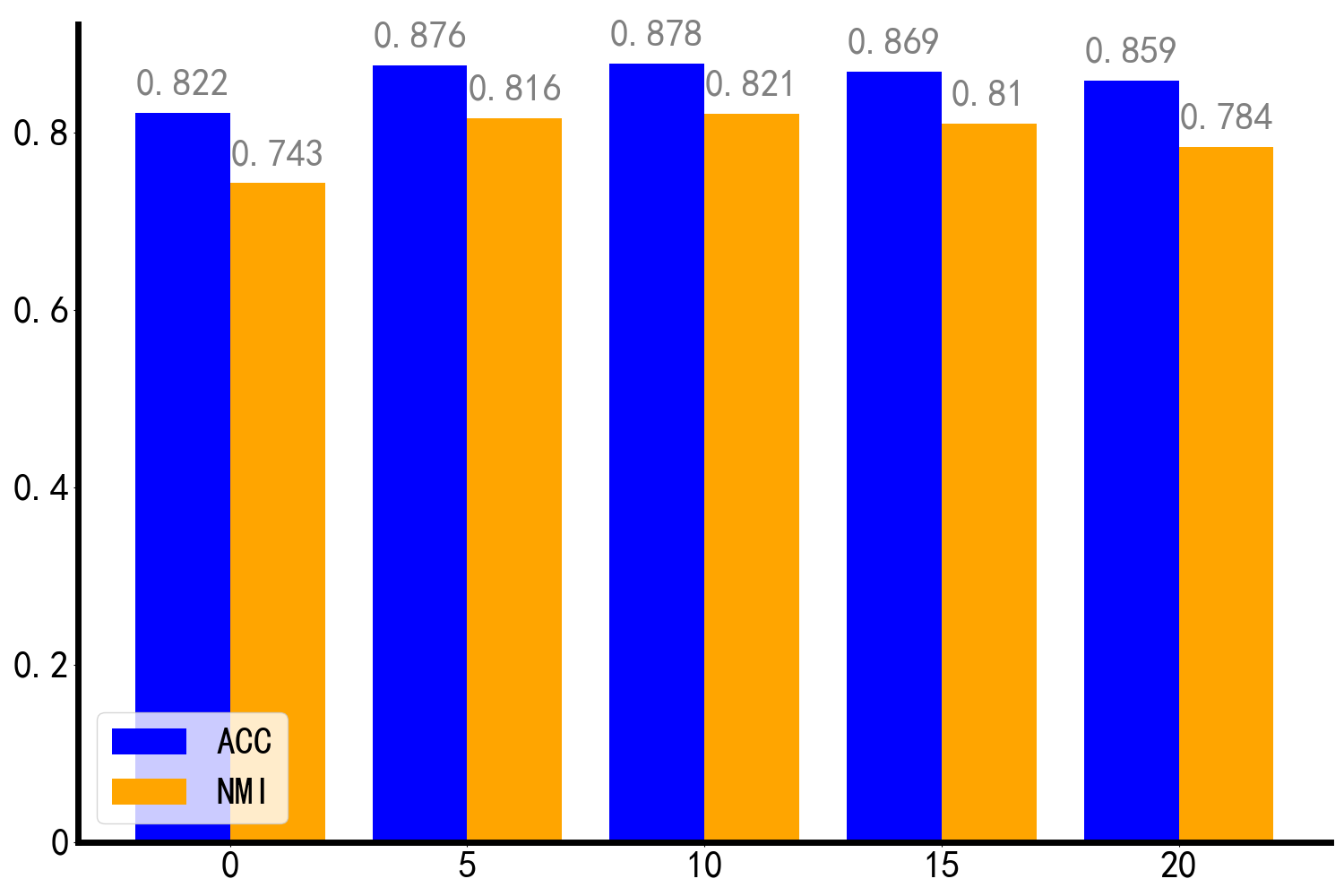}}
\caption{Parameter analysis.}
\label{Paraana}
\end{figure}

In this subsection, we analyze the parameter sensitivities of $\beta_1$ and $\beta_2$ in our objective function on MNIST and HHAR.
The results are shown in Fig~\ref{Paraana}.
In subfigure (a) and (c), $\beta_2$ is fixed as $10$. 
In subfigure (b) and (d), the initial $\beta_1$ is fixed as $5$. 
It can be seen that our method is not sensitive to these two parameters.

%\begin{minipage}[c]{0.5\textwidth}
%\centering
%\captionof{table}{Parameter analysis of $\beta_1$ on MNIST.}
%\begin{tabular*}{7cm}{@{\extracolsep{\fill}}cccccc}
%\hline
%$\beta_1$  & 0 & 5  &10 & 15  & 20 \\
%\hline
% ACC &0.892	&0.979	&0.976	&0.975	&0.972 \\
% NMI &0.871	&0.941	&0.934	&0.932	&0.926 \\
%\hline 
%\end{tabular*}
%\label{pp1}
%\end{minipage}
%\begin{minipage}[c]{0.5\textwidth}
%\centering
%\captionof{table}{Parameter analysis of $\beta_2$ on MNIST.}
%\begin{tabular*}{7cm}{@{\extracolsep{\fill}}cccccc}
%\hline
%$\beta_2$  & 0 & 5  &10 & 15  & 20 \\
%\hline
% ACC & 0.965	&0.975	&0.979	&0.974	&0.973\\
% NMI & 0.914	&0.933	&0.941	&0.931	&0.929  \\
%\hline
%\end{tabular*}
%\label{pp2}
%\end{minipage}

%\begin{minipage}[c]{0.5\textwidth}
%\centering
%\captionof{table}{Parameter analysis of $\beta_1$ on HHAR.}
%\begin{tabular*}{7cm}{@{\extracolsep{\fill}}cccccc}
%\hline
%$\beta_1$  & 0 & 5  &10 & 15  & 20 \\
%\hline
% ACC & 0.734	&0.878	&0.873	&0.867	&0.865 \\
% NMI & 0.791	&0.821	&0.816	&0.808	&0.802 \\
%\hline 
%\end{tabular*}
%\label{pp3}
%\end{minipage}
%\begin{minipage}[c]{0.5\textwidth}
%\centering
%\captionof{table}{Parameter analysis of $\beta_2$ on HHAR.}
%\begin{tabular*}{7cm}{@{\extracolsep{\fill}}cccccc}
%\hline
%$\beta_2$  & 0 & 5  &10 & 15  & 20 \\
%\hline
 %ACC &0.822	&0.876	&0.878	&0.869	&0.859\\
 %NMI &0.743	&0.816	&0.821	&0.810	&0.784 \\
%\hline
%\end{tabular*}
%\label{pp4}
%\end{minipage}

%%%%%%%%%%%%%%%%%%%%%%%%%%%%%%%%%%%%%%%%%%%%%%%%%%%%%%%%%%%%%%%%%%%%%%%%%%%%%%%
%%%%%%%%%%%%%%%%%%%%%%%%%%%%%%%%%%%%%%%%%%%%%%%%%%%%%%%%%%%%%%%%%%%%%%%%%%%%%%%

\end{document}